%% file: main.tex
\documentclass{article}

\PassOptionsToPackage{numbers,compress}{natbib}

 \usepackage[main, final]{neurips_2026}

\usepackage[utf8]{inputenc} 
\usepackage[T1]{fontenc}    
\usepackage{hyperref}       
\usepackage{url}            
\usepackage{booktabs}       
\usepackage{amsfonts}       
\usepackage{nicefrac}       
\usepackage{microtype}      
\usepackage{xcolor}         
\usepackage{amsmath}
\usepackage{amssymb}
\usepackage[pdftex]{graphicx}
\usepackage{booktabs}
\usepackage{cancel}
\usepackage{multicol}
\usepackage{float}
\usepackage{xcolor}

\usepackage{xcolor}
\usepackage{bm}

\definecolor{rankRed}{HTML}{C62828}
\definecolor{rankOrange}{HTML}{D35400}
\definecolor{rankYellow}{HTML}{9A7600}
\definecolor{rankLightGreen}{HTML}{548D22}
\definecolor{rankGreen}{HTML}{00866A}
\definecolor{rankDarkGreen}{HTML}{005A30}

\newcommand{\best}[1]{\ensuremath{\mathbf{#1}}}

\title{Isotropic Activation Functions Enable\\Deindividuated Neurons and Adaptive Topologies}

\author{%
  George Bird\\
  Department of Computer Science\\
  University of Manchester\\
  Manchester, UK\\
  \texttt{george.bird@postgrad.manchester.ac.uk} \\
}

\begin{document}

\maketitle

\begin{abstract}
Varying the number of neurons while preserving function is difficult because neural networks' dense connectivity creates strong computational interdependence between units. Elementwise activation functions, particularly their limited equivariance properties, are identified as a central hurdle to achieving this. We propose activation functions designed around chosen equivariances, enabling dense neural networks to adapt via singular value decomposition of their affine maps. 
This diagonalises layers into one-to-one, ordered connections, making it easier to assess the impact of individual connections on function. Low-impact neurons can be removed (neurodegeneration), and a thresholded buffer of largely inactive `scaffold' neurons is maintained (neurogenesis). These symmetry-led diagonalisation and structural changes prioritise function-invariance: neurogenesis exactly preserves function, neurodegeneration is asymptotically function-preserving with the pruned singular-value threshold. Moreover, this construction enables asymptotic 50\% parameter sparsification of dense networks while preserving function analytically. Thus, the approach enables real-time restructuring of the architecture in response to task demands, task appending, removal or changes.
Conceptually, this motivates broader applications of symmetry prescriptions at the small scale, even when no data-driven model-scale symmetry adherence is required. Following this, we build activations around symmetry, not from classical assumptions about individual neurons. Unlike typical elementwise forms, these alternatives feature explicit basis independence and do not individuate neurons. We leverage this to decompose and reinterpret neurons across sequential layers, simplifying structure and providing a natural route to adaptive topologies.
Additional developments include a new tunable parameter, the `intrinsic length', while diagonalisation offers new possibilities for interpretability and real-time monitoring, and isotropic functions enable parallelised precomputation of matrix-vector products.

\end{abstract}


\section{Introduction}
    Neural plasticity in the quantity and connectivity of neurons is biologically advantageous \cite{Axelrod2023, Wen2024, Caroni2012} and routinely occurs during mammalian brain development \cite{Mishra2016, LaRosa2018, Dori2021, Holtmaat2009}. Shown to improve neural efficiency through controlled pruning \cite{Riccomagno2015, Scholl2021}, with a benefit to initial overabundance \cite{Scholl2021}, whilst enabling the accumulation of knowledge \cite{Caroni2012, Holtmaat2009}, improving robustness \cite{Turrigiano2000} and function through growth \cite{Chklovskii2004, Holtmaat2009}. Hence, it is hypothesised that analogous behaviour within artificial neural networks will be similarly beneficial. The practical consequences of a network that can reliably adapt its architecture and connectivity in a real-time controlled manner may be transformative --- and this constitutes a broad goal of architecture-based continual learning approaches \cite{Wang2024}. 


    Contemporary primitives, e.g. activation functions, many initialisations, normalisers, optimisers, and more, are typically constructed using elementwise functional forms. Historically, forms were inspired by biological neural systems \cite{McCulloch1943}: 
    ``\textit{[...] the conformity of their relations to those of the logic of propositions, insure that the relations of psychons are those of the two-valued logic of propositions}'', an initial framing helping to naturalise the eventual pervasive implementation of activation vectors being decomposed in a basis-dependent manner, interpreted as individual artificial `neuron'-like units composed with scalar operations. Modern elementwise formulae apply univariate maps independently to components; a construction generalising from this original individuated neuron-derived premise. 


    This duality of individuated yet interdependent neuron construction makes adapting architectures while closely preserving function challenging---especially for neurodegeneration.
    This is one reason connectivity pruning \cite{Lecun1989b, Hassibi1992, Hassibi1993, Han2015, Guo2016, Lee2019, Evci2021}, rather than highly-connected whole-neuron removal \cite{Mozer1988, Chen2016, Ruso2016, Wei2016, Yoon2018, Kim2020}, is typical. 
    
    This `individuation' limitation is inherent to most contemporary primitives, with few exceptions. In particular, those classified by their equivalence class under reparameterisation \cite{Sussmann1992, Badrinarayanan2015} admitting permutation-structured groups, e.g., permutation, hyperoctahedral, scaled-permutations, etc. \cite{Godfrey2023}. We posit that generalising primitive construction instead through a broader symmetry-\textit{prescribed}\footnote{`Prescribed' indicates intentional introduction of a target symmetry for a purpose, contrasting to studying those `deduced' in a pre-existing form. We motivate this utility even at primitive-scale, without requiring data-driven model-scale symmetries.} approach is beneficial \cite{Bird2025a, Bird2025b, Bird2025c}. Exploring this allows novel functional forms and generally new behaviours, specifically via intentionally introduced reparameterisation invariances in the present work. These symmetry-prescribed primitive reformulations are central to the adaptive topology approach presented: replacing the typical, derived permutation symmetry with a chosen alternative.
    
    Using basis-independent formulations as the goal, one can replace the discrete permutation symmetry with a continuous one. Namely, \textit{orthogonal equivariance} symmetry naturally serves as a foundation for such altered activation functions. This is because this particular choice of continuous group prescription still permits non-linear functions on the magnitude, making it well-suited for activation functions; meanwhile, the orthogonal group is a supergroup of the permutation group, $\mathrm{O}\left(n\right)\supset S_n$, all while it ensures basis-independence by being direction-agnostic (`isotropic') in effect. Hence the orthogonal groups offer a suitably minimal and natural generalisation for such reformulations. \textit{Emergent} from this symmetry redefinition is a more generalised concept of neurons, which are `deindividuated', since no function-distinguished basis exists for decomposition \cite{Elhage2022, Bird2025a, Bird2025c}.

    Enabled by this is our methodology for dynamic topologies. It builds on the observation that these orthogonal-equivariant, symmetry-derived reformulations, termed ``isotropic primitives'', commute by definition with orthogonal-group actions. Considering this in tandem with singular value decomposition (SVD) for dynamical networks, or independent component analysis (ICA) \cite{Herault1984, Ans1985, Comon1994} for interpretable networks, both of which yield standard orthogonal matrices that only now commute due to the symmetry-led design. Combining SVD's properties thus enables a diagonalised topology by reparameterising the networks with its orthogonal matrices into a simpler one-to-one connectivity. This simplifies the complexity of full neurodegeneration by \textit{making it equivalent to connectivity pruning}. Similarly, an ICA representation derived from dataset activations may better represent a network for interpretability and real-time monitoring, up to ICA's own concept resolvability. 

    This diagonalisation procedure incurs \textit{no functional degradation} analytically, owing to the primitive's basis independence. In practice, only machine precision, and propagation thereof, cause negligible alteration. Adaptive topologies are then achieved by thresholding these diagonalised entries to trigger real-time adaptations in response to task demand and learning. These are exactly function-preserving for neurogenesis and, given by the thresholding, arbitrarily well approximated for neurodegeneration. Hence, equivalence classes expand to full architectural adaptations, not just static reparameterisations. 

    Overall, this is unique to continuous-symmetry-prescribed primitives, and it is not possible for classical elementwise functions constrained to discrete parameter relabelling. Hence, we demonstrate flexible use cases for these alternative isotropic formulations, whilst motivating a broader exploration of primitive symmetry prescription for benefits in \textit{general} practice. 
    The primary contributions are:

    \begin{enumerate}
        \item \textbf{Conceptual Contribution}: Functional forms are historically derived from individual neurons as fundamental units to compose; from this, symmetries are \textit{since deduced}. Instead, we propose an ontological reversal: treating symmetry as fundamental and neurons as emergent. Adopting symmetry-prescriptive foundations offers an expanded class of admissible functional forms that underpin \textit{neural} learning systems, unconstrained by biological cell-like conceptions and yielding generalised notions of neurons induced by group representations. \textit{Eqns.}~\ref{Eqn:Algebraic}~to~\ref{Eqn:Closure} formalise this in an organising taxonomy for current and novel functional forms by their descriptive and prescriptive symmetries. Induced effects may be assigned to and compared across categories or their compositions, offering utility for general applications by understanding symmetry-driven properties and biases. This constitutes the broader objective that the adaptive architectures begin to pursue and evidence. 
        
        \item \textbf{Reformulation Contribution}: With this enriched design space, isotropic primitives are designed to exhibit a crucial basis independence absent in elementwise formulae. 
        
        \item \textbf{Practical Contribution}: Leveraging this allows affine maps to be represented diagonally, with singular-value thresholding triggering architecture changes, alongside asymptotic $50\%$ perfect sparsification, ICA representations and parallelised matrix-vector products.
    \end{enumerate}

    \paragraph{Related Work:} Various approaches to architecture adaptation exist, including neuroadaptive-like methods \cite{Mozer1988, Chen2016, Ruso2016, Wei2016, Yoon2018, Kim2020} and also connectivity-pruning \cite{Han2015, Guo2016, Lee2019, Evci2021, Evci2022}. The former represent the closest related work, varying the number of neurons, not solely connectivity; examples include Net2Net \cite{Chen2016}, Network Morphism \cite{Wei2016} and Neuron Merging \cite{Kim2020}. Net2Net considers neurogenesis only through weight-column duplication in elementwise networks, showing function preservation when outgoing weights sum to the original; it also considers Net2DeeperNet as a method to append additional layers invariantly—Network Morphism extends this work, generalising both approaches. Neuron Merging considers an analogous case for neurodegeneration only and is restricted to solely ReLU activations. Moreover, due to distinctive primitives, the present approach bears limited similarity to these prior implementations. Works such as SNIP \cite{Lee2019} and RigL \cite{Evci2021} diverge more considerably by addressing connectivity sparsity instead, showing network performance can be maintained with significantly fewer connections rather than adapting width through neuron count. Separately, prior work \cite{Ganev2022} arrived at a related use of orthogonal equivariant activations, studied for the development of a special-case Step-ReLU universal approximation theorem. They also identified a markedly different exploitation of basis change via QR decomposition and preactivation span to obtain a distinct form of model compression \cite{Ganev2022}. This enriches the potential utility of such altered activations, and their compression appears to compound with our distinct connectivity-sparsification capabilities, and may aid in streamlining successive layers in neuroadaptation. \textit{App.}~\ref{App:Related} compares all these works in depth.



\section{Theoretical Background}

    An overview of the symmetry-prescribed framework is provided, and it is then used to define isotropic activation functions that commute with orthogonal-group actions. Finally, layerwise diagonalisation and sparsification are detailed, culminating in the adaptive topology procedure. This is detailed for multilayer perceptrons, whilst \textit{App}~\ref{App:Convolution} presents a conceptual generalisation for convolution\cite{Fukushima1980, LeCun1989}. 
\subsection{From Symmetry Taxonomy to Primitive Reformulation: Isotropic Activation Functions}

    Proposed is a broader organising taxonomy of function (including up to model-scale) symmetries. This is formalised via groups, $\mathcal{G}$, with their elements, $g\in\mathcal{G}$, acting on functions, $f$ of a (potentially singleton) functional class $\mathcal{F}$ through various matrix representations $\mathbf{g}\!=\!\varrho^{(i)}\left(g\right)$, in categorisable ways. These are expressed as three non-exhaustive relations \textit{Eqns.}~\ref{Eqn:Algebraic}~$\Rightarrow$~\ref{Eqn:Probabilistic}~$\Rightarrow$~\ref{Eqn:Closure}. These detail different `strength' symmetry adherence over the functional class, termed ``algebraic'', ``probabilistic'' and ``closure''. Algebraic is more standard \cite{Godfrey2023, Bronstein2021}, but is unconventional in its application over a \textit{functional class}. \textit{Eqn.}~\ref{Eqn:Probabilistic}'s probabilistic category is unnecessary for adaptive networks but provided for completeness, with intentionally flexible probability measure $\mathbb{P}$, capable of classifying initialisations.
    \begin{align}
        \text{Algebraic:}\qquad\forall g \in \mathcal{G}, \;\forall f\in\mathcal{F}:\quad
        & \varrho^{(1)}(g)\circ f\circ\varrho^{(2)}(g^{-1}) = f
        \label{Eqn:Algebraic}\\
        \text{Probabilistic:}\qquad\forall g \in \mathcal{G}, \;\forall f\in\mathcal{F}:\quad
        &\mathbb{P}\!\left(\varrho^{(1)}(g)\circ f\circ\varrho^{(2)}(g^{-1})\right)
        = \mathbb{P}(f)
        \label{Eqn:Probabilistic}\\
        \text{Closure:}\qquad\forall g \in \mathcal{G}, \;\forall f\in\mathcal{F}:\quad
        &\varrho^{(1)}(g)\circ f\circ\varrho^{(2)}(g^{-1}) \in \mathcal{F}
        \label{Eqn:Closure}
    \end{align}
     These taxonomise and structure the design space, assisting comparisons of symmetry-induced properties, utilisable behaviours and structural biases, both individually and in composition. Through careful representation choices, relations can serve as a foundation for activation functions, normalisers, initialisers, optimisers and further primitives, categorised by underlying symmetry. This facilitates novel prescriptive use and use as composable building blocks. In isolation, these do \textit{not} describe reparameterisation invariances (parameter symmetries), which typically require composition: usually pairing one-sided closures through algebraic intertwiners --- a setup utilised in the subsequent section.

    Defining isotropic functions as specifically algebraic-function constraints of standard representations of the orthogonal group family \textit{maximally} (no supergroup), $\varrho:\mathrm{O}\left(n\right)\!\rightarrow\!\mathrm{GL}_n\left(\mathbb{R}\right)$. These orthogonal matrices can be denoted in shorthand as $\mathbf{R}\!\in\!\mathrm{O}\left(n\right)$ for $n$-width layers, and the algebraic constraint of \textit{Eqn.}~\ref{Eqn:Algebraic} can then be simplified to \textit{Eqn.}~\ref{Eqn:IsotropicCommutator}, for a constrained subset of the functional class $\mathbf{f}:\!\mathbb{R}^n\!\rightarrow\!\mathbb{R}^n$.
    \begin{align}
        \forall\mathbf{R}\in\mathrm{O}\left(n\right),\forall\vec{x}\in\mathbb{R}^n:\qquad&\left[\mathbf{R}, \mathbf{f}\right] = \mathbf{R}\circ\mathbf{f}-\mathbf{f}\circ \mathbf{R}=0\quad\text{equally}\quad\mathbf{f}\left(\mathbf{R}\vec{x}\right)=\mathbf{R}\mathbf{f}\left(\vec{x}\right)
        \label{Eqn:IsotropicCommutator}
    \end{align}
    This is the desired standard orthogonal equivariant constraint for activation functions to \textit{maximally} adhere to, yielding rotationally symmetric (isotropic) properties. These can be maximally satisfied by the general multivariate functional form shown in \textit{Eqn.}~\ref{Eqn:IsotropicFunctionalForm}, for general non-linear univariate functions $\sigma,\tilde{\sigma},\bar{\sigma}$, and unit-vector $\hat{x}$. Any supergroup is also feasible for the subsequent diagonalisation, as it preserves this equivariance; however, consider the presence and nature of the respective non-linearity. Several placeholders may be suggested for $\sigma$, e.g. $\tanh$; however, superficially replicating existing activation functions should not be assumed to convey any continued optimality in this isotropic form.
    \begin{equation}   \mathbf{f}\left(\vec{x}\right)=\sigma\left(\left\|\vec{x}\right\|_2\right)\hat{x}\quad\quad \text{or}\quad\quad\mathbf{f}\left(\vec{x}\right)=\tilde{\sigma}\left(\left\|\vec{x}\right\|_2\right)\vec{x}\quad\quad \text{or}\quad\quad \mathbf{f}\left(\vec{x}\right)=\bar{\sigma}\left(\vec{x}^T\vec{x}\right)\vec{x}
        \label{Eqn:IsotropicFunctionalForm}
    \end{equation}
    \paragraph{Recovering Neurons:} Adopting this ontology entails treating neuron individuation not as an intrinsic property of the vector space or network but as an induced artefact of the \textit{chosen} group representations used to construct primitives, under \textit{Eqns.}~\ref{Eqn:Algebraic}~through~\ref{Eqn:Closure}. Matrix representations of the permutation group are generally not fixed under standard orthogonal, $\mathbf{R}\!\in\!\mathrm{O}\left(n\right)\!\setminus\!P$, conjugation: $\left\{\varrho\left(\pi\right)|\pi\!\in\!S_n\right\}\!=\!P\!\neq\!\left\{\mathbf{R}\mathbf{P}_{\pi}\mathbf{R}^T|\mathbf{P}_{\pi}\!\in\!P\right\}$. Thus, constraining a functional form to be \textit{maximally} equivariant to one such matrix representation introduces a \textit{representation-specific} basis dependence in the admitted form. For a conjugated standard permutation representation, $\mathbf{R}\mathbf{P}_{\pi}\mathbf{R}^T$, this anisotropy distinguishes various vectors, particularly the unordered columns of $\mathbf{R}$, selecting an \textit{absolute frame} in the function's codomain. These distinguished columns of $\mathbf{R}$ correspond to, and make special, up to relabelling, the classical historical notion of artificial neurons\footnote{Curiously columns of $\mathbf{C}\!=\!\alpha\mathbf{I}\!+\beta\vec{1}\vec{1}^T$ when similarly unit-scaled provide representation-induced, function-distinguished arcs, with two important cases: non-interfering \cite{Elhage2022} classical neurons \cite{Bird2025a, Bird2025c} and an equiangularly separated simplex \cite{Papyan2020, Kothapalli2023}.}. Such individuation ceases for orthogonal-prescribed isotropic functions: $\left\{\varrho'\left(Q\right)| Q\!\in\!\mathrm{O}(n)\right\}\!=\!O\!=\!\left\{\mathbf{R}\mathbf{Q}\mathbf{R}^T|\mathbf{Q}\!\in\!O\right\}$, where no direction is inherently distinguished by the group representation's admitted functional forms. Thus, group representations precede the individuated, characteristic \cite{Bau2017, Olah2017, Bird2025a} nature of emergent neurons, and are sufficient to recover them. Generalising to deindividuated neurons is crucial for this paper.
\subsection{Layerwise Affine Diagonalisation\label{Sec:Diagonalisation}}
    Full singular value decomposition of weights $\mathbf{W}$ produces $\mathbf{W}\!=\!\mathbf{R}\mathbf{\Lambda}\mathbf{Q}^T$, with diagonal $\mathbf{\Lambda}=\mathbf{\Lambda}\odot\mathbf{I}$, and orthogonal matrices $\mathbf{R},\!\mathbf{Q}\!\in\!\mathrm{O}\left(n\right)$, which can be used to transform an affine map in \textit{Eqn.}~\ref{Eqn:AffineDecomp}.%
    \begin{equation}
        \mathbf{W}\vec{x}+\vec{b}\quad=\quad\mathbf{R}\mathbf{\Lambda}\mathbf{Q}^T\vec{x}+\vec{b}
        \label{Eqn:AffineDecomp}
    \end{equation}
    Then, generalising this to performing the decomposition on the middle affine map of a three-layer perceptron and using the designed orthogonal equivariance of isotropic functions enables an overall function-invariant reparameterisation. This is unusual in that it leverages this \textit{continuous} symmetry definition across \textit{three} layers, acting twice concurrently on both sides of the layer. With the choice of SVD, this forms a diagonalised representation of the middle layer (or \textit{App.}~\ref{App:ICA}, an ICA representation).

\begin{figure}[htb]
        \centering
        \includegraphics[width=0.6\textwidth]{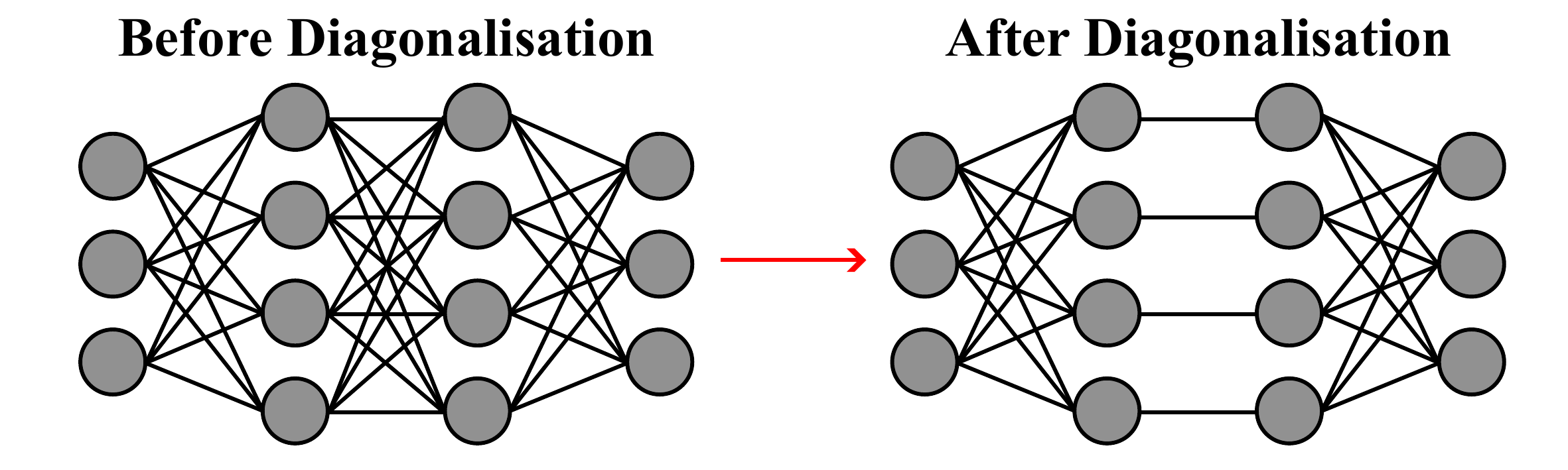}
        \caption{Illustrates the qualitative effects on a network undergoing full diagonalisation. The connectivity is simplified to a one-to-one correspondence between neurons on the chosen layer.}
        \label{Fig:Diagonalised}
    \end{figure}

    For a multilayer perceptron (MLP), this minimally necessitates a standard orthogonal algebraic (\textit{Eqn.}~\ref{Eqn:Algebraic}) constraint on the activation function, $f_{\text{Iso}}$, composed alongside two closure (\textit{Eqn.}~\ref{Eqn:Closure}) constraints on each sandwiching affine layer, $f_{\text{Aff}}$, such that a standard-representation acts inwardly and a trivial-representation outwardly, forming the appropriately required closure-invariances. The latter constraint is already satisfied by the existing $\mathrm{O}\left(n\right)\!\subset\!\mathrm{GL}_n(\mathbb{R})$ two-sided closure of affine maps, enabling the desired equivalence class without modification. One may also \textit{choose}, but not require, an orthogonally symmetric initialiser via \textit{Eqn.}~\ref{Eqn:Probabilistic}. This construction permits two, potentially differing, isotropic activation functions. The three-layer transformations are described in \textit{Eqns.}~~\ref{Eqn:Transform1}~through~\ref{Eqn:Transform5}. Layer widths are not assumed and may vary layer-wise, given this procedure's generality.
    \begin{align}
        \mathbf{f}
        &={f}_{\text{Aff.3}}\circ{f}_{\text{Iso.2}}\circ{f}_{\text{Aff.2}}\circ{f}_{\text{Iso.1}}\circ{f}_{\text{Aff.1}}
        \label{Eqn:Transform1}\\
        \mathbf{f}\left(\vec{x}\right)
        &= \mathbf{W}_3{f}_{\text{Iso.2}}\left(\mathbf{W}_2{f}_{\text{Iso.1}}\left(\mathbf{W}_1\vec{x}+\vec{b}_1\right)+\vec{b}_2\right)+\vec{b}_3
        \label{Eqn:Transform2}
    \end{align}
    Then one can apply the SVD to the weight $\mathbf{W}_2$ whilst inserting an identity matrix, $\mathrm{I}_n=\mathbf{R}\mathbf{R}^T$.
    \begin{align}
        \mathbf{f}\left(\vec{x}\right)
        &= \mathbf{W}_3{f}_{\text{Iso.2}}\Big((\underset{\mathrm{I}_n}{\underbrace{\mathbf{R}\mathbf{R}^T}})\Big(\mathbf{R}\mathbf{\Lambda}\mathbf{Q}^T{f}_{\text{Iso.1}}\left(\mathbf{W}_1\vec{x}+\vec{b}_1\right)+\vec{b}_2\Big)\Big)+\vec{b}_3
        \label{Eqn:Transform3}\\
        \mathbf{f}\left(\vec{x}\right)
        &= \underset{\mathbf{W}_3'}{\underbrace{\mathbf{W}_3\mathbf{R}}}{f}_{\text{Iso.2}}\Big(\mathbf{\Lambda}{f}_{\text{Iso.1}}\Big(\underset{\mathbf{W}_1'}{\underbrace{\mathbf{Q}^T\mathbf{W}_1}}\vec{x}+\underset{\vec{b}_1'}{\underbrace{\mathbf{Q}^T\vec{b}_1}}\Big)+\underset{\vec{b}_2'}{\underbrace{\mathbf{R}^T\vec{b}_2}}\Big)+\vec{b}_3
        \label{Eqn:Transform4}
    \end{align}
    Finally, reparameterising the affine layers with modified weights $\mathbf{W}'_i$, $\vec{W}$ and biases $\vec{b}'_i$.
    \begin{align}
        \mathbf{f}\left(\vec{x}\right)
        &= \mathbf{W}_3'{f}_{\text{Iso.2}}\left(\mathbf{\Lambda}{f}_{\text{Iso.1}}\left(\mathbf{W}_1'\vec{x}+\vec{b}_1'\right)+\vec{b}_2'\right)+\vec{b}_3
        \label{Eqn:Transform5}\\
        \mathbf{f}\left(\vec{x}\right)
        &= \mathbf{W}_3'{f}_{\text{Iso.2}}\left(\vec{W}\odot {f}_{\text{Iso.1}}\left(\mathbf{W}_1'\vec{x}+\vec{b}_1'\right)+\vec{b}_2'\right)+\vec{b}_3
        \label{Eqn:Transform6}
    \end{align}
    Hence, the central layer is diagonalised, whilst the MLP remains function-invariant overall to such a transformation. This layer is simplified to one-to-one connectivity, as depicted in \textit{Fig.}~\ref{Fig:Diagonalised}. The singular values can be ordered, enabling a perturbative-like understanding of the layer's now simplified action --- especially when incorporating normalisation or a magnitude-bounded $\mathbf{f}$. Similarly this can be expressed with a Hadamard product, $\odot$, with parameterisation $\vec{W}=\operatorname{diag}\left(\mathbf{\Lambda}\right)$, as shown in \textit{Eqn.}~\ref{Eqn:Transform6}.

    Practically, complete diagonalisation is not necessary for adaptive topologies or interpretability; instead, a left-sided partial diagonalisation, as shown in \textit{Eqn.}~\ref{Eqn:LeftActing}, is sufficient and marginally more computationally efficient to enact. However, at inference time, full diagonalisation may provide substantial benefit by its significant sparsity whilst remaining analytically function-invariant.
    \begin{equation}
        \mathbf{f}\left(\vec{x}\right)
        = \mathbf{W}_3'{f}_{\text{Iso.2}}\left(\mathbf{\Lambda}\mathbf{Q}^T{f}_{\text{Iso.1}}\left(\mathbf{W}_1\vec{x}+\vec{b}_1\right)+\vec{b}_2'\right)+\vec{b}_3
        \label{Eqn:LeftActing}
    \end{equation}
    Finally, unlike traditional permutation reparameterisations, orthogonal transforms exhibit typical forward-pass function invariance but, novelly, display \textit{variance} in their coupling to most second-order gradient-descent algorithms after the transform. This produces, sequentially backwards from the transform, a training-time-only breaking of the reparameterisation symmetry --- absent at inference time. This optimiser connection may offer an insightful direction for future study, explored in \textit{App.}~\ref{App:GradientCoupling}.
    
\subsubsection{Sparsification\label{Sec:Sparsification}}
    A significant quality arising in isotropic-MLPs stems from this function-invariant diagonalisation. Beyond the adaptive or interpretability considerations, it affords considerable connectivity sparsity. 

    Two directions may be taken, which may be performed just prior to inference time: Joint sparsity optimisation (JSO), which optimises over the multi-layer orthogonal direct-sum Lie group to minimise an objective function of parameter count via a continuous proxy --- many such forms are applicable, including $L_1$-norm-based approaches. Simpler still is an alternating layerwise diagonalisation, which asymptotically achieves an overall 50\% parameter reduction, despite the model function being identical up to floating-point precision. For an odd-depth MLP, \textit{Eqn.}~\ref{Eqn:OddLength} shows the sparsity factor, while \textit{Eqn.}~\ref{Eqn:EvenLength} is for the even-depth case, given $2D+1$ and $2D$ affine layers respectively (for fixed $N$-width, $D$-depth, and $\dim\mathbf{W}+\dim\vec{b}=N^2+N$ parameter counting per layer).
    \begin{align}
        S_{\text{odd}} &= \frac{2DN+\left(D+1\right)\left(N^2+N\right)}{\left(2D+1\right)\left(N^2+N\right)}&&=\frac{1+D^{-1}}{2+D^{-1}}+\frac{2D}{\left(2D+1\right)\left(N+1\right)}\label{Eqn:OddLength}\\
        S_{\text{even}} &= \frac{2\left(D-1\right)N+\left(D+1\right)\left(N^2+N\right)}{2D\left(N^2+N\right)}&&=\frac{1+D^{-1}}{2}+\frac{D-1}{D\left(N+1\right)}\label{Eqn:EvenLength}
    \end{align}
    Both sparsify to 50\% asymptotically as $N, D\rightarrow\infty$, considerably reducing memory footprint and computation. This is not a sparse approximation, but an analytical equivalence under this orthogonal basis transform, offering significant overhead improvements. The formulas serve as upper bounds that may be improved with JSO and/or acceptable function approximation. Moreover, it appears to act near-multiplicatively with QR-compressions \cite{Ganev2022}, while remaining applicable regardless of the MLP's layer-to-layer width structure, unlike QR-compression.

\subsection{Left-Sided Neuroadaptation}
    Neuroadaptation requires less: only the left-sided partial diagonal reparameterisation (\textit{Eqn.}~\ref{Eqn:LeftActing}). This section covers both function-invariant `neurogenesis' involving the maintenance of `scaffold' neurons, and `neurodegeneration': neuron removal that closely approximates function invariance, especially through the introduction of a new parameter: the `intrinsic length', capable of absorbing residual bias.

    Beginning with the addition of scaffold neurons, this neurogenesis process is achieved by embedding the existing $\mathbb{R}^n$ activation space into $\mathbb{R}^{n+1}$. Under the conventional individuated picture, this corresponds to appending additional neurons to the layer, connected so that the model is functionally independent of them. Here, we term these `scaffold neurons', indicating neurons present structurally that do not yet, until specialisation, contribute any function. Alternatively, this is comparable to further increasing the dimensionality of the generalised notion of a neuron, since no basis distinguishes them.

    Such embedding appears largely arbitrary, yet the initialisation may, in practice, involve nuances ranging from explicit to spontaneous symmetry breaking in the new neuron's associated parameters. This raises implications for learning and is related to the aforementioned optimiser-based symmetry violation under orthogonal transforms. In any case, the new singular values should be zeroed, despite a freedom in the corresponding new row of $\mathbf{W}_3'$ denoted in \textit{Eqn.}~\ref{Eqn:LeftActing} for forward growth. 

    Neurodegeneration is the reverse process: producing a well-chosen subspace with appropriate reparameterisations to closely approximate function-invariance. Given the diagonally ordered, positive singular-value parameters, as one tends to zero, the `diagonal-neuron' becomes increasingly independent of the preceding layer. Pruning is achieved by removing corresponding entries and reparameterising surroundings. Moreover, this outgoing connectivity pruning now has exact correspondence to neuron removal, due to the now one-to-one diagonal dependence. The thresholding of the zero-tending singular value can be selected to produce a suitably minimal degradation to model functionality. \textit{Isotropic} normalisers (\textit{App.}~\ref{App:Normalisers}) may help prevent activation-magnitude compensations to shrinking singular values. However, a singular-value-independent bias parameter persists.

    A novel parameter, the scalar "intrinsic length" denoted $o$, is a crucial insight restoring a full neurodegeneration symmetry in the $\mathbf{\Lambda}_{ii}\rightarrow 0$ limit. It transforms under pruning to gauge away the troublesome `residual bias', $\vec{b}_i$. The diagonalised layer, with this novel parameter, can be expressed as \textit{Eqn.}~\ref{Eqn:AffineWithIntrinsicLength}, while \textit{Eqn.}~\ref{Eqn:IntrinsicSymmetry} displays the restored symmetry in the $\mathbf{\Lambda}_{00}\rightarrow 0$ limit, provided the bias in the linear term is absorbed into the next layer's biases appropriately. Intrinsic length's transform also preserves any existing positivity; if desired, it may be parameterised as such, e.g. $o=\exp{\lambda}$.
    \begin{equation}
        \mathbf{f}\left(\vec{x}\right)=\bar{\sigma}\left(\left\|\mathbf{\Lambda}\vec{x}+\vec{b}\right\|_2^2+o\right)\left(\mathbf{\Lambda}\vec{x}+\vec{b}\right)
        \label{Eqn:AffineWithIntrinsicLength}
    \end{equation}
    \begin{equation}
        \left\|\mathbf{\Lambda}\vec{x}+\vec{b}\right\|_2^2+o=\sum_{j\neq0}\left(\mathbf{\Lambda}_{jj}x_j+b_j\right)^2+\big(\underset{\rightarrow 0}{\underbrace{\mathbf{\Lambda}_{00}x_0}}+b_0\big)^2+o=\left\|\mathbf{\Lambda}'\vec{x}+\vec{b}'\right\|_2^2+\underset{=o'}{\underbrace{b_0^2+o}}
        \label{Eqn:IntrinsicSymmetry}
    \end{equation}
    This neurodegenerates the architecture, entering the functional class of narrower-width multilayer perceptrons, with the reparameterisation symmetry restored in the limit, and arbitrarily well approximated by chosen thresholding. Furthermore, to counteract the persistent growth of the intrinsic length, it can be made trainable. This may offer broader benefits, such as optimising the non-linearity's action or, if negative, allowing a non-linear sign change to representations, thereby permitting antipodal arrangements that may be desirable \cite{Elhage2022}. Numerous interpretations arise for it, such as a parameterised isotropic activation function preserving orthogonal equivariance, or a new affine translation, embedding the subspace of activations away from the origin, by a rotationally-degenerate bias-like vector, $\vec{o}$, in the orthogonal complement space, constrained by $\vec{o}^T\vec{o}=o$ and $\vec{o}\perp \mathbf{\Lambda}\vec{x}+\vec{b}$.

    Additionally, the remaining residual weights ($\mathbf{\Lambda}_{ii}=\epsilon$) in $\mathbf{f}$'s linear term can be better mitigated via the Moore-Penrose pseudo-inverse. Given diagonalised weights $\mathbf{\Lambda}$ and the subsequent pruned form $\mathbf{\Lambda}'$ alongside the next layer's weight matrices before $\mathbf{W}$ and after pruning $\mathbf{W}'$, then the following map should be closely preserved: $\mathbf{W}'\mathbf{\Lambda}'\mathbf{Q}^T\vec{x}\approx\mathbf{W}\mathbf{\Lambda}\mathbf{Q}^T\vec{x}$, by minimising the least-squares difference in these maps. $\mathbf{W}'$ can then be determined by the pseudo-inverse, where rank permits, shown in \textit{Eqn.}~\ref{Eqn:MoorePenrose}. Otherwise, continue using classically pruned $\mathbf{W}'$. If the threshold is particularly small, the inverse may be poorly conditioned; thus, this method necessitates care. Regardless, the standard smallest singular value deletion reduces this to simple corresponding column deletion of $\mathbf{W}'$.
    \begin{equation}
        \mathbf{W}'=\mathbf{W}\mathbf{\Lambda}\mathbf{\Lambda}^{'T}\left(\mathbf{\Lambda}'\mathbf{\Lambda}^{'T}\right)^{-1}
        \label{Eqn:MoorePenrose}
    \end{equation}
\subsubsection{Neurogenesis and Jacobians of Isotropic Activation Functions\label{Sec:Jacobians}}

    Concerns regarding the trainability of scaffold neurons are mitigated by analysing the Jacobian, $J_{\mathbf{f}}$, of isotropic activation functions and \textit{App.}~\ref{App:NullLike}'s discussion. Alongside standard connectivity mixing, isotropic Jacobians distribute learning gradients globally even in the case of an effective activation subspace. Hence, scaffold neurons may train rapidly, even if independent of the model's function and zero-valued associated parameters. The term proportional to the identity, $\mathrm{I}$, in \textit{Eqn.}~\ref{Eqn:IsotropicJacobian}'s Jacobian is generally non-zero, increasing gradient flow through scaffold neurons, hastening differentiation and operationalisation. However, canonical (Kronecker) basis vector directions $\hat{e}_i$ of scaffold neurons are zeroed in \textit{Eqn.}~\ref{Eqn:AnisotropicJacobian}'s Jacobian. With a suitable choice of $\tilde{\sigma}$, merely a coordinate singularity exists at $\vec{0}$. 
    \begin{align}
        \text{Isotropic:}\qquad\mathbf{f}\left(\vec{x}\right)&=
        \tilde{\sigma}\left(\left\|\vec{x}\right\|_2\right)\vec{x}
        &&\Rightarrow
        &&J_{\mathbf{f}} =
        \tilde{\sigma}\left(\left\|\vec{x}\right\|_2\right)\mathrm{I}+
        \frac{\tilde{\sigma}'\left(\left\|\vec{x}\right\|_2\right)}{\left\|\vec{x}\right\|_2}\vec{x}\vec{x}^T
        \label{Eqn:IsotropicJacobian}\\
        \text{Elementwise:}\qquad\mathbf{f}\left(\vec{x}\right)&=\sum_{i=1}^{n}\sigma\left(\vec{x}\cdot\hat{e}_i\right)\hat{e}_i
        &&\Rightarrow
        &&J_{\mathbf{f}} = \sum_{i=1}^{n}\sigma'\left(\vec{x}\cdot\hat{e}_i\right)\hat{e}_i\hat{e}_i^T
        \label{Eqn:AnisotropicJacobian}
    \end{align}

\subsubsection{Practical Considerations of Scheduling and Thresholding}

    Singular value decomposition itself features an $\mathcal{O}\left(m^2n\right)$ computational scaling dependence for $\mathbb{R}^{n\times m}$, $m<n$, matrices. Due to this cost, it is appropriate to consider only intermittent scheduling for neuroadaptation. This scheduling aligns with existing literature \cite{Evci2021}, but here it stems from the computational cost of the SVD rather than dense gradient computation. For straightforward implementation, this work considers layerwise decomposition only once per training epoch.  

    Although many methods may be feasible and compatible with the neuroadaptation described, as stated, we focus on the following distinct approach: maintaining a buffer of `scaffold neurons' in the diagonalised representation. The size of this buffer corresponds to a new hyperparameter, $\Xi\in\mathbb{Z}_+$, representing the desired number of singular values specified by a thresholding method. Because isotropic activation functions are not elementwise, different values of $\Xi$ are inequivalent, leading to tangible differences in training and inference. A learning-induced decrease in the number of neurons in the thresholded 'scaffold buffer' then triggers neurodegeneration, removing the smallest singular-value neuron with the least functional impact; an insufficient number in the buffer triggers neurogenesis---offering principled, real-time network adaptability. We note that many thresholding methods could indicate when neuroadaptation should occur; we compare a non-exhaustive list of methods in detail in \textit{App.}~\ref{App:Thresholding}. However, we propose and test a minimalistic approach using $0<\vartheta<1$ to demarcate when singular values have a negligible, or rarely contributing, effect. Singular values below 1 cause multiplicative decay in activations, suggesting their insignificance in a model's function. Minimalistic thresholding is attractive in the subsequent verification experiments, since its simplicity mitigates additional failure modes while performing well with contextually appropriate adaptations.



\subsubsection{Algorithmic Implementation}
    This section lists the exact transforms for forward neuroadaptation of an isotropic layer, for varying $m$: $(\mathbb{R}^n\!\rightarrow)\mathbb{R}^m\!\rightarrow\!\mathbb{R}^p$. For efficiency, several steps during and reparameterisation-\textit{contractions} after adaptation ($\mathbf{W}_1'\!=\!\mathbf{\Lambda}'\mathbf{Q}^T$) are done, meaningfully altering optimisation trajectories (\textit{App.}~\ref{App:GradientCoupling}).

    The chosen scheduler determines the timing of these actions, and they are repeated per layer specified. First, perform left-sided partial diagonalisation (\textit{Eqn.}~\ref{Eqn:LeftActingNew}) --- such diagonalisation need not be later undone, but $\mathbf{W}_1'\!=\!\mathbf{\Lambda}'\mathbf{Q}^T$ contraction is done. From the diagonal singular value matrix, $\mathbf{\Lambda}$, determine the number of scaffold neurons from singular values below the specified threshold $\mathbf{\Lambda}_{ii}<\vartheta$.
    \begin{equation}
        \mathbf{f}\left(\vec{x}\right)
        = \mathbf{W}_2{f}_{\text{Iso.}}\left(\mathbf{\Lambda}\mathbf{Q}^T\vec{x}+\vec{b}_1; o\right)+\vec{b}_2
        \label{Eqn:LeftActingNew}
    \end{equation}
    If the number of scaffold neurons falls below the desired count $\omega=\left|\left\{\mathbf{\Lambda}_{ii}|\mathbf{\Lambda}_{ii}<\vartheta\right\}\right|<\Xi$, iteratively enact neurogenesis for $(\Xi-\omega)$ neurons, described as follows:
    \begin{enumerate}
        \item Append a row of zeros to the bottom of $\mathbf{\Lambda}\in\mathbb{R}_{\geq 0}^{m \times n}$, to form a matrix $\mathbf{\Lambda}'\in\mathbb{R}_{\geq 0}^{\left(m+1\right) \times n}$.
    
        \item For $\vec{b}_1\in \mathbb{R}^m$ this is expanded to include a new last entry, $b_{+}$ to form $\vec{b}_1'\in \mathbb{R}^{\left(m+1\right)}$. For $o$-positivity, $b_{+}$ must be chosen such that $b_{+}^2+o'=o$, while $b_{+}=0$ is needed for $\vec{b}_2$'s step.
        
        \item Hence, with $o\in\mathbb{R}_{\geq 0}$, an adjustment follows accordingly from the prior step: $o'=o-b_{+}^2$.

        \item The weight matrix, $\mathbf{W}_2\in\mathbb{R}^{p\times m}$, acquires a new column, $\vec{W}_{+}$, forming $\mathbf{W}_2\in\mathbb{R}^{p\times \left(m+1\right)}$. The column $\vec{W}_{+}$ exhibits freedom in its initialisation only during the forward pass, due to zero premultiplication; however, it does couple with the optimisation discussed below.
        
        \item For vector, $\vec{b}_2\in\mathbb{R}^p$, no adjustment needs to be made, if $b_{+}=0$, due to identical impact.
    \end{enumerate}
    Initialising $\vec{W}_{+}$ as the (near-)zero vector can result in spontaneous symmetry breaking in the next optimisation step, whilst a non-zero vector acts as an explicit symmetry breaking, biasing learning accordingly, discussed in \textit{App.}~\ref{App:NullLike}. Hence, a range of options is available for empirical exploration: a linear combination of existing columns to achieve greater fidelity around the existing span, or various choices for semi-orthogonality, etc. Choices about symmetry-breaking initialisations may result in scaffold neurons diverging in specialisation, with their number having significant consequences.

    Iterative pruning occurs when the number of scaffold neurons exceeds the desired count, $\omega>\Xi$.
    \begin{enumerate}
        \item The row of $\mathbf{\Lambda}\in\mathbb{R}_{\geq 0}^{m \times n}$ with smallest norm is deleted, forming $\mathbf{\Lambda}'\in\mathbb{R}_{\geq 0}^{\left(m-1\right) \times n}$.
    
        \item Similarly, delete the corresponding entry,  $b_{\times}$, of $\vec{b}_1\in \mathbb{R}^m$ to form $\vec{b}_1'\in \mathbb{R}^{\left(m-1\right)}$.
        
        \item Adjust $o\in\mathbb{R}_{\geq 0}$ accordingly by $o'=o+b_{\times}^2$.

        \item Deletion of $\mathbf{W}_2\in\mathbb{R}^{p\times m}$ corresponding column, $\vec{W}_{\times}$, then yields $\mathbf{W}_2\in\mathbb{R}^{p\times \left(m-1\right)}$.
       
        \item For vector, $\vec{b}_2\in\mathbb{R}^p$, an adjustment arising from the prior bias term can be considered: $\vec{b}_2'=\vec{b}_2+\mathbb{E}\left[\bar{\sigma}\left(\cdot\right)\right]\vec{W}_{\times} b_{\times}$. This better approximates the functionality, where the expectation may be computed batch-wise or test-set-wise, or if isotropic normalisation ensures it is one.
    \end{enumerate}
    Better still, a further non-linear additive term and learnable parameter, $+\sigma\left(\cdot\right)\vec{\psi}$, can be used to absorb such terms identically, $\vec{\psi}'=\vec{\psi}+\vec{W}_{\times} b_{\times}$ --- essential is its regularised decay $\vec{\psi}\rightarrow\vec{0}$. This is for two reasons: it must be negligible before next-layer pruning, so it doesn't require further correction, and this also suggests an output-to-input ordering for neuroadaptation. Secondly, translation inconsistency due to the non-linear behaviour of $\psi$ may be troublesome, so it should decay over a short yet sufficient time for the network to compensate appropriately. Across both procedures, $\mathbf{Q}^T\in\mathbb{R}^{n \times n}$ need not be changed and can then be contracted with $\mathbf{\Lambda}'$ once concluded. Repeating this procedure over all specified layers then produces the automatic adaptive architecture. One can choose the final-layer representations, but alternating diagonalisation may help via the aforementioned sparsity benefit.

%

    
\section{Empirical Verifications of Theory and Mechanisms\label{Sec:Empirical}}
    Naturally, the present derivations apply to multilayer perceptrons; hence, the claims must be assessed in reference to various instances of this architecture. We use $\sigma(a)=\tanh a$ as a working placeholder, in analogy to standard tanh, but it is not assumed to be optimal\footnote{`Isotropic-tanh' also saturates, $\lim_{\alpha\rightarrow\infty}\sigma\left(\alpha\right)= 1$, which alleviates concerns regarding magnitude growth in activations compensating for small singular values. Thus, zero-tending singular values do suggest a negligible contribution to the function.}. Consequently, we empirically verify the theoretical claims about function invariance and neuroadaptation, rather than broader competitiveness due to function-optimality confounds. Beginning with \textit{Tab.}~\ref{Tab:CaltechPresentable} as the key demonstration of function preservation and invariance when neuroadapting a Caltech256 \cite{Griffin2007} MLP architecture, with invariances quantified through test-set mean relative change measured for the output before and after adaptation. 

    Table~\ref{Tab:CaltechPresentable} shows all variations for intrinsic length and $\psi$-correction worked, with the best invariances observed for constant or trainable intrinsic lengths, but varying for $\vec{\psi}$. Tiny relative changes ($\approx10^{-4}\%$) for neurogenesis and diagonalisation indicate negligible function alteration, confirming theoretical predictions and may arise from machine-error amplification in forward propagation. This rises modestly ($\approx3\%$) when removing the lowest $50\%$ of neurons, and increases to ($\approx 60\%$, width $100\!\rightarrow\!25$) when rank decreases severely and non-negligible singular values are removed. In practice, however, thresholded neuroadaptation should not remove such large singular values. \textit{App.}~\ref{App:ExperimentOne} supports this conclusion with extended results across KL-divergence, accuracy and relative-changes for all CIFAR-10, CIFAR-100 and Caltech256. Overall, this shows that functional degradation is mitigated despite \textit{no retraining} after adaptation---also explaining columnwise consistency. Training-set optimisation, at $19.0\pm0.2\mathrm{s}$, and test-set evaluation, at $2.04\pm0.01\mathrm{s}$, made the epoch-wise diagonalisation and neurogenesis, $29\pm1\mathrm{ms}$, or neurodegeneration, $39\pm2\mathrm{ms}$, comparatively negligible.
    
\begin{figure}[htb]
        \centering
        \includegraphics[width=\textwidth]{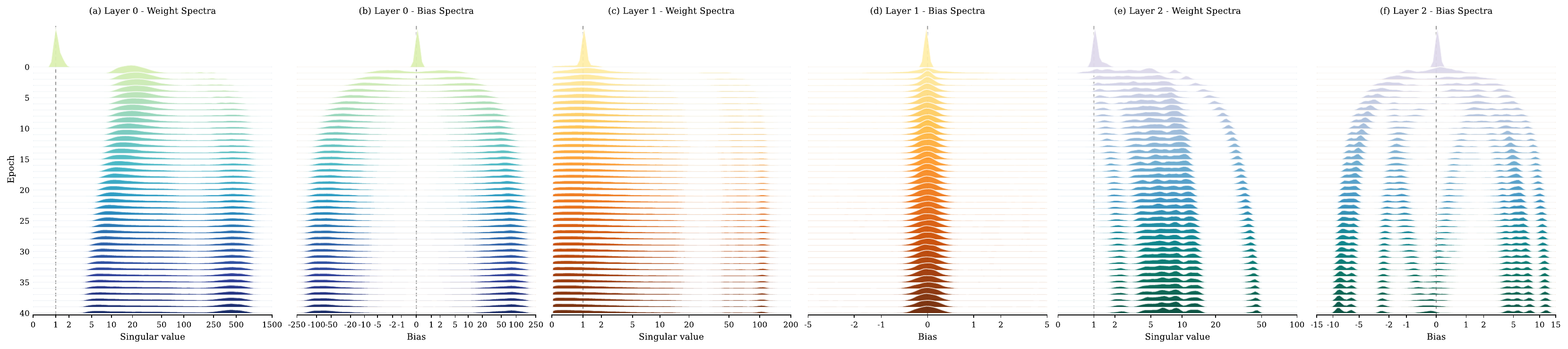}
        \caption{Experiment Two (\textit{App.}~\ref{App:ExperimentTwo}) displays the evolving layerwise singular value and bias spectra for a $\left[3072, 100, 100, 10\right]$ fixed isotropic MLP trained on CIFAR-10 \cite{Krizhevsky2009} classification. The spectra are obtained from $40$ independent repeats, showing strong consistency. Singular values begin at one, indicated by the dashed line, due to orthogonal initialisation, and biases at zero due to initialisation.}
        \label{Fig:IsotropicSpectra}
    \end{figure}

    \begin{table}[tb]
        \caption{Experiment One (\textit{App.}~\ref{App:ExperimentOne}): measures function invariance after neuroadaptation: appending $+k$, removing $-k$, or diagonalisation $0$ the $\mathbb{R}^{100}\rightarrow\mathbb{R}^{100}$ layer of a $[12288, 100, 100, 256]$ Caltech256 trained classifier. Mean $\pm$ standard error (SE) are reported over $20$ independent repeats. Columns specify intrinsic length $o$ and the ${\psi}$-vector approach, yielding $7\times20$ total networks. $\Sigma_0$ and $\Sigma_1$ denote the before- and after-sums of singular values, respectively (with their SE omitted for legibility; all $\approx0.4$). $\epsilon_{\mathrm{rel}}$ is the test-set mean $L_2$ logit change divided by the initial value, reported as a percentage.}
        \label{Tab:CaltechPresentable}
        \centering
        \scriptsize
        \setlength{\tabcolsep}{2.2pt}
        \renewcommand{\arraystretch}{1.12}
        \resizebox{\textwidth}{!}{%
        \begin{tabular}{r*{14}{c}}
        \toprule
        \text{Neuroadaptation} & \multicolumn{2}{c}{\shortstack{$o_{\mathrm{none}},\ \vec{\psi}_{\mathrm{decay}}$\\$\Sigma_0=240$}} & \multicolumn{2}{c}{\shortstack{$o_{\mathrm{constant}},\ \vec{\psi}_{\mathrm{decay}}$\\$\Sigma_0=240$}} & \multicolumn{2}{c}{\shortstack{$o_{\mathrm{trainable}},\ \vec{\psi}_{\mathrm{decay}}$\\$\Sigma_0=241$}} & \multicolumn{2}{c}{\shortstack{$o_{\mathrm{trainable}},\ \vec{\psi}_{\mathrm{none}}$\\$\Sigma_0=240$}} & \multicolumn{2}{c}{\shortstack{$o_{\mathrm{trainable}},\ \vec{\psi}_{\mathrm{constant}}$\\$\Sigma_0=237$}} & \multicolumn{2}{c}{\shortstack{$o_{\mathrm{trainable}},\ \vec{\psi}_{\mathrm{trainable}}$\\$\Sigma_0=237$}} & \multicolumn{2}{c}{\shortstack{$o_{\mathrm{trainable}},\ \vec{b}_{\mathrm{adjusted}}$\\$\Sigma_0=240$}} \\
        \cmidrule(lr){2-3}
        \cmidrule(lr){4-5}
        \cmidrule(lr){6-7}
        \cmidrule(lr){8-9}
        \cmidrule(lr){10-11}
        \cmidrule(lr){12-13}
        \cmidrule(lr){14-15}
         & $\Sigma_1$ & $\epsilon_{\mathrm{rel}}$ (\%) & $\Sigma_1$ & $\epsilon_{\mathrm{rel}}$ (\%) & $\Sigma_1$ & $\epsilon_{\mathrm{rel}}$ (\%) & $\Sigma_1$ & $\epsilon_{\mathrm{rel}}$ (\%) & $\Sigma_1$ & $\epsilon_{\mathrm{rel}}$ (\%) & $\Sigma_1$ & $\epsilon_{\mathrm{rel}}$ (\%) & $\Sigma_1$ & $\epsilon_{\mathrm{rel}}$ (\%) \\
        \midrule
        $+75$ & $240$ & $(8.4\!\pm\!0.2)\!\times\!10^{-4}$ & $240$ & $(8.5\!\pm\!0.2)\!\times\!10^{-4}$ & $241$ & $\mathbf{(8.2\!\pm\!0.2)\!\times\!10^{-4}}$ & $240$ & $(8.6\!\pm\!0.1)\!\times\!10^{-4}$ & $237$ & $(8.5\!\pm\!0.2)\!\times\!10^{-4}$ & $237$ & $(8.3\!\pm\!0.2)\!\times\!10^{-4}$ & $240$ & $(8.3\!\pm\!0.1)\!\times\!10^{-4}$ \\
        $+50$ & $240$ & $(8.4\!\pm\!0.2)\!\times\!10^{-4}$ & $240$ & $(8.5\!\pm\!0.2)\!\times\!10^{-4}$ & $241$ & $\mathbf{(8.2\!\pm\!0.2)\!\times\!10^{-4}}$ & $240$ & $(8.6\!\pm\!0.1)\!\times\!10^{-4}$ & $237$ & $(8.5\!\pm\!0.2)\!\times\!10^{-4}$ & $237$ & $(8.3\!\pm\!0.2)\!\times\!10^{-4}$ & $240$ & $(8.3\!\pm\!0.1)\!\times\!10^{-4}$ \\
        $+25$ & $240$ & $(8.4\!\pm\!0.2)\!\times\!10^{-4}$ & $240$ & $(8.5\!\pm\!0.2)\!\times\!10^{-4}$ & $241$ & $\mathbf{(8.2\!\pm\!0.2)\!\times\!10^{-4}}$ & $240$ & $(8.6\!\pm\!0.1)\!\times\!10^{-4}$ & $237$ & $(8.5\!\pm\!0.2)\!\times\!10^{-4}$ & $237$ & $(8.3\!\pm\!0.2)\!\times\!10^{-4}$ & $240$ & $(8.3\!\pm\!0.1)\!\times\!10^{-4}$ \\
        $+5$ & $240$ & $(8.4\!\pm\!0.2)\!\times\!10^{-4}$ & $240$ & $(8.5\!\pm\!0.2)\!\times\!10^{-4}$ & $241$ & $\mathbf{(8.2\!\pm\!0.2)\!\times\!10^{-4}}$ & $240$ & $(8.6\!\pm\!0.1)\!\times\!10^{-4}$ & $237$ & $(8.5\!\pm\!0.2)\!\times\!10^{-4}$ & $237$ & $(8.3\!\pm\!0.2)\!\times\!10^{-4}$ & $240$ & $(8.3\!\pm\!0.1)\!\times\!10^{-4}$ \\
        $+1$ & $240$ & $(8.4\!\pm\!0.2)\!\times\!10^{-4}$ & $240$ & $(8.5\!\pm\!0.2)\!\times\!10^{-4}$ & $241$ & $\mathbf{(8.2\!\pm\!0.2)\!\times\!10^{-4}}$ & $240$ & $(8.6\!\pm\!0.1)\!\times\!10^{-4}$ & $237$ & $(8.5\!\pm\!0.2)\!\times\!10^{-4}$ & $237$ & $(8.3\!\pm\!0.2)\!\times\!10^{-4}$ & $240$ & $(8.3\!\pm\!0.1)\!\times\!10^{-4}$ \\
        $0$ & $240$ & $(8.4\!\pm\!0.2)\!\times\!10^{-4}$ & $240$ & $(8.5\!\pm\!0.2)\!\times\!10^{-4}$ & $241$ & $\mathbf{(8.2\!\pm\!0.2)\!\times\!10^{-4}}$ & $240$ & $(8.6\!\pm\!0.1)\!\times\!10^{-4}$ & $237$ & $(8.5\!\pm\!0.2)\!\times\!10^{-4}$ & $237$ & $(8.3\!\pm\!0.2)\!\times\!10^{-4}$ & $240$ & $(8.3\!\pm\!0.1)\!\times\!10^{-4}$ \\
        $-1$ & $240$ & $(2.4\!\pm\!0.1)\!\times\!10^{-1}$ & $240$ & $(2.4\!\pm\!0.1)\!\times\!10^{-1}$ & $240$ & $(2.6\!\pm\!0.2)\!\times\!10^{-1}$ & $240$ & $\mathbf{(2.2\!\pm\!0.1)\!\times\!10^{-1}}$ & $237$ & $(2.7\!\pm\!0.1)\!\times\!10^{-1}$ & $236$ & $(2.6\!\pm\!0.1)\!\times\!10^{-1}$ & $240$ & $(2.3\!\pm\!0.2)\!\times\!10^{-1}$ \\
        $-5$ & $236$ & $(3.5\!\pm\!0.2)\!\times\!10^{-1}$ & $236$ & $(3.7\!\pm\!0.2)\!\times\!10^{-1}$ & $236$ & $(3.8\!\pm\!0.3)\!\times\!10^{-1}$ & $236$ & $\mathbf{(3.4\!\pm\!0.2)\!\times\!10^{-1}}$ & $233$ & $(6.7\!\pm\!0.2)\!\times\!10^{-1}$ & $233$ & $(6.4\!\pm\!0.2)\!\times\!10^{-1}$ & $236$ & $(3.7\!\pm\!0.3)\!\times\!10^{-1}$ \\
        $-25$ & $212$ & $(7.2\!\pm\!0.2)\!\times\!10^{-1}$ & $212$ & $(7.3\!\pm\!0.2)\!\times\!10^{-1}$ & $212$ & $(7.5\!\pm\!0.3)\!\times\!10^{-1}$ & $212$ & $(7.1\!\pm\!0.2)\!\times\!10^{-1}$ & $209$ & $(1.0\!\pm\!0.0)\!\times\!10^{0}$ & $209$ & $(9.9\!\pm\!0.2)\!\times\!10^{-1}$ & $212$ & $\mathbf{(7.1\!\pm\!0.3)\!\times\!10^{-1}}$ \\
        $-50$ & $170$ & $(3.7\!\pm\!0.1)\!\times\!10^{0}$ & $170$ & $(3.6\!\pm\!0.1)\!\times\!10^{0}$ & $170$ & $(3.9\!\pm\!0.2)\!\times\!10^{0}$ & $170$ & $\mathbf{(3.6\!\pm\!0.2)\!\times\!10^{0}}$ & $167$ & $(4.2\!\pm\!0.1)\!\times\!10^{0}$ & $167$ & $(4.1\!\pm\!0.2)\!\times\!10^{0}$ & $170$ & $(3.7\!\pm\!0.1)\!\times\!10^{0}$ \\
        $-75$ & $99$ & $(6.3\!\pm\!0.0)\!\times\!10^{+1}$ & $99$ & $(6.3\!\pm\!0.0)\!\times\!10^{+1}$ & $99$ & $(6.2\!\pm\!0.1)\!\times\!10^{+1}$ & $99$ & $(6.3\!\pm\!0.0)\!\times\!10^{+1}$ & $98$ & $(6.3\!\pm\!0.0)\!\times\!10^{+1}$ & $97$ & $\mathbf{(6.2\!\pm\!0.0)\!\times\!10^{+1}}$ & $99$ & $(6.3\!\pm\!0.0)\!\times\!10^{+1}$ \\
        \bottomrule
        \end{tabular}%
        }
    \end{table}
    \textit{Fig.}~\ref{Fig:IsotropicSpectra} shows the collated evolution of isotropic parameters, which evolve very similarly despite random (orthogonal) initialisation, perhaps because the symmetry induces a flat loss curve across the group's continuous orbit. These differ strongly from elementwise forms (\textit{App.}~\ref{App:ExperimentTwo}), showing their incomparability despite superficial similarities. Layers $0$ and $2$ show strong specialisation, with distinct separating bands---decay of small singular value tails suggests utility in their neurodegeneration.
    \begin{figure}[htb]
        \centering
        \includegraphics[width=\textwidth]{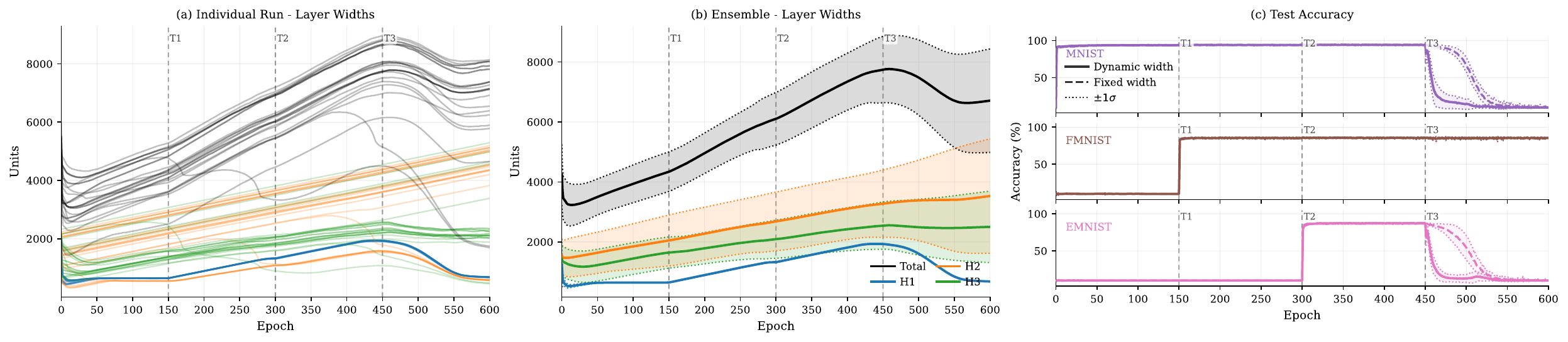}
        \caption{Experiment Three (\textit{App.}~\ref{App:ExperimentThree}) demonstrates thresholded neural adaptation when an architecture of $\left[2352, \text{H1}, \text{H2}, \text{H3}, 30\right]$ adapts to a changing task schedule, across $25$ independent repeats for random hidden widths $\text{H1}, \text{H2}, \text{H3}\!\sim\!\mathrm{U}\left(500, 2500\right)$. From epoch $0$ through $150$, only MNIST \cite{LeCun1998} cross-entropy terms contribute to the loss; at $\text{T1}$, the total loss is equally averaged between both MNIST and FMNIST \cite{Xiao2017}. At $\text{T2}$, the total loss is equally averaged between MNIST, FMNIST, and A-J letters from EMNIST \cite{Cohen2017}, and finally, at $\text{T3}$, \textit{only} FMNIST's loss contributes.}
        \label{Fig:DifferentDatasets}
    \end{figure}

    Measuring neuroadaptation under a changing task schedule is essential, with \textit{Fig.}~\ref{Fig:DifferentDatasets} demonstrating how added learning pressures tend to cause a bounded neurogenesis until stabilising at a new width, whilst removing datasets causes neurodegeneration of unneeded excess to a new stable width --- reducing EMNIST and MNIST test accuracies correspondingly to demand. This indicates \textit{Fig.}~\ref{Fig:IsotropicSpectra}'s intended decay of low-singular-value bands is robust (if desirable one could preserve task-corresponding singular values \cite{Kirkpatrick2017}). Overall, it demonstrates clearly that thresholded scaffold neurons, and their buffer, operate as intended. Adaptive networks have matched or improved accuracy over static baselines\footnote{At T3, epoch 450, the (control)/(adaptive) mean and standard error test accuracy, was recorded at $(94.2\pm0.1)\%$/$(94.2\pm0.1)\%$ for MNIST, $(85.6\pm0.2)\%$/$(85.4\pm0.1)\%$ for FMNIST and $(87.3\pm0.2)\%$/$(87.3\pm0.2)\%$ for A-J EMNIST}. The first layer, H1, shows remarkable consistency across repeats at all times, suggestive of an optimal width. Curiously, distinct width groupings appear and bifurcate in H2, and clearly for H3, beginning from the start to the T3 boundary. The linearly increasing H2 group appears metastable, often converging rapidly to a stable, narrower width. We speculate this reflects function consolidation, in which inter-task learning transfer merges separately acquired task-specific behaviours into a shared approach that supports multiple tasks---reducing redundancy in both function and neurons. Adding datasets appears to trigger this, perhaps by bridging intermediate internal representations, strengthened by its occurrence in deeper layers, suggestive of redundancy exploited in more generalised representations shareable across tasks. Individual runs indicate bifurcation may involve layer couplings, typically splitting between standard contracting $H1\!\geq\!H2\!\geq\!H3$, which stabilises, and $H2\!\geq\!H3\!\geq\!H1$, which grow. Work identified subsequently on QR-compression \cite{Ganev2022} may force width contraction but alters learning. Overall, adaptation remains effective and responsive to task demands: growing and shrinking with task addition and removal.


\section{Conclusion}


    We develop a symmetry-led approach to adaptive topologies. Crucially, through this we demonstrate the utility of prescribed primitive-scale symmetries in their own right, which need not correspond to symmetries of the data and model, but instead are used solely within internal representations to enable otherwise inaccessible structural transformations. Activation functions are made to exhibit orthogonal equivariance that permits function-invariant diagonalisation and neuroadaptation. Verification across various datasets confirms negligible functional degradation under substantial neurogenesis, while mitigating drift under neurodegeneration. 
    Together, this suggests the taxonomy and reversed ontology underlying the method offer novel behaviours and opportunities: from neuroadaptation to sparsification and ICA-interpretable representations, this sweep offers concrete examples of unusual capabilities enabled through symmetry-prescribed primitives. Moreover, the reversal reveals neuron individuation can be an emergent consequence rather than a fundamental assumption. This motivates a broader exploration of prescribed primitives and the expanded functional class of neural systems they permit for general applications.
    Appendices cover optimiser and normaliser considerations, sparsification verification, nested functional classes, ICA representations, a sketch of convolution and transformer neuroadaptation and the \textit{parallelised precomputation of all matrix-vector products}. 

    \textbf{Scope and Limitations.} This approach to neuroadaptation has a defined scope, as expanded in \textit{App.}~\ref{App:Limitations}. The empirics focus on mechanistic verification of the approach. This is for several reasons, including that the methodology depends on isotropic primitives, which have not benefited from the longstanding refinement of modern elementwise functions. This offers a substantial avenue for further exploration. Additionally, isotropic forms still lack a general UAT \cite{Hornik1989, Cybenko1989}; yet the intervening decades between \citet{Rosenblatt1958} and elementwise-UATs did not stall this form's practical adoption. Separately, the primary methodology centres on MLP architectures for clarity. Accordingly, we prioritise mechanistic verification over competitive SOTA performance. Further implementation considerations include explicit symmetry breaking during neurogenesis while retaining function-invariance.

    \begin{ack}\end{ack}
    Work supported by the Engineering and Physical Sciences Research Council (EPSRC), grant number [EP/W524347/1]. I thank Dr Omar Rivasplata and Dr Mohammad Saeed Bahramy for their continued supervision and support throughout my PhD.


\bibliographystyle{unsrtnat}
\bibliography{references.bib}
\newpage
\appendix
\section{Considering Gradient Non-Invariance\label{App:GradientCoupling}}

    Despite the forward-pass invariance of various reparameterisation approaches, some reparameterisations diverge in their optimisation trajectories under these transforms. This section will overview these considerations.

\subsection{Transforming Optimiser States}

    Practically, the core consideration is that of the various optimisers' running statistics. These include the momentum of momentum gradient descent \cite{Sutskever2013}, the first- and second-order moment moving-average estimates of the ADAM \cite{Kingma2017} and ADAM-W \cite{Loshchilov2019} optimisers, among many others. Several can be exactly transformed under a basis change to ensure invariance; however, some must be approximated. This requires consideration of each mathematical formulation and its motivation to determine which tensor rank the state should transform under. This section will consider derivations of the momentum and ADAM state transforms.

    For simplicity, consider a vector parameter, $\vec{a}$, being updated by an optimiser of form $\vec{a}_{n+1}=\vec{a}_n-\eta\vec{U}\left(\vec{a}_n\right)$. Assume the network of interest has a construction such that it exhibits forward-pass invariance to orthogonal transforms on the parameter $\vec{a}$, denoted: $\vec{a}'=\mathbf{R}\vec{a}$. For the isotropic MLP, this transform is compensated for elsewhere, such as in the next and previous layers; for example, the diagonalisation of \textit{Sec.}~\ref{Sec:Diagonalisation}. Due to the designed function invariance, the loss is similarly invariant, $\mathcal{L}'\left(\vec{a}'\right)=\mathcal{L}\left(\vec{a}\right)$ as long as any additional terms are assumed also invariant, such as regularisers, e.g. $\mathcal{L}_2\propto \vec{a}^T\vec{a}$. Naturally, we may also want the optimisation trajectories to be invariant under basis changes, as this acts like a new coordinate system and representation for the matrices.

    Therefore, to ensure continued invariance under this reparameterisation, $\mathcal{L}'\left(\vec{a}'_n\right)=\mathcal{L}\left(\vec{a}_n\right)$, for all $n$, requires the trajectories to be equivariant to the reparameterisation transform. This is described in \textit{Eqn.}~\ref{Eqn:OptimiserTransform}, and effectively uses \textit{Eqn.}~\ref{Eqn:Algebraic} applied to a general optimiser.
    
    \begin{equation}
        \vec{a}_{n+1}'=
        \left(\vec{a}_{n}'-\eta \vec{U}'\left(\vec{a}'_n\right)\right)=
        \left(\mathbf{R}\vec{a}_{n}-\eta\mathbf{R}\vec{U
        }\left(\mathbf{R}^T\vec{a}'_n\right)\right)=
        \mathbf{R}\vec{a}_{n+1}
        \quad\Rightarrow\quad \vec{U}'\left(\vec{a}'\right)=\mathbf{R}\vec{U}\left(\mathbf{R}^T\vec{a}'\right)=\mathbf{R}\vec{U}\left(\vec{a}\right)
        \label{Eqn:OptimiserTransform}
    \end{equation}
    
    Hence, network and loss invariance under an orthogonal transform motivates an optimiser algebraic-equivariance condition on the optimiser dynamics under reparameterisation
    $\vec{U}'\left(\vec{a}'\right)=\mathbf{R}\vec{U}\left(\vec{a}\right)$. This is an intertwining relation between the original and transformed optimiser states, that reduces to a standard algebraic equivariance when $\vec{U}'=\vec{U}$. Standard gradient descent satisfies this trivially by the chain rule.

    \begin{equation}
    \vec{U}'\left(\vec{a}'\right)\equiv\nabla_{\vec{a}'}\mathcal{L}'\left(\vec{a}'\right)=\left(\frac{\partial\vec{a}'}{\partial \vec{a}}\right)^{-T}\nabla_{\vec{a}}\mathcal{L}\left(\vec{a}\right)=\mathbf{R}\nabla_{\vec{a}}\mathcal{L}\left(\vec{a}\right)=\mathbf{R}\vec{U}\left(\vec{a}\right)
        \label{Eqn:GDTransform2}
    \end{equation}

    Overall, this can be understood intuitively: if $\nabla_{\varrho}\mathcal{L}$ is locally the direction of steepest descent in parameter space, for parameter $\varrho$, then transforming to a new orthonormal basis in parameter space, $\varrho'=\mathbf{R}\varrho$, does not change the vector of steepest descent, only its representation in the chosen basis --- the underlying loss surface is invariant due to function invariance. 

    However, this must be reevaluated for momentum, $\vec{U}_{\text{mome.}}$, and ADAM optimisers, $\vec{U}_{\text{ADAM}}$. For momentum, with factor $\alpha\in\left[0, 1\right]$, this requires a transform to the update vector as $\vec{M}_n'=\mathbf{R}\vec{M}_n$. Hence, this transforms the same as the underlying parameter, $\vec{a}$, as may be expected.

    \begin{align}
        \vec{U}_{\text{mome.}} = \vec{M}_n\quad,\quad \vec{M}_{n+1}&=\alpha\vec{M}_{n}+\left(1-\alpha\right)\nabla_{\vec{a}}\mathcal{L}\left(\vec{a}\right)\\
        \text{requiring:}\quad\vec{M}_{n+1}'&=\mathbf{R}\vec{M}_{n+1}=\alpha\mathbf{R}\vec{M}_{n}+\left(1-\alpha\right)\underset{\nabla_{\vec{a}'}\mathcal{L}'\left(\vec{a}'\right)}{\underbrace{\mathbf{R}\nabla_{\vec{a}}\mathcal{L}\left(\vec{a}\right)}}
        \label{Eqn:MomentumTransform}
    \end{align}

    For ADAM, the first moment transforms, $\vec{M}$ identically, $\vec{M}_n'=\mathbf{R}\vec{M}_{n}$; however, the second moment itself is a diagonal approximation of a true second moment due to elementwise approximations. \textit{Eqn.}~\ref{Eqn:ADAM} displays ADAM in elementwise form, with the $\varepsilon$-positivity term neglected for simplicity. 

    \begin{equation}
        \vec{U}_{\text{ADAM}}= \sum_{i}\frac{\vec{M}_n\cdot\hat{e}_i}{\sqrt{\vec{V}_n\cdot\hat{e}_i}}\hat{e}_i\qquad\qquad
        \left\{\begin{matrix}
        &\vec{M}_{n+1}
        = \beta_1\vec{M}_{n}
        + \left(1-\beta_1\right)\nabla_{\vec{a}}\mathcal{L}\left(\vec{a}\right)
        \\
        &\vec{V}_{n+1}
        = \beta_2\vec{V}_{n}
        + \left(1-\beta_2\right)
        \left(\nabla_{\vec{a}}\mathcal{L}\left(\vec{a}\right)\odot
        \nabla_{\vec{a}}\mathcal{L}\left(\vec{a}\right)\right)
        \end{matrix}\right.
        \label{Eqn:ADAM}
    \end{equation}

    This product of gradients term in $\vec{V}_n$ behaves as the diagonal of a rank-2 tensor quantity in terms of transform: $\tilde{V}=\left(\nabla_{\vec{a}}\mathcal{L}\left(\vec{a}\right)\odot\nabla_{\vec{a}}\mathcal{L}\left(\vec{a}\right)\right)=\operatorname{diag}\left(\nabla_{\vec{a}}\mathcal{L}\left(\vec{a}\right)\left(\nabla_{\vec{a}}\mathcal{L}\left(\vec{a}\right)\right)^T\right)$, considering that it is a projection of the outer product of two rank-1 tensors. Thus, the elementwise squaring of the gradient induces basis dependence by removing off-diagonal information, which is irrecoverable. Hence it is proposed to transform it as a rank-2 matrix but preserve the off-diagonal zero projection, as shown in \textit{Eqn.}~\ref{Eqn:DiagonalApproximation}, where $\mathbf{R}^{\odot2}=\mathbf{R}\odot\mathbf{R}$ is the an elementwise square of the matrix, while $\operatorname{diag}:\mathbb{R}^{n\times n}\rightarrow\mathbb{R}^n$ is the canonical diagonal function which extracts the matrix diagonal into a vector, and the reverse process $\operatorname{diag}^{\dagger}:\mathbb{R}^{n}\rightarrow\mathbb{R}^{n\times n}$ which embeds a vector along a matrix diagonal.

    \begin{equation}
        \tilde{V}'=\operatorname{diag}\left(\mathbf{R}\left(\operatorname{diag}^{\dagger}\tilde{V}\right)\mathbf{R}^T\right)=\mathbf{R}^{\odot2}\tilde{V}=\left(\mathbf{R}\odot\mathbf{R}\right)\tilde{V}
        \label{Eqn:DiagonalApproximation}
    \end{equation}

    For hyperoctahedral-prescribed networks, including the more typical permutation-prescribed networks, all these optimiser transforms are exact and consistent with a (signed) relabelling. This includes ADAM's second moment. Hence, typically, networks do not exhibit a distinction between the forward pass reparameterisation symmetry contrasted with a backward pass broken reparameterisation symmetry; their optimisation trajectories remain equivariant and exhibit a training-time function invariance, $\vec{a}'_{n}=\mathbf{P}\vec{a}_n$, for all $n$ and constant permutation matrix $\mathbf{P}$. Yet this is not the case for orthogonal transforms; the second-moment transform is only approximate. The network is symmetric under orthogonal reparameterisations during the forward pass, but it couples with certain optimisers in a non-invariant manner, such that reparameterisations yield different optimisation trajectories.

    Nevertheless, one should still apply the approximate transform to ADAM's second moment to avoid inappropriate scaling of the effective loss surface, and it is proposed that this approximated rank-2 transform is more consistent with the quadratic gradient term than treating it as a rank-1 quantity  $\vec{V}'=\mathbf{R}\vec{V}$. Hence, an inference-time reparameterisation orthogonal symmetry exists, but concurrently, a training-time equivalent symmetry does not generally exist. 

    Briefly, Muon \cite{Jordan2024} with its natural orthogonal construction appears impervious to such orthogonal basis changes. It takes a first-order moment $\mathbf{M}\in\mathbb{R}^{n\times m}$ on the weights and adapts it via Newton–Schulz. approximately equivalent to setting the moment's singular values to one via SVD. Hence, as long as this first-order moment transforms appropriately under our orthogonal basis changes (analogous to the weights' own transformation), this operator should remain unaffected in the trajectories it produces.

    Overall, the practical takeaway is that, provided an optimiser is equivariant under the symmetry and the function and loss are invariant under it, an appropriate reparameterisation of the model and optimiser yields an equivariant trajectory.
    
\subsection{Tensor Contraction Inequivalence}

    A separate, classical route in which reparameterisation can result in function invariance but optimiser non-invariance is considering the presence and absence of various tensor contractions, such as whether the SVD decomposition preserves $\mathbf{\Lambda}\mathbf{Q}^T$ as separate or contracted $\mathbf{W}'=\mathbf{\Lambda}\mathbf{Q}^T$. This consideration is more broadly pertinent, including implications for the normaliser's constrained-affine parameterised maps. Generally, this arises when optimisation is performed on the parameter space, with these non-contracted maps increasing dimensionality, thereby altering the loss surface and steepest descent, resulting in different optimiser trajectories.
    
    Considering the map $\vec{y}_m=\mathbf{W}_{mn}\vec{x}_n$ with decomposition $\mathbf{A}\mathbf{B}=\mathbf{W}$, one can see two different functions after a gradient step when assuming standard gradient descent and notation $\vec{g}=\nabla_{\varrho} \mathcal{L}$ for appropriate parameter $\varrho$.

    \begin{align}
        \vec{y}_m'&=\left(\mathbf{W}_{mn}-\eta \vec{g}_m\vec{x}_n\right)\vec{x}_n\qquad\Rightarrow\qquad \Delta\vec{y}=\vec{y}'-\vec{y}=-\eta\left\|\vec{x}\right\|_2^2\vec{g}
        \label{Eqn:ContractedForm}
    \end{align}
    \begin{align}
        \vec{y}_m'&=\left(\mathbf{A}_{mk}-\eta \vec{g}_m \mathbf{B}_{ki}\vec{x}_i\right)
       \left(\mathbf{B}_{kn}-\eta \vec{g}_j \mathbf{A}_{jk}\vec{x}_n\right)\vec{x}_n\\&\Rightarrow\qquad\Delta\vec{y}=\vec{y}'-\vec{y}=-\eta\left(\vec{g}\left\|\mathbf{B}\vec{x}\right\|_2^2+\left\|\vec{x}\right\|_2^2\mathbf{A}\mathbf{A}^T\vec{g}\right)+\eta^2\left\|\vec{x}\right\|_2^2 \left(\vec{x}^T\mathbf{B}^T\mathbf{A}^T\vec{g}\right)\vec{g}
        \label{Eqn:UnContractedForm}
    \end{align}
        
    This result shows that leaving a layer in a \textit{decomposed} diagonalised parameterisation changes optimisation. 

    The approach used in this paper is to decompose the specified weight matrix into three temporary matrices every epoch, take appropriate neuroadaptive/sparsification actions, and then immediately recombine the temporaries into the appropriate contracted matrices before continuing training. Consequently, one does not optimise over $\mathbf{R}$, $\mathbf{Q}$ and $\mathbf{\Lambda}$ but over $\mathbf{W}_i$ and $\mathbf{W}_i'$ instead for before and after the neuroadaptive actions respectively. We do this contraction because optimisation will likely change the subsequent SVD rotation factors and thus require contraction and re-decomposition anyway, while also improving efficiency by reducing the trainable parameter count. Thus, $\mathbf{R}$, $\mathbf{Q}$ and $\mathbf{\Lambda}$ never become persistent trainable parameters in this setup.
    

\subsection{Null-Like Reparameterisations\label{App:NullLike}}

    Of greater significance for this paper are the null-like reparameterisations that arise when the singular values of a partially or fully diagonalised layer tend to zero, or are created at zero. Many such reparameterisations leave the network functionally identical; yet couple to the optimisation in non-trivial ways. This introduces a mode to steer the training of such networks by explicitly breaking symmetry through the chosen reparameterisations/initialisation. Considering the map of \textit{Eqn.}~\ref{Eqn:Diagonal2MLP}, representing a two-layer MLP with a diagonalised first layer and isotropic activation $\mathbf{f}\left(\vec{x}\right)=\bar{\sigma}\left(\vec{x}^T\vec{x}\right)\vec{x}$.

    \begin{align}
        \vec{y}_i=\mathbf{W}_{ij}\left[\mathbf{f}\left(\vec{z}\right)\right]_j+\vec{b}^{(2)}_i && \vec{z}_j=\vec{\Lambda}_j\vec{x}_j+\vec{b}^{(1)}_j
        \label{Eqn:Diagonal2MLP}
    \end{align}

    Computing the gradients of $\mathbf{W}$ and $\vec{\Lambda}$ yields \textit{Eqn.}~\ref{Eqn:Grads}.

    \begin{align}
        \frac{\partial \mathcal{L}}{\partial \mathbf{W}_{pq}} = \vec{g}_p\left[\mathbf{f}\left(\vec{z}\right)\right]_q && \frac{\partial \mathcal{L}}{\partial \vec{\Lambda}_{p}} = \vec{g}_i\mathbf{W}_{ij}\mathbf{J}_{jp}^{\mathbf{f}}\vec{x}_p && \frac{\partial \mathcal{L}}{\partial \vec{b}^{(1)}_{p}} = \vec{g}_i\mathbf{W}_{ij}\mathbf{J}_{jp}^{\mathbf{f}}
        \label{Eqn:Grads}
    \end{align}
    \begin{equation*}
        \mathbf{J}_{jp}^{\mathbf{f}}\left(\vec{z}\right)=\bar{\sigma}\left(\vec{z}^T\vec{z}\right)\delta_{jp}+2\bar{\sigma}'\left(\vec{z}^T\vec{z}\right)\vec{z}_j\vec{z}_p
    \end{equation*}

    These null-like reparameterisations emerge as a singular value tends to zero, $\vec{\Lambda}_0\rightarrow 0$ with bias $\vec{b}^{(1)}_0\rightarrow 0$. Then as a consequence $\vec{z}_0\rightarrow 0$ and $\left[\mathbf{f}\left(\vec{z}\right)\right]_0\rightarrow 0$ and from this a function-invariant unconstrained reparameterisation of $\mathbf{W}_{i0}$ is admitted.
    
    By numerical substitution into \textit{Eqn.}~\ref{Eqn:Grads}, one can determine that such a reparameterisation not only couples with the optimiser but produces a zero-valued fixed point for both $\mathbf{W}_{i0}$ and $\vec{\Lambda}_0$ --- stalling training. Hence, careful consideration of such null-like reparameterisations is essential, especially for neurogenesis initialisations, and an explicit symmetry breaking of $\mathbf{W}_{i0}\in\mathbb{R}^n\setminus\{\vec{0}\}$ is necessary. However, any non-zero choice biases the optimisation trajectory accordingly. Hence, a `no-free-lunch' result arises; one must bias the behaviour of the network's new neuron through a choice of $\vec{W}^{+}\neq\vec{0}$ or $b^{+}\neq 0$. To name several approaches, one may choose $\vec{W}^{+}\neq\vec{0}$ randomly, or a vector in the existing column-vector span of $\mathbf{W}$, or create a signed-bias in $b^{\star}$ or take it to equal the current intrinsic length, $b^{\star}=\pm\sqrt{o}$ --- many valid initialisations can help avoid this vanishing gradient problem, all with optimisation consequences.


\section{Considerations of Normalisation\label{App:Normalisers}}

    Within isotropic multilayer perceptrons, one may wish to implement a form of normalisation, especially post-composed with the activation function. Isotropic normalisers then mitigate activations growing and compensating for small singular values, which would affect the negligible impact assumption of $\mathbf{\Lambda}_{ii}\rightarrow0$. Isotropic-tanh, $\sigma\left(a\right)=\tanh a$, similarly mitigates this by saturating to 1. However, BatchNorm \cite{Ioffe2015}, LayerNorm \cite{Ba2016}, among others, do not exhibit standard orthogonal algebraic (\textit{Eqn.}~\ref{Eqn:Algebraic}) equivariance re-individuating neurons despite isotropic activation functions. RMSNorm \cite{Zhang2019} does display this orthogonal equivariance, but pre- and post-composition with an isotropic activation presents difficulties because the vector magnitude is constant.

    This is displayed in \textit{Eqn.}~\ref{Eqn:PreComposition} and \textit{Eqn.}~\ref{Eqn:PostComposition}, for pre- and post-composition of $L_2$-Norm (scaled RMSnorm), denoted as $\mathbf{N}\left(\vec{x}\right)=\vec{x}/\left\|\vec{x}\right\|_2$ (neglecting $\varepsilon$) and generic isotropic activation function, $\mathbf{f}=\bar{\sigma}\left(\vec{x}^T\vec{x}\right)\vec{x}$, for $\bar{\sigma}:\mathbb{R}^+\rightarrow\mathbb{R}^+$ and constant $C\in\mathbb{R}$. This section assumes parameterless normalisers with a generally neglected, negligible $\varepsilon$ for notational simplicity.

    \begin{equation}
        \mathbf{f}\left(\mathbf{N}\left(\vec{x}\right)\right)=\bar{\sigma}\left(\hat{x}^T\hat{x}\right)\hat{x}=\bar{\sigma}\left(1\right)\hat{x} = C\hat{x}=C\mathbf{N}\left(\vec{x}\right)
        \label{Eqn:PreComposition}
    \end{equation}

    \begin{equation}
        \mathbf{N}\left(\mathbf{f}\left(\vec{x}\right)\right)=\frac{\bar{\sigma}\left(\vec{x}^T\vec{x}\right)\vec{x}}{\left\|\bar{\sigma}\left(\vec{x}^T\vec{x}\right)\vec{x}\right\|_2}=\frac{\bar{\sigma}\left(\vec{x}^T\vec{x}\right)\vec{x}}{|\bar{\sigma}\left(\vec{x}^T\vec{x}\right)|\left\|\vec{x}\right\|_2}=\frac{\vec{x}}{\left\|\vec{x}\right\|_2}=\hat{x}=\mathbf{N}\left(\vec{x}\right)
        \label{Eqn:PostComposition}
    \end{equation}

    Consequently, $\mathbf{N}\circ\mathbf{f}=\mathbf{N}$, $\mathbf{f}\circ\mathbf{N}=C\mathbf{N}$, alongside $\mathbf{N}$'s idempotence $\mathbf{N}\circ\mathbf{N}=\mathbf{N}$. Thus, when implementing normalisation with adaptive isotropic networks, we encourage alternative radial normalisation, such as reintroducing batch statistics. This is described in \textit{Eqn.}~\ref{Eqn:BatchRMSNorm}, which acts like RMSNorm/$L_2$-Norm using a batched estimate of the mean norm, $\mathbb{E}_{j\in B}\left[\left\|\vec{x}_j\right\|_2\right]$ and samples $\vec{x}_i$. 

    \begin{equation}
        N_{b}\left(\left\{\vec{x}_i\right\}\right)=\frac{\vec{x}_i}{\mathbb{E}_{j\in B}\left[\left\|\vec{x}_j\right\|_2\right]}
        \label{Eqn:BatchRMSNorm}
    \end{equation}

    Similarly, `FrobNorm' over a matrix of batched samples $\mathbf{X}\in\mathbb{R}^{n\times b}$, using a Frobenius norm $\left\|\cdot\right\|_F$, could be used to similar effect, shown in \textit{Eqn.}~\ref{Eqn:FrobNorm}, with a specific instance for RMSNorm comparability.

    \begin{equation}
        N_{f}\left(\mathbf{X}\right)\propto\frac{\mathbf{X}}{\left\|\mathbf{X}\right\|_F}\qquad\Big(\text{e.g.} =\frac{\mathbf{X}}{(b\sqrt{n})^{-1}\left\|\mathbf{X}\right\|_F}\Big)
        \label{Eqn:FrobNorm}
    \end{equation}

    LayerNorm, $\mathrm{LN}:\mathbb{R}^n\rightarrow\mathbb{R}^n$, does exhibit an equivariance to $\mathrm{O}\left(n-1\right)$, with a $\varrho:g\rightarrow\mathrm{GL}_n\left(\mathbb{R}\right)$ representation which leaves $\vec{1}$ fixed: $\left\{\varrho'\left(g\right)|g\in\mathrm{O}\left(n\right), \varrho'\left(g\right)'\vec{1}=\vec{1}\right\}\cong\left\{\varrho\left(g\right)|g\in\mathrm{O}\left(n-1\right)\right\}$. In general, this ordinarily prohibits the usage of SVD diagonalisation, which necessitates a full $\mathrm{O}\left(n\right)$ equivariance; however if one generalises the form of LayerNorm to parameterise mean as instead a weighted-mean, then a joint equivariance to $\mathrm{O}\left(n\right)$ is restored through a direct-sum. This generalised form is displayed in \textit{Eqn.}~\ref{Eqn:OrthogonalLN}, $\mathrm{LN}_{\hat{n}}:\mathbb{R}^n\times\mathbb{R}^n\rightarrow\mathbb{R}^n$, with equivariance displayed in \textit{Eqn.}~\ref{Eqn:LNEquivariance} for standard representation, notated in shorthand as $\forall\mathbf{R}\in\mathrm{O}\left(n\right)$. A further \textit{Eqn.}~\ref{Eqn:Algebraic} prescription for normalisers.

    \begin{equation}
        \mathrm{LN}_{\hat{n}}\left(\vec{x};\hat{n}\right)=\frac{\vec{x}-\hat{n}\hat{n}^T\vec{x}}{\left\|\vec{x}-\hat{n}\hat{n}^T\vec{x}\right\|_2}
        \label{Eqn:OrthogonalLN}
    \end{equation}

    \begin{equation}
        \mathrm{LN}_{\mathbf{R}\hat{n}}\left(\mathbf{R}\vec{x};\mathbf{R}\hat{n}\right)=\mathbf{R}\mathrm{LN}_{\hat{n}}\left(\vec{x};\hat{n}\right)
        \label{Eqn:LNEquivariance}
    \end{equation}

    Thus, considering \textit{Eqn.}~\ref{Eqn:LNEquivariance}, one may use LayerNorm in the neuroadaptive isotropic multilayer perceptron, since the weighted mean transforms as \textit{Eqn.}~\ref{Eqn:LNEquivariance} under an orthogonal basis change.

    Similar generalisations may be considered to reinterpret parameterless BatchNorm, which typically exhibits standard hyperoctahedral algebraic equivariance. In this formula, each vector element is divided by its standard deviation across the batch. This element-wise approximation to full covariance-whitening induces hyperoctahedral equivariance --- analogous to the second-moment elementwise approximation of the ADAM optimiser \cite{Kingma2017}. Typical composed parameterisations also break this symmetry down to a closure form of the symmetry.
    
    However, surrounding weights are closed under orthogonal transformations too, separate from the elementwise activations. Hence, training may naturally explore rotated representations. Consequently, the reference basis used to compute diagonal (elementwise) variance may shift during training. Thus, one may view BatchNorm as a batch estimate of the diagonal of the full covariance matrix. The full covariance matrix transforms as a rank-2 tensor $\mathbf{C}'=\mathbf{R}\mathbf{C}\mathbf{R}^T$, with its diagonal approximately transforming as $\vec{C}'=\left(\mathbf{R}\odot\mathbf{R}\right)\vec{C}$ for Hadamard product $\odot$, as previously derived in \textit{Eqn.}~\ref{Eqn:DiagonalApproximation}. These approximate transformations may be reasonable for three reasons: first, training may induce comparable rotations of the representation regardless; second, BatchNorm already relies on noisy batch-level estimates, so the approximate transformation may not introduce qualitatively different noise during training; and third, transformation to running statistics need not be undertaken during training, since each forward pass recomputes the relevant diagonal batch statistics in the current basis. 

    Thus, one may argue that since training-time evolution will likely continuously rotate the basis in which the covariance is diagonally approximated anyway, at least until the weights stabilise, this may justify an explicit basis transform used in the diagonalisation too. Hence, this treats BatchNorm with an approximate $\mathrm{O}\left(n\right)$ equivariance, rather than permutation $S_n$ equivariance only --- especially during training time.  
    
    Similarly, one may periodically take the post-activation function vectors over a chosen set of inputs and determine the true covariance. Singular value decomposition of this symmetric matrix yields orthogonal matrices, which, in an isotropic network, can then act on the weights to produce representations in a new basis. In this basis, BatchNorm no longer produces a diagonal approximation to the whitening transform, but the exact whitening transform in reference to the chosen set due to the diagonalised covariance matrix. This is maintained at least until further training. At late training/inference time, this explicit transform may be more problematic because weights may have settled into a consistent basis for the diagonalised covariance, which downstream parameters may have specialised to. Therefore, we propose regularly reparameterising weights to diagonalise the covariance during training to mitigate the risk of downstream parameters specialising on a specific approximation. At inference time, the transform can ensure diagonal whitening is equivalent to a full whitening transform.
    
    At inference time, the estimated statistics must also be transformed appropriately, with mean, $\vec{\mu}$, transforming as a rank-1 quantity $\vec{\mu}'=\mathbf{R}\vec{\mu}$ and standard deviation, $\vec{\sigma}$ as a diagonal rank-2 approximation $\vec{\sigma}'\odot\vec{\sigma}'=\left(\mathbf{R}\odot\mathbf{R}\right)\left(\vec{\sigma}\odot\vec{\sigma}\right)$, then taking the elementwise square-root to recover $\vec{\sigma}'$ --- or recompute this diagonal from scratch as a stored statistic. If one wishes to undertake this with neuroadaptation, then SVD weight-diagonalisation and adaptation must precede the transform that diagonalises the covariance matrix. This treatment of BatchNorm not as elementwise standardisation with permutation equivariance, but as diagonally approximated full-whitening with approximate orthogonal equivariance, is important for enabling the latter independent component analysis approach on pre-whitened vectors, as detailed further in \textit{App.}~\ref{App:ICA} --- especially in tandem with a network which is regularly reparameterised to produce layerwise diagonalised covariance during training to prevent basis-specialisation.

    Separately, this indicates the taxonomy's utility even when compositional symmetries are absent, i.e. two surrounding weights are closed orthogonally, even though the intermediate activation breaks orthogonal algebraic equivariance symmetry, preventing the parameter symmetries or broader model symmetries. Yet, this unusual construction can thus be analysed with separate building blocks from the taxonomy, and composed together, revealing how one can handle BatchNorm's symmetry-breaking properties to behave favourably in this context. The continuous change of basis conducted in neuroadaptation and/or diagonalisation periodically during training also helps achieve a similar effect for elementwise approximations used within optimisers --- they may better approximate full standardisation than what their diagonal approximations initially intended. The training-time orthogonal closure of weights likely also causes unaccounted for effective basis changes to optimiser running statistics which it is unaffected by, encouraging the approximate second-order transforms proposed. Whilst periodic changes of basis may reduce an optimiser's dependence on any single coordinate system.

\newpage

\section{Alternative Methodologies to Thresholding \label{App:Thresholding}}

    The specified symmetry construction determines the form of neuroadaptation described and its function-preserving properties, but it does not uniquely determine when to perform it. This paper's overall approach uses scheduled and conditional neuroadaptation, but these methods and the ones enclosed are changeable. Alternatives can be explored more broadly while still using the function-preserving neuroadaptation process. The clearest alternatives are demonstrable in the current method's thresholding formulations.

    Here, thresholding is used to indicate when a neuron has an insignificant contribution to the overall model function, and therefore constitutes a so-called `scaffold neuron' --- a neuron which is available to specialise and learn if required, ultimately allowing the possibility for a network to expand its width/capacity if thresholding is surpassed, but until then is deemed otherwise negligible. 
    
    This is an important part of this particular approach to scheduled, conditional, neuroadaptation, but has many possible formulations. This section expands on several alternatives, comparing pros and cons, contextualised through specific isotropic activations and their spectral behaviours. The discussion starts by comparing different statistical strategies, then how these may be applied to activation or gradient information, or complement them with batch statistics. 

    Arguably the most straightforward approach is the one utilised in the main text and empirical section: setting a constant value against which singular values are compared. The chosen value was $0.95$ --- principally used to indicate whether a singular value amplifies or attenuates the prior activation in its map (assuming prior uniformity, aided somewhat by magnitude saturation). It is interpretable, given the use of orthogonal initialisation that begins the singular values at $1$, so a small offset below this, such as the decrease to $0.95$, is meaningful because it shows declining impact. The exact value of $0.95$ is not inherently principled; any similar $1-\epsilon$ consideration likely suffices. Overall, the benefits of this approach are that it is trivial to understand and is not elaborate, which minimises the risk of introducing additional complicated failure modes. Therefore, it remained an appropriate choice in the empirical section because its focus on mechanistic verification benefits from a minimalistic approach. However, this approach has downsides: a fixed cut is not adaptive to overall singular value spectral shifts in the data, which may occur during training, nor does it account for a magnitude-saturating activation function, which may make the relative proportions of singular values more meaningful than their absolute values. Hence, spectral information and statistics appear to be a favourable improvement, albeit less minimalistic.

    Consequently, a natural extension is to use statistics such as the mean, median, standard deviation or maximum of singular values in the thresholding formulation. First to consider is a mean multiplier $c\mathbb{E}\left(\sigma_i\right)$, a simple method that deems a neuron unimportant if its singular value is substantially smaller than the mean. Its positives include its simplicity and, crucially, scalability to overall spectral shifts---particularly useful if a layer operates at a different scale. However, large outliers may be problematic; even a few can produce aggressive neurodegeneration in the network. This is addressed by using a median, which mitigates the risk of large outliers while retaining scalable thresholding behaviour; however, its multiplier is rather unintuitive and distribution-dependent in how many scaffold neurons it may classify. A similar point applies to a multiplier of the maximum, which is also separately highly susceptible to outliers.

    Mean minus a multiplier of standard deviations, or comparably median minus a multiplier of median absolute deviation, may be more robust and interpretable as a threshold, selecting small tails of the singular value distribution. Setting the multiplier sufficiently high indicates statistically unusually low singular values, which could be interpreted as providing relatively little function (assuming roughly uniform-scale pre-activations). This also favours saturating isotropic activation functions, as it better measures relative contribution. Despite this, singular value distributions are non-negative, and the observed spectra in \textit{Fig.}~\ref{Fig:IsotropicSpectra} may be significantly non-Gaussian. This could, in principle, cause negative thresholding and result in uncontrolled neurogenesis. 
    
    Briefly, quantiles are largely incompatible with the scaffold-neuron-triggered neuroadaptation described, as they would fix the number of neurons at all times and therefore never trigger such processes. Comparably constant, predefined growth and pruning should be avoided due to the repeated replacement of small singular values; thus, a reactive implementation is essential.

    On relative methods: setting a threshold against a relative value such as $\lambda_i=N \sigma_i/\sum_j\sigma_j$ or $\lambda_i'=\sqrt{N}\sigma_i/\sqrt{\sum_j\sigma_j^2}$ may be practical, especially for the aforementioned case of saturating activation functions, and it is also scale-invariant. Here, $N$ indicates a layer width $N=\min\left(m,n\right)$ for the weights $\mathbb{R}^{n\times m}$, which must be accounted for if consistent scaling is desired. Despite this, this thresholding approach remains strongly affected by large outliers. 

    Finally, the spectral gap approach appears promising, though more implementationally nuanced, for selecting scaffold neurons. This appears especially attractive given the larger spectral gaps observed across the singular value time-evolution plots, shown in \textit{Fig.}~\ref{Fig:IsotropicSpectra} and the range of results of \textit{App.}~\ref{App:ExperimentTwo}. For example, one could threshold at the first spectral gap position from zero and deem all diagonalised neurons below it as `scaffold-neurons'. Retaining a limited number would preserve the gap's presence, while also enabling these neurons to transition from the scaffold-band, across the band-gap, to the active-band---then triggering neurogenesis, and vice versa. This seems especially well suited to the observed spectra, and we propose it as the most promising practical implementation outside of mechanistic verification. Its complexity, especially in quantifying such spectral gaps, may introduce peculiar failure modes tied to this thresholding approach, although the nature and mitigation of these can be explored in subsequent work. Conversely, if no persistent gap forms in the spectrum, such as in a very dense case, this method may be unsuitable; however, gaps appeared persistent in the isotropic results of \textit{App.}~\ref{App:ExperimentTwo}.

    This concludes a non-exhaustive list of thresholding approaches. The remaining paragraphs clarify how these may relate directly to singular values or their gradients and how batch statistics may enrich these categorising methods.

    One canonical separation in judging importance is activations-based thresholding or thresholding on their gradients. Generally, such considerations are assumed to apply only to activations, but they may also generalise to gradients. Neuroadaptation is suggested to run at scheduled intervals alongside standard gradient-based parameter optimisation; this means gradient information is readily available to the thresholding stage. A nuance is that the activation-gradient may not be in the diagonalised-neuron basis, but it can be trivially transformed by performing SVD on the corresponding weights, then applying the corresponding orthogonal transforms to the available gradients. Composing this with the gradient of the outputs to their squared-sum, enables a measure of how much that diagonalised neuron impacts functionality. 
    Alternatively, one could consider the magnitude of the gradient with respect to the loss \cite{Lee2019}, among others. Across these, many of the above thresholding methods can be employed to such gradients to determine outliers. For example, having a gradient within a small multiple of the standard deviation around zero would suggest little impact on the function's outputs --- thus it may be considered a scaffold neuron. Various other permutations could be considered, offering practical avenues to improve when neuroadaptation should occur. 
    
    One could also use these methods to examine whether rarely significant connections tend to produce non-robust maladaptations in the output, rather than a simple lack of impact, when preceding activations are not assumed to be uniform.  

    Finally, batch information could represent an underexplored indicator that could potentially complement the discussed approaches. For example, batch statistics coupled with such methods could indicate variability in a neuron's significance: if a diagonalised neuron's resultant activation is consistently low, even across a batch, then it is arguably a better candidate for being a scaffold neuron. This also relates to removing the assumption of pre-activation uniformity.

    Overall, this discussion indicates many alternatives to thresholding methods that may be productive to explore. The current minimalistic approach is preferable for the intended mechanistic verification and for reducing obscure failure modes, but alternatives may remain preferable if the intention is for competitive or production usage, and further exploration of these approaches and others not listed is encouraged.
    \newpage

\section{Parallelising Matrix-Vector Products and Nested Functional Classes\label{App:MatrixVector}}

    Generalising isotropic multilayer perceptrons yields an asymptotic nested functional class, much like residual networks, but arising from the functional form permitted by the chosen orthogonal symmetry. This is given by all possible isotropic networks of depth $d$ being denoted as the set $\mathcal{F}_d$ such that $\mathcal{F}_d\subseteq\mathcal{F}_{d+1}$. This structure may be significant, offering an alternative to classical residual \cite{He2016} connections, with an architectural constraint to maintain or monotonically increase width with depth, and a saturating isotropic function. Moreover, this approach clarifies a crucial decomposition showing that all matrix-vector computations can be parallelised up front for isotropic multilayer perceptrons.

    Considering the two-layer isotropic affine map provided by \textit{Eqn.}~\ref{Eqn:IsotropicExpansion1} and then re-expressing this equation using notation $\mathbf{f}=\bar{\sigma}\big(\left\|\vec{x}\right\|_2^2\big)\vec{x}$ for preactivation $\vec{x}$.

    \begin{equation}
        \vec{x}^{(d+1)}= \mathbf{W}^{(d+1)} \mathbf{f}\Big(\underset{\vec{x}^{(d)}}{\underbrace{\mathbf{W}^{(d)}\mathbf{f}\left(\vec{x}^{(d-1)}\right)+\vec{b}^{(d)}}}\Big)+\vec{b}^{(d+1)}
        \label{Eqn:IsotropicExpansion1}
    \end{equation}

    \begin{equation}
        \vec{x}^{(d+1)}= \bar{\sigma}\left(\big\|\vec{x}^{(d)}\big\|_2^2\right)\mathbf{W}^{(d+1)} \vec{x}^{(d)}+\vec{b}^{(d+1)}
        \label{Eqn:IsotropicExpansion2}
    \end{equation}

    Then expanding this recurrence relation, yields \textit{Eqn.}~\ref{Eqn:IsotropicExpansionTwo} for two layers, for new contracted variables $\mathbf{W}'$ and $\vec{b}^{(d)'}$.
    
    \begin{equation}
        \vec{x}^{(d+1)}= \bar{\sigma}\left(\big\|\vec{x}^{(d)}\big\|_2^2\right)\bar{\sigma}\left(\big\|\vec{x}^{(d-1)}\big\|_2^2\right)\underset{=\mathbf{W}^{(d+1)'}}{\underbrace{\mathbf{W}^{(d+1)}\mathbf{W}^{(d)}}} \vec{x}^{(d-1)}+\bar{\sigma}\left(\big\|\vec{x}^{(d)}\big\|_2^2\right)\underset{=\vec{b}^{(d+1)'}}{\underbrace{\mathbf{W}^{(d+1)}\vec{b}^{(d)}}}
        +\vec{b}^{(d+1)}
        \label{Eqn:IsotropicExpansionTwo}
    \end{equation}

    Which can subsequently be generalised again for $d$-layer isotropic perceptron, as shown in \textit{Eqn.}~\ref{Eqn:IsotropicExpansionGnrl}. For notational simplicity a further reparameterisation to $\mathbf{W}_{\circ}^{(d)}$ and $\vec{b}_{\circ}^{(d, i)}$ is utilisThe input $\vec{x}^{(0)}$ does not undergo $\mathbf{f}$'s non-linear transform due to it being the input; hence, indexing starts at $1$.

    \begin{align}
        \vec{x}^{(d+1)}= \vec{b}^{(d+1)}+S_1^d\mathbf{W}_{\circ}^{(d)}\vec{x}^{(0)}+
        \sum^{d}_{i=1}S_i^d\vec{b}_{\circ}^{\left(i,d\right)} && 
        S^d_m=\prod^d_{j=m}\bar{\sigma}\left(\big\|\vec{x}^{(j)}\big\|_2^2\right)
        \label{Eqn:IsotropicExpansionGnrl}
    \end{align}

    In this case, the layerwise architecture still matters for the $\bar{\sigma}$ terms, with the directionality of intermediate $\vec{x}$ also affecting the various non-linear terms. This reparameterisation generally constrains the parameters $\mathbf{W}_{\circ}^{(d)}$ and $\vec{b}_{\circ}^{(i, d)}$; however, it is suggested to broaden the functional class by allowing both to be new, trainable, real matrices of appropriate shape. Hence, instead of interdependence between $\mathbf{W}_{\circ}^{(d)}$ depthwise, they are now treated as independent parameters, enabling up to full rank matrices, not limited to being less than or equal to the minimum rank of the original matrices in the contraction --- the independent formulations are not subject to constraint $\operatorname{rank}\left(\mathbf{W}_{\circ}^{(d)}\right)\leq\min_k\operatorname{rank}\left(\mathbf{W}^{(k)}\right)$. Thus, unconstraining to general real matrices broadens the functional class and also enables further parallelisation at training time instead of requiring an initial sequential contraction of the intermediate weight matrices. They may also be trained as independent weight matrices.

    Under the assumptions of a maintained or monotonically increasing width, with a non-linear term of an isotropic activation function, $\bar{\sigma}$, capable of saturating to a non-zero constant, with intrinsic length parameter able to ensure such general saturation --- then in this saturation limit, the parameters of layer $d+1$ are able to embed the parameters of layer $d$, thus reproducing its function in that embedded (sub)space. In the uncontracted parameters, this is equivalent to a multiplication by a scaled identity, where the scaling is the inverse of the constant-saturated value of the non-linearity, which now acts as a constant linear scaling. Thus, given this architectural arrangement and asymptotic saturation limit, increasing depth retains the expressibility of all shallower networks while exhibiting a strictly equal-to-enlarged function class. This is not the case for typical multilayer perceptrons.

    Furthermore, one may note that by this construction, \textit{all} matrix-vector products, over all layers, $\vec{z}^{(d)}=\mathbf{W}_\circ^{(d)}\vec{x}^{(0)}$, may be precomputed \textit{in parallel} by producing the $\vec{z}^{(d)}$ contraction. This is displayed in \textit{Eqn.}~\ref{Eqn:IsotropicExpansionCntrct}, showing sequential computation only need occur for inner products, $a=\vec{x}^{(d)}\cdot\vec{x}^{(d)}$ in $\mathcal{O}\left(n\right)$ per-layer-scaling for $n$ width, a single non-linear operation, $\bar{\sigma}\left(a\right)$ in $\mathcal{O}\left(1\right)$ (compared to elementwise $\mathcal{O}\left(n\right)$ per-layer-scaling), scalar-vector product $\mathcal{O}\left(n\right)$ and vector-vector addition, $\mathcal{O}\left(n\right)$. This remains valid for the general $\mathbf{W}_{\circ}^{(d)}$ or $\vec{b}_{\circ}^{(i, d)}$ the original constrained forms: $\mathbf{W}_{\circ}^{(d)}=\prod^{d+1}_{i=1}\mathbf{W}^{(i)}$ and $\vec{b}_{\circ}^{(i, d)}=\left(\prod^{d+1}_{k=i+1}\mathbf{W}^{(k)}\right)\vec{b}^{(i)}$.
    
    \begin{equation}
        \vec{x}^{(d+1)}= \vec{b}^{(d+1)}+S_1^d \vec{z}^{(d)}+
        \sum^{d}_{i=1}S_i^d\vec{b}_{\circ}^{\left(i,d\right)}
        \label{Eqn:IsotropicExpansionCntrct}
    \end{equation}
    
    Hence, the computationally demanding matrix–vector products are \textit{all} parallelised and precomputed, whilst comparatively trivial scalar-vector products, inner products, vector additions and simple $\mathcal{O}\left(1\right)$ activations are sequential, potentially offering significantly improved GPU utilisation. 

    Between submission and acceptance of this paper, we developed these ideas into a distinct standalone workshop paper now accepted for the OPT'26 workshop at NeurIPS \cite{Bird2026a} that explores these principles more substantively alongside empirics.

\section{Architectural Generalisations\label{App:Transformers}}

This section discusses how the isotropic-neuroadaptive approach could be generalised to architectures beyond multilayer perceptrons.

\subsection{Sketching Adaptive-Channel Convolutional Networks\label{App:Convolution}}

    The procedure for isotropic-adaptive networks can be generalised for convolutional networks \cite{Fukushima1980, LeCun1998}. This section sketches these results and will develop them in future work.

    A dynamical convolution architecture is enabled by implementing isotropic activation functions, $\mathbf{f}$, channel-wise --- taking the norm over vectors in $\mathbf{R}^C$ for $C$ channels. In analogy to the multilayer perceptron setup, a partial diagonalisation of each convolution's kernel parameters, $\mathbf{K}_1\in\mathbb{R}^{C_{\text{out}}\times\left(C_{\text{in}}\times K_h\times K_w\right)}$, is possible as shown by \textit{Eqns.}~\ref{Eqn:Conv1}~through~\ref{Eqn:Conv4}, for $\mathbf{R}\in\mathrm{O}\left(C_{\text{out}}\right)$, $\mathbf{Q}\in\mathrm{O}\left(C_{\text{in}}K_h K_w\right)$ and diagonal $\mathbf{\Lambda}$.

    \begin{align}
        \mathbf{Y}&=\mathbf{K}_2\ast\mathbf{f}\left(\mathbf{K}_1\ast\mathbf{X}+\vec{b}_1\right)+\vec{b}_2
        \label{Eqn:Conv1}\\
        \mathbf{Y}&=\mathbf{K}_2\ast\mathbf{f}\left(\left(\mathbf{R}\mathbf{\Lambda}\mathbf{Q}^T\right)\ast\mathbf{X}+\vec{b}_1\right)+\vec{b}_2
        \label{Eqn:Conv2}\\
        \mathbf{Y}&=\underset{=\mathbf{K}_2'}{\underbrace{\mathbf{K}_2\mathbf{R}}}\ast\mathbf{f}\left(\underset{=\mathbf{K}_1'}{\underbrace{\mathbf{\Lambda}\mathbf{Q}^T}}\ast\mathbf{X}+\underset{=\vec{b}_1'}{\underbrace{\mathbf{R}^T\vec{b}_1}}\right)+\vec{b}_2
        \label{Eqn:Conv3}\\
        \mathbf{Y}&=\mathbf{K}_2'\ast\mathbf{f}\left(\mathbf{K}_1'\ast\mathbf{X}+\vec{b}_1'\right)+\vec{b}_2
        \label{Eqn:Conv4}
    \end{align}
    In effect, the spatial indices of convolution must be kept consistent, but rotations can commute with isotropic functions, thereby partially diagonalising channels and enabling the described adaptation. Generalising the approach is non-trivial when considering both padding and residual constructions, requiring further development; yet this notated approach presents a route to broader generalisation. Similarly, such a fully diagonalised, partially diagonalised or ICA-based approach may offer interpretable insights into representations, with the caveat that representations may change under isotropic functions \cite{Bird2025c}.

    \subsection{Attention Mechanisms}

    The self-attention mechanism \cite{Lin2017, Vaswani2023} within a transformer \cite{Vaswani2023} architecture is also a candidate where neuroadaptation appears feasible. Single-headed self-attention is characterised by the equation presented in its canonical form in \textit{Eqn.}~\ref{Eqn:AttentionCanonical} and in a tensor-indexed form in \textit{Eqn.}~\ref{Eqn:AttentionIndexed} (notation following from \citet{Vaswani2023}).

    \begin{equation}
        \operatorname{Attention}\left(Q, K, V\right) = \operatorname{softmax}\left(\frac{QK^T}{\sqrt{d_k}}\right)V
        \label{Eqn:AttentionCanonical}
    \end{equation}

    The following uses notation $\mathbf{W^{(K)}}, \mathbf{W^{(Q)}}\in\mathbb{R}^{d\times d_k}$, and $\mathbf{W^{(V)}}\in\mathbb{R}^{d\times d_v}$ for key, query and value weight matrices respectively and $\mathbf{X}\in\mathbb{R}^{T\times d}$ for the input (and $T$ tokens). The subscript $m$ on softmax indicate it acts over columns indexed by $m$.
    
    \begin{equation}
    \operatorname{Attention}\left(\mathbf{X}; \mathbf{W^{(K)}}, \mathbf{W^{(Q)}}, \mathbf{W^{(V)}}\right) = \operatorname{softmax}_{(m)}\left(\frac{\mathbf{X}_{ij}\mathbf{W^{(Q)}}_{jq}\mathbf{W^{(K)}}_{lq}\mathbf{X}_{ml}}{\sqrt{d_k}}\right)_{im}\mathbf{X}_{mp}\mathbf{W^{(V)}}_{pn}
        \label{Eqn:AttentionIndexed}
    \end{equation}

    Due to the $\mathbf{W^{(Q)}}\mathbf{W^{(K)}}^T$ contraction, an identity, $\mathbf{I}$, can be inserted between them and separated into a pair of orthogonal actions, $\mathbf{R}$, as follows: $\mathbf{I}=\mathbf{R}^T\mathbf{R}$. Thus, these exhibit a form of orthogonal closure (\textit{Eqn.}~\ref{Eqn:Closure}) in their parameterisation. Consequently, one could perform a right- or left-sided partial diagonalisation of the respective weight, exposing its singular values and rank.

    Then one can act with the neuroadaptation approach to change the dimension $d_k$ within this self-attention. However, care is needed because the full closure is for the general linear group, $\operatorname{GL}\left(d_k, \mathbf{R}\right)$, and can thus be used to reparameterise the matrices in a way that affects singular values. One could construct an agreed special representation for the parameters to threshold, such as balancing singular values between terms, as shown in \textit{Eqn.}~\ref{Eqn:AttentionSingularBalancing} (and resizing $\mathbf{U}\mathbf{\Lambda}^{\frac{1}{2}}$ and $\mathbf{V}\mathbf{\Lambda}^{\frac{1}{2}}$ such that they are $\mathbb{R}^{d\times d_k}$ matrices).

    \begin{equation}
        \mathbf{M}=\mathbf{W^{(Q)}}\mathbf{W^{(K)}}^T = \mathbf{U}\mathbf{\Lambda}\mathbf{V}^T=
        \underset{=\mathbf{\bar{W}^{(Q)}}}{\underbrace{\left(\mathbf{U}\mathbf{\Lambda}^{\frac{1}{2}}\right)}}
        \underset{=\left(\mathbf{\bar{W}}^{(K)}\right)^T}{\underbrace{\left(\mathbf{\Lambda}^{\frac{1}{2}}\mathbf{V}^T\right)}}
        \label{Eqn:AttentionSingularBalancing}
    \end{equation}

    Additionally, the $d_k$ denominator needs addressing, such as rescaling both matrices by a factor $\sqrt[4]{d_k'/d_k}$.

    Similarly, one could consider $\mathbf{W^{(V)}}$, where one could cautiously neuroadapt $d_v$ by considering a representation of the orthogonal group $\operatorname{O}\left(d_v\right)$ acting on the weight matrix subsequent to the multi-head concatenation, $\mathbf{W^{(H)}}$. Consequently, one could adapt each head independently, but it would end up with potentially varying $d_v$ per head. Similarly, depending on the construction, general-linear symmetries may apply, with analogous considerations as before for $d_k$.
    
    Trivially, the MLP aspects of a transformer could utilise neuroadaptation if multilayered --- in exact comparison to the MLP implementation presented in this paper. This could occur for the other axis of $\mathbf{W^{(H)}}$, if it has a subsequent layer before recombination into the residual term. The dimension of $\mathbf{X}$ corresponding to $d$ may be varied, if constructed as a multilayer $\mathbf{X}=\mathbf{W}\mathbf{Y}$, since singular value decomposition of $\mathbf{W}$ could have its orthogonal transforms simultaneously absorbed into $\mathbf{W^{(Q)}}, \mathbf{W^{(K)}}, \mathbf{W^{(V)}}$, enabling neuroadaptation. But it is essential to consider whether the particular architecture being used allows this to affect the residual stream.

    Token dimensions and layers that combine into residual streams pose problems for neuroadaptation. The former should not be transformed with an orthogonal symmetry, since it represents discrete, time-ordered tokens. It is therefore a conceptually and mathematically poor candidate for neuroadaptation in any straightforward way. Similarly, reparameterisation that requires orthogonal transforms to commute with residuals adds considerable complexity, as many parameters at different locations require simultaneous orthogonal closure with the transform. Therefore, this requires substantial nuance to implement, which is out of scope for the present work.

    Overall, several promising directions for neuroadaptation in transformer models exist. These allow several internal dimensions to change neuron counts while preserving function. However, token counts at present appear unsuitable for such approaches. This presents an interesting line of enquiry for future generalisations of this neuroadaptive approach.

    \subsection{Hybrid Architectures}

    As discussed in the neuroadaptability of attention, a network need not be wholly isotropic to permit a layer's neuroadaptation --- two affine layers on either side of an isotropic activation are sufficient for adaptation, enabling the designed reparameterisation invariance. Therefore, once such properties and biases are suitably understood, one could construct other layers with various symmetry-led properties to induce a repertoire of desirable behaviours coexisting throughout the network in differing locations. Hybrid constructions are therefore permitted, where various primitives can be used in network construction.

\section{Independent Component Analysis (ICA) Network Representation\label{App:ICA}}

    Independent component analysis (ICA) is a standard statistical technique for linearly separating a mixture of signals into often meaningful, statistically independent channels for interpretation --- typically relying on non-Gaussianity for separation. This could be used for interpretable isotropic networks at inference time, up to the inherent concept resolvability of ICA itself for hidden-layer representations. Given interpretable components produced by ICA, one may monitor the activation elements to determine a network's current expression of internal concepts. 

    This approach may be utilised only on the premise that the multilayer perceptron is constructed from sequential blocks of affine-map $\rightarrow$ isotropic activation function $\rightarrow$ BatchNorm. It must then be represented in its covariance-eigenbasis, diagonalising the post-activation function's covariance. The network should also be trained such that it is not contingent on any particular basis inherited from BatchNorm's diagonal-covariance approximation but largely invariant to orthogonal transforms of the affine layers (discussed in the BatchNorm section of \textit{App.}~\ref{App:Normalisers}). Thus, this ensures a pre-whitened signal ready for the subsequent ICA step.

    One may then use various approaches to determine the orthogonal matrix required by ICA to transform the whitened data --- such as a loss that encourages non-gaussianity and optimising in the corresponding $\mathfrak{so}\left(n\right)$ Lie algebra space. Having determined the appropriate orthogonal transform\footnote{I am grateful to Dmitry Kangin for our discussion of isotropic functions and their orthogonal equivariance, during which he reminded me of the orthogonal transformations arising from independent component analysis and suggested their potential relevance. This section subsequently develops that connection further.}, $\mathbf{Q}$, one may reparameterise the network in one of three distinct approaches, given that an identity is postcomposed with BatchNorm $\mathrm{I}=\mathbf{Q}^T\mathbf{Q}$ and that the $\mathbf{Q}^T$ matrix is used to reparameterise the subsequent affine-layer weights: $\mathbf{W}_{(n+1)}'=\mathbf{W}_{(n+1)}\mathbf{Q}^T$. These proceed identically in algebra to the partial-diagonalisation insertion of an identity, but with a different pair of rotations and corrections to BatchNorm's statistics, detailed below. Transforms are detailed in \textit{Eqns.}~\ref{Eqn:ICAstart}~through~\ref{Eqn:ICAend}, for parameterless BatchNorm, $BN\big(\vec{x};\vec{\mu},\vec{C}\big)$ and isotropic activation function $\mathbf{f}$.

    \begin{align}
        \vec{x}_{(n+1)}&=\mathbf{W}_{(n+1)}\vec{z}_{\text{whi.}}+\vec{b}_{(n+1)} &&\vec{z}_{\text{whi.}}=\mathrm{BN}\left(\mathbf{f}\left(\mathbf{W}_{(n)}\vec{x}_{(n)}+\vec{b}_{(n)}\right);\vec{\mu},\vec{C}\right)\label{Eqn:ICAstart}\\
        \vec{x}_{(n+1)}&=\mathbf{W}_{(n+1)}\Big(\underset{=\mathrm{I}}{\underbrace{\mathbf{Q}^T\mathbf{Q}}}\Big)\vec{z}_{\text{whi.}}+\vec{b}_{(n+1)} &&\vec{z}_{\text{whi.}}=\mathrm{BN}\left(\mathbf{f}\left(\mathbf{W}_{(n)}\vec{x}_{(n)}+\vec{b}_{(n)}\right);\vec{\mu},\vec{C}\right)\\
        \vec{x}_{(n+1)}&=\underset{=\mathbf{W}_{(n+1)}'}{\underbrace{\mathbf{W}_{(n+1)}\mathbf{Q}^T}}\underset{=\vec{z}_{\text{ICA}}}{\underbrace{\mathbf{Q}\vec{z}_{\text{whi.}}}}+\vec{b}_{(n+1)} &&\vec{z}_{\text{whi.}}=\mathrm{BN}\left(\mathbf{f}\left(\mathbf{W}_{(n)}\vec{x}_{(n)}+\vec{b}_{(n)}\right);\vec{\mu},\vec{C}\right)
    \end{align}

    From this point, there is an explicit $\mathbf{Q}$ matrix product retained, which may remain as is, necessitating an additional matrix-vector product per sample, or be absorbed into prior parameters to increase efficiency. However, the latter option is nuanced, requiring the assumption that the network has been trained to be approximately invariant to basis changes to BatchNorm described in \textit{App.}~\ref{App:Normalisers} if an elementwise $\vec{C}$ is desired, or promoting this to a full rank-2 transform requiring a covariance matrix $\mathbf{C}$ (particularly its inverse $\mathbf{C}^{-1}$).
    
    \begin{align}
        \vec{x}_{(n+1)}&=\mathbf{W}_{(n+1)}'\underset{=\vec{z}_{\text{ICA}}}{\underbrace{\mathbf{Q}\vec{z}_{\text{whi.}}}}+\vec{b}_{(n+1)} &&\vec{z}_{\text{whi.}}=\mathrm{BN}\left(\mathbf{f}\left(\mathbf{W}_{(n)}\vec{x}_{(n)}+\vec{b}_{(n)}\right);\vec{\mu},\vec{C}\right)\\
        \vec{x}_{(n+1)}&=\mathbf{W}_{(n+1)}'\vec{z}_{\text{ICA}}+\vec{b}_{(n+1)} &&\vec{z}_{\text{ICA}}=\mathrm{BN}\Big(\mathbf{f}\Big(\underset{=\mathbf{W}_{(n)}'\vec{x}_{(n)}+\vec{b}_{(n)}'}{\underbrace{\mathbf{Q}\Big(\mathbf{W}_{(n)}\vec{x}_{(n)}+\vec{b}_{(n)}\Big)}}\Big);\vec{\mu}',\vec{C}'\text{ or }\mathbf{C}'\Big)\\
        \vec{x}_{(n+1)}&=\mathbf{W}_{(n+1)}'\vec{z}_{\text{ICA}}+\vec{b}_{(n+1)} &&\vec{z}_{\text{ICA}}=\mathrm{BN}\Big(\mathbf{f}\Big(\mathbf{W}_{(n)}'\vec{x}_{(n)}+\vec{b}_{(n)}';\vec{\mu}',\vec{C}'\text{ or }\mathbf{C}'\Big)\label{Eqn:ICAend}
    \end{align}

    These latter reparameterisation approaches temporarily assume the orthogonal equivariance of the full-whitening and isotropic activation function, commuting with $\mathbf{Q}$ such that $\mathbf{Q}$ may be absorbed into the preceding affine layer's weights, $\mathbf{W}_{(n)}'=\mathbf{Q}\mathbf{W}_{(n)}$, and biases $\vec{b}_{(n)}'=\mathbf{Q}\vec{b}_{(n)}$. 
    
    Assumed to be in a covariance-diagonalised representation ready for inference-time, $\mathbf{C}=\operatorname{diag}^{\dagger}\big(\vec{C}\big)$, one may then choose to appropriately transform the mean, $\vec{\mu}'=\mathbf{Q}\vec{\mu}$, and covariance matrices $\mathbf{C}'=\mathbf{Q}\mathbf{C}\mathbf{Q}^T$. If full $\mathbf{C}'$ is retained, this necessitates a matrix-vector product per sample for full equivariance, rather than the diagonalised vector-vector Hadamard product. Alternatively, one may choose to retain the efficient vector-vector product nature of BatchNorm by approximating the original function with a diagonal-approximated covariance $\vec{C}'=\left(\mathbf{Q}\odot\mathbf{Q}\right)^T\vec{C}$ (inexact). Either of the latter two methods seems preferable to the first exact approach, which adds an unnecessary matrix-vector product.

    \begin{table}[h]
    \caption{Two primary approaches for a network in the `ICA-representation' with respective trade-offs.}
    \centering
    \resizebox{\textwidth}{!}{%
    \begin{tabular}{lll}
    \hline
    \textbf{Variant} & \textbf{Functional Invariance} & \textbf{Computational Cost} \\
    \hline
    $Q$ absorbed with full covariance 
    & Exact, under assumptions 
    & Loses diagonal BatchNorm efficiency \\
    
    $Q$ absorbed with diagonal covariance 
    & Approximate 
    & Retains BatchNorm's efficiency \\
    \hline
    \end{tabular}%
    }
    \end{table}

    In both exact cases, the `neurons' associated with the vectors decomposed in the standard basis now represent independent components as specified by ICA and can be interpreted accordingly. The inexact approach approximates such components given the BatchNorm diagonal preweighting approximation. Thus, this ICA representation of the network may enable real-time monitoring of the network's hidden states and interpretability therein, at least up to the inherent concept resolvability of ICA.
    

    Alongside diagonalising via singular value decomposition, this approach may enhance interpretability but does not concurrently permit sparsification for more efficient computation and memory, unless in the exceedingly unlikely case that both transforms coincide. Thus, this introduces an interpretability-sparsification tradeoff within isotropic networks. This approach still retains isotropic MLP `big-$\mathcal{O}$' scalings detailed in \textit{App.}~\ref{App:MatrixVector} which may be preferable to standard MLP scalings. Corresponding approaches may also apply to the sketched convolutional form of \textit{App.}~\ref{App:Convolution} for interpretability. Unlike sparsification, which may be applied only to every other layer, the `ICA-representation' of the network can be introduced for every hidden-layer map. 
    
    Overall, this demonstrates the flexibility of standard orthogonal prescribed primitives beyond neuroadaptation.

    


\section{Notation}
\begin{multicols}{2}
\begin{itemize}
    \item $\mathbb{R}_{\geq 0}$: Non-negative real numbers.
    \item $\mathbb{Z}_{+}$: Positive integers.
    \item $\vec{a}$: Vectors. $\mathbf{A}$: Matrices.
    \item $\circ$: Function composition. 
    \item $\odot$: Hadamard, or elementwise, product.
    \item $\|\vec{x}\|_2$: Euclidean norm of $\vec{x}$.
    \item $\hat{x}$: Unit-normalised vector $\hat{x}=\vec{x}/\|\vec{x}\|_2$.
    \item $\hat{e}_i$: $i$th standard basis vector.
    \item $\delta_{ij}$: Kronecker delta.
    \item $\mathrm{I}_n$ or $\mathrm{I}$: Identity matrix.
    \item $\operatorname{diag}(\mathbf{A})$: Vector formed from the diagonal entries of matrix $\mathbf{A}$.
    \item $\operatorname{diag}^{\dagger}(\vec{v})$: Diagonal matrix whose diagonal entries are the components of $\vec{v}$.
    \item $\mathbb{P}$: Probability measure.
    \item $\mathcal{O}(\cdot)$: Big-$\mathcal{O}$ notation.
    \item $\mathcal{G}$: Group and $g\in\mathcal{G}$: an element of that group.
    \item $\varrho^{(1)}(g)$, $\varrho^{(2)}(g)$, $\varrho\left(g\right)$: Different group representations.
    \item $\mathrm{O}(n)$: Orthogonal group in dimension $n$.
    \item $\mathrm{GL}_n(\mathbb{R})$: General linear group of invertible $n\times n$ real matrices.
    \item $S_n$: Permutation group on $n$ elements.
    \item $\mathbf{P}_{\pi}$: Matrix representation of permutation $\pi$.
    \item $P=\{\varrho(\pi)\mid \pi\in S_n\}$: Set of all permutations in the group in a chosen matrix representation.
    \item $O=\{\varrho(Q)\mid Q\in\mathrm{O}(n)\}$: Set of all elements in the orthogonal group in a chosen matrix representation.
    \item $[\mathbf{R},\mathbf{f}]$: Commutator $\mathbf{R}\circ\mathbf{f}-\mathbf{f}\circ\mathbf{R}$.
    \item $\mathcal{F}$: Functional class.
    \item $\mathcal{F}_d$: Functional class of isotropic networks of depth $d$.
    \item $f$, $\mathbf{f}$: Function, emboldened notation is a vector-valued function.
    \item $f_{\mathrm{Aff.}}$: Affine map.
    \item $f_{\mathrm{Iso.}}$: Isotropic activation function.
    \item $\sigma$: Scalar nonlinearity used to construct an isotropic activation, with alternative formulations $\tilde{\sigma}$, $\bar{\sigma}$ defined by \textit{Eqns.}~\ref{Eqn:IsotropicFunctionalForm}.
    \item $J_{\mathbf{f}}$: Jacobian of function $\mathbf{f}$.
    \item $\mathbf{N}(\vec{x})$: Parameterless $L_2$ normalisation map, $\mathbf{N}(\vec{x})=\vec{x}/\|\vec{x}\|_2$.
    \item $\mathrm{LN}$: LayerNorm.
    \item $\mathrm{BN}$: BatchNorm.
    \item $\mathbf{\Lambda}$: Diagonal or rectangular diagonal matrix of singular values.

    \item $\vartheta$: Singular-value threshold.
    \item $\Xi$: Desired number of scaffold neurons to maintain below threshold.
    \item $\omega$: Current number of scaffold neurons, defined by singular values below $\vartheta$.
    \item $o$: Intrinsic length parameter introduced for neuroadaptation.
    \item $b_{+}$: Bias component associated with an appended neuron/channel
    \item $b_{\times}$: Bias component associated with a deleted neuron/channel
    \item $\vec{W}_{+}$: Appended downstream weight column associated with a scaffold neuron/channel.
    \item $\vec{W}_{\times}$: Deleted downstream weight column associated with a neuron/channel.
    \item $\vec{\psi}$: Compensatory vector parameter used to absorb additive terms arising during pruning.
    \item $\epsilon$: Mean per-example $L_2$ discrepancy between logits or activations before and after adaptation or sparsification.
    \item $\varepsilon$: Small positive stabilisation constant in normalisers.
    \item $\Sigma_0$: Mean initial sum of singular values in Experiment One.
    \item $\Sigma_1$: Sum of singular values after neuroadaptation in Experiment One.
    \item $\theta$: Generic model parameter; also used when counting non-zero parameters.

    \item $\mathcal{L}$: Loss function.  $\nabla_{\vec{a}}\mathcal{L}$: Gradient of the loss with respect to parameter vector $\vec{a}$.
    \item $\eta$: Learning rate.
    \item $\mathbf{R}^{\odot 2}=\mathbf{R}\odot\mathbf{R}$: Shorthand for elementwise square of an orthogonal matrix. Generally $\mathbf{R}^{\odot n}$ for elementwise power.
\end{itemize}
\end{multicols}

\section{Plain-Language Glossary}
    This glossary provides straightforward, plain-worded explanations of several terms used in this paper. These are not precise formal or mathematical definitions; they are meant to reduce the burden of niche terminology and provide a directional sense of the intended meaning. 

    First, we begin with some general terminology, encompassing logical and philosophical meanings:
\begin{itemize}
    \item \textbf{Prescribed}: A central descriptor used in the paper to indicate our approach to symmetry: it is used to indicate that symmetry is not a property that is found as a happenstance when examining activation functions (or other operations), but instead is implemented on purpose with the desire to introduce some property that requires it. It is used in the same sense as `prescribing' a medicine: an act of intentionally adding something to yield some consequence. This distinguishes it from approaches that find and examine symmetries already incidentally present in a deep learning system.
    
    \item \textbf{Deduced}: Indicates that some knowledge is derived logically from something known. In particular, we use it to indicate that properties such as symmetry can be determined from known functional forms---like how standard elementwise activation functions can be shown to obey a particular symmetry relation. We use this in contrast to `prescribed', to indicate methods which take known functions and derive conclusions from them (loosely: prescribed acts in reverse: taking a property and designing a function to obey it).

    \item \textbf{Ontology}: Used in the sense of which structures are considered fundamental to a construction, distinguishing them from what follows from those foundations. It indicates which concepts form the base of a framework, from which others are constructed. It is somewhat connected to the weaker term 'definitional', but carries a stronger philosophical meaning of the most foundational building blocks needed to capture the broader structure. In this case, for the broad topic of neural network systems, classically, there are some units, `neurons', that interconnect in some manner, the `network', transforming and intercommunicating information. Here, the neuron-unit is foundational to that description.

    \item \textbf{Emergent}: A property that occurs indirectly as a consequence of another or others; it is not a basic part of the system, but arises from the construction. For example, elementwise individual neurons were originally used as foundational building blocks of neural systems circa the 1950s \cite{Rosenblatt1958}, but since circa the 1990s \cite{Sussmann1992} have been understood to have symmetry properties. This is not solely intended as `strong emergence' where a property is more than its parts.
    
    \item \textbf{Ontological Reversal}: This swaps something seen as emergent/derivative with something foundational or definitional. In this case, a neuron-unit is classically considered a fundamental constituent of a neural network, from which symmetries may arise; instead, the `ontological reversal' replaces this with symmetry-fundamentality, from which neurons can then be recovered as an emergent property.  
    
\end{itemize}

    The following terms belong to more mathematical origins, in particular linear algebra (including eigenanalysis) and group theory.
    
\begin{itemize}
    \item \textbf{Group Theory}: A broad mathematical topic examining the relations among elements under an operation. 
    
    \item \textbf{Symmetry}: Indicates that a specified property remains unchanged or varies with it whilst another is transformed.
    
    \item \textbf{Group}: A collection of objects interrelated through an operation with specified rules. A classic example is 2-dimensional 90-degree rotations; you can reach 180 degrees by composing several different combinations of 90-degree actions together.
    
    \item \textbf{Orthogonal}: Classically used as `perpendicular', such as two vectors perpendicular to one another; this can be achieved through rotations and reflections. An orthogonal \textit{group} is a collection of rotations and reflections. It gets the name `orthogonal' because, in its standard representation, one property of the columns of its matrix is that they are perpendicular vectors.

    \item \textbf{Group Representation}: The elements of a group are often very abstract, like a rotation. There are many ways to write them down, whilst keeping the group's structure; for example, a rotation represented as a rotation matrix. This is one representation of that group; another may be the imaginary number symbol (or even quaternions) and how it relates to rotations through euler's formula.
    
    \item \textbf{Group action}: Groups may contain many elements; a group action is taking a single member of the group and applying it to something. For example, applying a 90-degree rotation to a rectangle.

    \item \textbf{Commutation/Commute}: Meaning that the order of application doesn't matter: A then B or B then A.

    \item \textbf{Commutator}: A specific operation related to commutation; in this case, action A then B minus action B then A.

    \item \textbf{Basis independence}: Not tied to any particular basis (coordinate system). For example, mass does not depend on whether we choose south to be north on the Earth, or east to west. In neural network terms, one may consider something basis-independent if the observed quantity does not fundamentally change when all matrices, vectors, etc. are measured in a different basis. This property is not present when using elementwise functions, as their transforms depend on specific directions corresponding to the coordinate system used.
    
    \item \textbf{Conjugation}: A combination of group actions acting on an object in a particular way. In this case, one could take a group element, like a 2D clockwise 60-degree rotation, then conjugate it with an anticlockwise 120-degree rotation: this yields anticlockwise by 120 degrees, then 60-degrees clockw120 degrees20-degrees clockwise (in this case, they commute, so the anticlockwise and clockwise 120-degree actions cancel). Conjugation is commonly used to change an object's basis.
    
    \item \textbf{Reparameterisation}: Representing the same thing differently. If one chooses matrices such that both can be rotated in opposing ways that cancel, and they always occur as multiplying one another, then one could choose to show that multiplication as they originally were or with the rotations applied first --- they appear different superficially, but achieve the same result.
    
    \item \textbf{Equivariance}: A property that varies in a predictable way in response to another's variations.
    
    \item \textbf{Invariance}: A property that does not vary in response to another's variations.

    \item \textbf{Functional Invariance}: When something changes within the system, the system's overall output remains the same.
    
    \item \textbf{Functional Form}: A general structural family of equations, or more straightforwardly a general equation with placeholders for specific maps. For example, an elementwise activation function describes no single activation function, but a collection of familiar specific instances: ReLU, Leaky-ReLU, sigmoid, etc.
        
    \item \textbf{Equivalence class}: Given some form of criterion, a collection of objects which are treated as the same. For example, a collection of differently parameterised networks may be treated as an equivalence class if their function is identical for all given inputs.
    
    \item \textbf{Singular value}: Non-negative real numbers that are square roots of the eigenvalues of a matrix premultiplied by its transpose.
    
    \item \textbf{Diagonalisation}: The process of reparameterising a matrix (and nearby matrices) so that all non-diagonal elements of the target matrix are zeroed, leaving only a diagonal of non-zero elements.
    
    \item \textbf{Standard representation}: loosely, a representation of a group acting on its corresponding vector space. Very loosely, this is the most typical way to represent a group element; for example, 3D rotations are represented as 3-by-3 rotation matrices.
    
    \item \textbf{Trivial representation}: This maps all group elements to a single identity element, since it preserves the group relations. This is used particularly in invariance equations, since all of a group's elements become an identity transformation, so they leave a quantity unchanged, as required by `invariance'.
    
\end{itemize}

The following terms stem from more neuroscientific, general deep learning principles, or newly introduced terminology.

\begin{itemize}
    \item \textbf{Neuroadaptation}: In this paper's narrower sense, this means changing a network's construction by varying the number of neurons. More broadly, it may refer to the network's overall structure, including interconnectedness.
    
    \item \textbf{Neurogenesis}: Creating new neurons to the system. In this paper, this constitutes enlarging a layer's width by adding new neurons.
    
    \item \textbf{Neurodegeneration}: Typically, this has a broad meaning referring to neuronal damage, loss-of-function, etc. In this instance, we use it narrowly to mean the removal of neurons from the system, effectively the antonym of neurogenesis. 
    
    \item \textbf{Connectivity pruning}: Removing interconnections between neurons, not the neurons themselves, contrasting with neurodegeneration.

    \item \textbf{Sparsification}: Removing connections in a matrix, particularly to reduce the computational burden. Typically, this removes connections deemed unimportant and therefore approximates function; however, in this paper it arises as a mathematical consequence of reparameterisation rather than direct conditional removal, so it is not an approximation.

    \item \textbf{Individuated Neuron}: A unit in a neural learning system that is inherently distinct and separable from other units in an agreeable way, regardless of the basis used to measure it. In this paper, the elementwise functions distinguish directions (those of the standard, canonical basis) and therefore create a distinction to those directions, to which we assign the classical term `neuron'. It is unnatural to deviate from those directions as a representing neural unit, since they are inherently special because of network functions.
    
    \item \textbf{Deindividuated}: Something that may typically be an individual unit having its distinctive separation from other units removed. Neurons are usually single units, but this paper uses mathematics that makes such separation between neurons not fundamental --- differing coordinate systems (bases) may disagree on the natural way to separate them into units.
    
    \item \textbf{Isotropic}: Means the same (iso-) in every direction (-tropic). This paper uses it to mean a function which applies equally in all directions. This does not imply homogeneity; a special centre may still exist even if it is isotropic. This is standard Greek and is frequently used in several scientific disciplines, including machine learning. For example, in astrophysics, a common assumption is that the universe is isotropic on large scales --- \textit{approximately} uniform no matter the patch of sky we observe.
    
    \item \textbf{Anisotropic}: Not (an-) the same (-iso-) in every direction (-tropic). Something that varies with direction.
    
    \item \textbf{Isotropic Activation Function}: A special form of activation function that applies the same transform regardless of direction. This contrasts with standard elementwise activation functions, which inherently vary with direction (anisotropic activation functions).
    
    \item \textbf{Primitives}: Basic building blocks of a neural network, although not necessarily ontological parts. For example, activation functions, affine maps, normalisations and initialisations are all largely irreducible components frequently used in composition or more generally to construct a neural network.
    
    \item \textbf{Neural Learning Systems}: The system of algorithms structured based upon neurons and their interconnectivity --- this need not actually be machine learning if hand-chosen implementations occur.
    
    \item \textbf{Scaffold neuron}: A neuron present in the system which is currently inactive, providing little to no meaningful participation in the computation being performed; at its extreme, it is present structurally only. The aim is that these 'inactive'-like neurons may acquire function as needed through the learning algorithm, specialising into a computational role.
    
    \item \textbf{Intrinsic length}: A newly introduced quantity, sometimes a trainable parameter; it changes a vector's measured length regardless of its entries. For technical reasons (even in the positive-definite case, it violates absolute homogeneity $\left\|\alpha\vec{x}\right\|\neq |\alpha|\left\|\vec{x}\right\|$ and zero-norm $\big\|\vec{0}\big\|\neq0$), it does not strictly satisfy the mathematical notion of a length due to several of its axioms. 
    
    \item \textbf{Functional class}: Given a general functional form, or many composed together in a system, the set of all possibilities which are reachable by choosing specific instances of that form. For example, all two-layer perceptrons, with any possible weights and biases. 

    \item \textbf{Closure Symmetry}: The weakest of our three proposed symmetry strengths, but encompasses all stronger ones. It states that given a collection of objects, if each is acted on by a specified transformation in a group, the resulting transformed object becomes another entry in that original collection. Hence, the transformation leaves all objects still within their original collection.

    \item \textbf{Probabilistic Symmetry}: A stronger, but intentionally flexible, symmetry condition that indicates the transformation leaves some probability or statistical property unchanged. For example, if vectors of the same length are equally likely to be initialised, then rotating one vector to another remains equally probable, as length has not changed.

    \item \textbf{Algebraic Symmetry}: The strongest of the three proposed symmetry constraints, it indicates that applying the transform to an object in a collection results in the exact same object as you started with (not just another member of the collection).
    
\end{itemize}

\section{Limitations, Assumptions, and Applicability\label{App:Limitations}}
    This section clarifies and discusses the known limitations of the approaches discussed, ranging from diagonalisation and architectural considerations to approximations.

    Due to the nature of the diagonalising transforms, this approach requires two parameterised maps that exhibit closure invariance under standard orthogonal transforms on the respective sides. One may use no or any number of interspaced standard orthogonal algebraic equivariant functions to achieve this. In practice, this likely necessitates two densely parameterised layers for the orthogonal transform to act on, and a non-linear function that is orthogonal-equivariant, e.g. isotropic activation functions. Isotropic multilayer perceptrons are straightforward networks that exhibit these properties, with a potential generalisation to convolutional networks described in \textit{App.}~\ref{App:Convolution} and to other state-of-the-art networks, such as transformers \cite{Vaswani2023} (\textit{App.}~\ref{App:Transformers}), with more nuanced relations. Thus, it seems possible to broaden the architectural constraints beyond MLP networks, but further work is needed to determine the exact implementation details. Some residual networks may be redesigned to exhibit this property, but diagonalisation introduces complications, as all residually connected layers must be transformable simultaneously. Nevertheless, MLP architectures remain pertinent in applications with unknown inductive biases, inhibiting the specialisation of the architecture class. They are also widely used as backbone subcomponents in larger networks, including transformers, where such neuroadaptive techniques can be applied when two or more such layers occur sequentially.
    
    Another consideration is the orthogonal-prescribed primitives in an MLP, and whether it retains the expressibility of typical construction with permutation-prescription. The adaptive topology construction does not depend on any single isotropic activation function, but on the general form. Furthermore, the competitiveness of this form cannot be ascertained from any single function or application, since it represents a broad functional form whose inherent performance relative to elementwise formulae is currently unknown. This is especially evident in the incomparability of even superficially analogous functions like isotropic-tanh and standard-tanh in the experiments of \textit{App.}~\ref{App:ExperimentTwo} showing markedly different spectra-learning dynamics. Moreover, the chosen isotropic-tanh is merely a superficial analogue of standard tanh (itself not typically optimal), intended as a provisional placeholder that saturates below 1 to prevent the need for normalisations that introduce confounders. However, because neuroadaptation depends only on isotropic forms, the verification empirics did not require the function to be optimal; isotropic-tanh was sufficient to confirm the underlying mechanisms. Nevertheless, the results of \textit{App.}~\ref{App:ExperimentSix} are encouraging, showing that despite this placeholder, it typically outperforms standard-tanh for deeper and wider MLP architectures on more challenging visual-classification datasets.
    
    Similarly, no \textit{general} universal approximation theorem (UAT) exists for this form, which does not imply its absence either. Despite this, the pragmatic usage of elementwise formulae proceeded for the intermediate $31$ years between Rosenblatt's canonical work \cite{Rosenblatt1958} and Cybenko's first universal approximation theorem \cite{Cybenko1989}, alongside their underappreciated specificity to particular architectures. Thus, this unknown is not immediately discouraging; additionally, precedent exists for incidental normed nonlinearities \cite{Sabour2017, Bird2025c} with motivation outside a symmetry-prescription formalism (not to be confused with radial-basis functions \cite{Broomhead1988}). Furthermore, a specific-function, 2-layer UAT has been demonstrated for the `step-ReLU' radial function \cite{Ganev2022}, which, under our terminology, is an instance of an isotropic function. Regardless, in the worst-case scenario, the newly enabled behaviours may, in several circumstances, outweigh mild accuracy suboptimality. To optimise for competitiveness or production, we motivate a substantial search over the class of isotropic functions; this remains a limitation for future research, given that the underlying mechanisms are now substantiated.

    This also relates to the weight-parameterisations and redundancies therein. Under the standard permutation-prescription, the unique weight-space is $\mathbb{R}^{n\times m}/ S_n$ in shorthand, with a generic group size of $\left|S_n\right|=n!$ multiplying the functionally equivalent parameterisations. Further reduction occurs for hyperoctahedral-prescribed activation functions to $\mathbb{R}^{n\times m}/ B_n$, in shorthand, with group size $\left|B_n\right|=2^{n}n!$ multiplying unique parameterisations\footnote{These should not be confused with the two-sided closures $B_n$, $B_m$, or $S_n$, $S_m$ or $O\left(n\right)$, $O\left(m\right)$ also applicable, but would reduce parameterisation for every other layer, instead one-side closures are detailed which are relevant for \textit{every} layer except the output.}.
    Whether this retains expressibility, given increasingly redundant parameterisations, remains unclear. Parameter count and function-class expressivity are fundamentally different quantities. Orthogonal-prescribed multilayer perceptrons further reduce the number of effective, unique parameters, as they are a supergroup, $S_n\subset B_n\subset \mathrm{O}\left(n\right)$, yielding a continuous orbit of reparameterisations. Pursuing this to larger supergroups eventually yields $\operatorname{GL}_n(\mathbb{R})$, where the induced `linear activation functions' collapse expressibility to linear regression; therefore, expressibility is compromised in the extreme case.
    
    Whether these degeneracies and reduced unique parameterisations are harmful or helpful for expressibility is presently unclear. Such considerations may, reductionistically, come down to whether better minima emerge along what would be the orthogonal orbit, but when using activations described maximally by typical elementwise subgroups, e.g. $S_n$, $B_n$, etc. With these, a potentially \textit{usefully} nuanced loss structure arises because of the absence of continuous redundancy, whereas such minima wouldn't arise for the flattened-loss orbits induced by orthogonal-group prescriptions. Conversely, such local minima could be troublesome for optimisation if unfavourable, whereas a flattened orbit may offer better learning efficiency by avoiding them, since it has a zero gradient in the orbit direction, reducing exploration of the continuous orbit-flattened loss. This appears highly specific to the particularities of the seemingly non-comparable, non-linear relationship of the two functional forms. Some results do suggest a parameter-symmetry dependence on training phenomena \cite{Lim2024}.
    
    Forming a canonical parameterisation may be more intuitive. We propose the following directionally, assuming that $0$ singular-valued neurons are pruned and that positive values are distinct, which is likely after training. An example is the partial-diagonal representations for all but the final layer $\mathbf{R}\mathbf{\Lambda}\in \mathbb{R}^{n\times m}$, where $\mathbf{R}$ has degrees of freedom $0.5n(n-1)$ from its Lie-algebra parameterisation, and magnitude-ordered singular values $\mathbf{\Lambda}$ with $\min\left(n,m\right)$ degrees of freedom --- these can also be transformed to have $\det{\mathbf{R}}=+1$. However, a sign-flip redundancy remains in $\mathbf{R}$ despite the positive determinant, and subspace redundancy remains when $m<n$. Further conventions can be explored to reduce this ambiguity. A benefit is that, if a network can be put into a unique canonical form, we could explore an intuitive sense of `model-distance' --- though this may still be possible despite the subspace and column-signed ambiguities, by taking such an inter-model quasi-distance in the Lie-Algebra coordinates (given you restrict them to a non-repeating domain).

    
    A further limitation is that neurodegeneration is, in general, unavoidably approximate in its function invariance. This is because, unlike neurogenesis, one cannot \textit{generally} embed the original parameters, $\operatorname{rank}\left(\mathbf{W}'\right)\leq\operatorname{rank}\left(\mathbf{W}\right)$ --- non-zero singular values are removed. However, this motivated several compensatory parameters, the intrinsic length and psi-vector, which were shown (\textit{Tab.}~\ref{Tab:CaltechPresentable} and \textit{App.}~\ref{App:ExperimentOne}). Using $\vec{\psi}$ induces an ordering from final-to-first-layer neuroadaptation, ensuring psi-vector compensation propagates to only one subsequent layer. For the same reason, this $\vec{\psi}$-parameter should decay before the next neuroadaptation, which may also help mitigate non-linear instability in its action (although baseline percentages of \textit{Tabs.}~\ref{Tab:Caltech256_KL}, \ref{Tab:CIFAR10_KL} and \ref{Tab:CIFAR100_KL} suggest this is inconsequential as they remain comparable).
    
    Furthermore, with a saturating isotropic activation function, or with the discussed normalisation, activation magnitudes cannot grow to compensate for a decaying singular value --- thus reliably implying that such singular values have little impact on overall model functionality. Hence, despite the inherent limitations and function invariance during neurodegeneration, the methodology presented does take several steps to mitigate this function drift --- confirmed then empirically in \textit{Tab.}~\ref{Tab:CaltechPresentable} results. 
    
    Similarly, the results of \textit{App.}~\ref{App:NullLike} implies that neurogenesis initialisation should not be zero in $\vec{W}^{+}$ due to inducing an optimisation fixed-point, and $b^{+}=0$ is a necessity to ensure true function invariance, as otherwise the subsequent biases would need a non-linear correction. This produces an explicit symmetry break at initialisation, biasing the network along a particular optimisation trajectory. As discussed, these explicit symmetry-breaking initialisations can be explored to achieve various effects. Neither is explicit symmetry breaking inherently problematic, as evidenced by the many initialisers in standard use. 


    It is also essential to clarify that several claims hold only asymptotically; for example, the $50\%$ sparsification figure holds only in the infinite-width and depth limit, while the finite equations are highly architecture-specific to constant-width isotropic multilayer perceptrons. Considerations also pertain to isotropic nested functional classes, which apply only to equal- or monotonically increasing-width isotropic MLPs in the regime where the final activation function becomes 'saturated' (constant in its non-linear term). Nevertheless, several isotropic functions, including isotropic-tanh, do display such saturating behaviour, particularly when considering the positive-definite intrinsic length. Further considerable sparsification was demonstrated, \textit{App.}~\ref{App:ExperimentFive}, for finite width and depth architectures, consistent with \textit{Eqns.}~\ref{Eqn:OddLength2}~and~\ref{Eqn:EvenLength2}.

    Other considerations, such as hyperparameter optimisation, SVD-scheduling, initialisations, isotropic function-searches and thresholding alternatives, could be explored if one wished to demonstrate competitive performance beyond validating the theoretical claims prioritised in this paper. Importantly, the current neuroadaptation approach was shown to add negligible compute time compared to standard training and test-set evaluation, showing that the method's practical cost is dominated by underlying model training rather than the adaptation procedure itself. Thus, one may also increase the frequency of neuroadaptation beyond once per epoch with little ramification on compute time, and alternative variations should be mindful of efficiency.

    Further limitations regarding diagonalisation approximations for rank-2 quantities, as used in existing optimisers and normalisers, can also affect invariance in the optimisation trajectory and forward-pass inference, respectively. However, they can be approximately transformed to explicitly compensate; moreover, the concern is mitigated by the likely rotationally degenerate parameterisations being explored regardless through training, with little impact observed thus far on such quantities. 
    
    Finally, the empirical component is intended as a mechanism verification rather than a competitive or SOTA-scale implementation. This reflects the paper's primarily theoretical-conceptual contribution: general isotropic mechanisms for structural invariance that support a broader symmetry-prescriptive ontology. Building on this foundation, orthogonal-prescribed primitives were used to admit orthogonal reparameterisations, enabling behaviours including basis independence, SVD-based diagonalisation, sparsification and function-preserving neuroadaptation. These analytical results therefore require empirical verification first; the contribution lies primarily in these novel attributes afforded by a symmetry-prescriptive ontology, rather than in any single optimised experimental procedure. Alongside the restriction to multilayer perceptrons, limiting SOTA possibilities. 
    
    Accordingly, the experiments focus on invariance, adaptation dynamics, and computational overhead as the most direct, meaningful tests of the proposed mechanism. The experiments are therefore not presented as evidence of competitiveness but as controlled, minimalist tests of whether the proposed symmetry mechanism behaves as predicted, while minimising confounders to the mechanism. Large-scale competitive benchmarks would introduce architecture choice, activation-function optimality, optimiser interactions, normalisation, data augmentation, and more as confounders; hence, these minimal mechanism-verification experiments were preferred. Moreover, comparing elementwise neuroadaptive approaches to our isotropic formulation introduces significant confounders again, stemming from the markedly different learning dynamics expressed by the differing functional forms. One cannot determine whether differences result from the neuroadaptive method or the underlying functions. This cannot be rectified by reformulating prior techniques onto isotropic primitives or the proposed ones onto elementwise forms, as each typically \cite{Chen2016, Ruso2016, Yoon2018, Kim2020} presupposes and is dependent on its respective form.
    
    Architecturally, one may consider whether elementwise functions have circularly benefited current model constructions. Transformer or convolutional systems may themselves be predicated on primitive choices.
    Thus, given an altered symmetry-prescribed foundation, unknown selection confounders may exist in more elaborate constructions, encouraging the straightforward verification on multilayer perceptrons. Overall, results with placeholder activation functions should therefore be interpreted as verification of the mechanism, not as assessments of the broader functional class enabled by symmetry-prescribed primitives. Overall, establishing novel mechanistic possibilities upfront, before refining every component for competitive optimality, is preferable and enables testing of the paper's core claims. However, substantial future work is required to determine the exact pipeline optimisation, hyperparameter search and activation function selection needed for competitive SOTA comparisons.

\section{Related Works\label{App:Related}}
    
    The works of `Net2Net' \cite{Chen2016}, particularly `Net2WiderNet', and `Network Morphism' \cite{Wei2016}, display some relation to this work in terms of neurogenesis considerations. These works effectively propose an approach to embed a narrower network architecture within a wider one, while preserving function. The observation in the Net2WiderNet work is that for an elementwise activation function, one may arbitrarily duplicate an existing column vector to widen a layer’s weight matrix, shown in \textit{Eqn.}~\ref{Eqn:Net2WiderNet1} with $\vec{W}_{(1)}\in\operatorname{column}\left(\mathbf{W}_{(1)}\right)$, provided that the corresponding outgoing weights in the subsequent layer are divided among the duplicated units so that their sum equals the original outgoing weight, shown on the right of \textit{Eqn.}~\ref{Eqn:Net2WiderNet1} with $\vec{C}_{\text{duplicity}}$ being a vector correcting for duplicity. The appended vectors are functions of their corresponding weights: $\vec{W}^{+}_{(i)}\left(\mathbf{W}_{(i)}\right)$.

    \begin{align}
        \mathbf{W}'_{(1)}=\left(\mathbf{W}_{(1)}, \vec{W}^{+}_{(1)}\left(\mathbf{W}_{(1)}\right)\right) &&\mathbf{W}'_{(2)}=\vec{C}_{\text{duplicity}}\odot\begin{pmatrix}\mathbf{W}_{(2)}\\\vec{W}^{+}_{(2)}\left(\mathbf{W}_{(2)}\right)\end{pmatrix}
        \label{Eqn:Net2WiderNet1}
    \end{align}

    This is arguably a form of neurogenesis for elementwise activations that increases a layer's neuron count, but is otherwise distinct from the singular-value approach enabled by isotropic formulations, discussed in this paper. As a result, embeddings differ considerably; for example, the SVD approach appends a zero column to the singular values and then makes an unconstrained choice in the corresponding row of the subsequent layer. This is achieved only by exploiting the reparameterisation symmetry prescriptively introduced into primitives for this purpose. Additionally, neurodegeneration is not covered by the Net2Net formulations, and their neurogenesis requires some function-invariance noise to break the optimisation symmetry and drive neural differentiation --- this noise results in a small error in function preservation. Our approach using isotropic primitives does not require this function-invariance breaking noise: the explicit symmetry breaking introduced in this paper is qualitatively different, preserving the function exactly via null-like reparameterisations whilst desirably breaking the optimiser's symmetry. It would appear that near-zero noisy initialisations may also be possible within the Net2WiderNet framework. Network morphism addresses a problem similar to Net2Net's, but within a more general framework using non-identity factorisations.

    Overall, although they share the aim of increasing a network's width, these approaches differ significantly from this paper's in how they achieve function invariance. Further, the results of \textit{App.}~\ref{App:Experiments} and \citet{Bird2025c} suggests an incomparability of functional forms even for superficially similar non-linear terms, with Net2WiderNet contingent on elementwise formulae, and this work is centred on symmetry-led isotropic formulae. However, both formulations demonstrate function-invariance under neurogenesis.
    
    Additionally, some techniques from Net2DeeperNet and Network Morphism may be combined with the present work by parameterising the isotropy $\mathbf{f}_{\ast}\left(\vec{x}; \alpha\right)=\alpha\vec{x}+(1-\alpha)\mathbf{f}\left(\vec{x}\right)$, $\alpha\in[0,1]$ in order to insert new layers. Alternatively, one may exploit the local linearity about the origin $\left.\mathbf{J}_{\mathbf{f}}\right|_{\vec{0}}\propto\mathrm{I}_{n\times n}$ of isotropic activation functions to similar effect, given a normalisation that guarantees the magnitudes will not enter a highly non-linear region of the activation function. Thus, overlap could be established through Net2DeeperNet.

    Rigging-The-Lottery (RigL) \cite{Evci2021} more clearly draws a distinction with the present work. Motivationally, RigL responds to the Lottery-Ticket hypothesis \cite{Frankle2019}. This hypothesis posits that a sparse network found through pruning can match or improve the original network by training that sparse architecture alone from its original initialisation; RigL demonstrates the performance of a training-time sparsely adaptive network. Thus, Rigging-the-Lottery focuses on training networks with sparse representations from the start and progressively performing further connectivity pruning or re-enabling connectivity throughout learning. These actions are based on criteria such as comparison to dense-network gradients for growth and weight-magnitude-based pruning. Thus, this is an example of adaptable connectivity rather than neuroadaptability introduced in this paper; the dense network's width remains static. This connectivity pruning does change the directionality of the weight-matrix columns, unlike a singular-value thresholding-of-importance approach, which may be undesirable. With such a singular-value approach, these weight columns have a single non-zero entry; thus, decay or removal does not affect directionality, while it also makes connectivity pruning equivalent to full neurodegeneration. Furthermore, this paper's diagonalised representation yields a significant $1-\min\left(m^{-1},n^{-1}\right)$ connectivity sparsity for a single weight matrix with an identically invariant function, which experiments could explore in terms of the lottery-ticket hypothesis.
     
    Together with the Single-Shot Network Pruning (SNIP) \cite{Lee2019} (also a connectivity-pruning technique), these papers also indicate an alternative thresholding approach, namely loss sensitivity, which could be similarly explored for this paper's neurodegeneration pipeline --- and is directionally discussed in \textit{App.}~\ref{App:Thresholding}. 
    
    Independently, neuron merging \cite{Kim2020} provides a useful comparison on neurodegeneration. This approach uses a weight `factorisation/decomposition', which retrospectively may be considered to leverage a scaled-permutation-described network. It does this for neurodegeneration by attempting to compensate for functional loss by merging the removed neuron's contribution into a retained, similar neuron. Together with a strong ReLU constraint, it effectively uses an (unstated) positive-scaled permutation (positive-monomial) equivariance exhibited by ReLU \cite{Godfrey2023} to propagate such scalings through the activation function. 
    
    The decomposition leverages the interdependence among the weight column vectors to reexpress the removed (individuated) neurons as a scaling of a single remaining column vector (not a general linear combination), using cosine similarity for comparability. However, it is important to consider that this is not an algebraic weight-factorisation like SVD; it is an approximate nearest-retained-neuron compensation rule that presupposes column-wise redundancy for good approximation, distinct from a spectral ranking of function contributions. Thus, a strong approximation enters earlier than the neurodegenerative approximation generally requires, especially compared to the present isotropic-SVD approach.
    
    We argue that the singular-value structure also provides a clearer proxy for the importance of (diagonalised-)neurons than columnwise comparisons, as the former, combined with bounded-norm activations, indicates the functional impact of neurodegeneration. This contrasts with cosine similarity and the separate pruning-criterion heuristics required by the individuated-neuron constraint, which only roughly indicate a neuron's functional redundancy among existing neurons.
    
    The isotropic constraint in this work is also a broad functional-form constraint rather than a specific-instance constraint like elementwise-ReLU in the neuron-merging work. However, this may be generalised retrospectively to positive-monomial-prescribed primitives for this paper, with ReLU as the practical and commonplace example. Overall, this approach has similarities but is explicitly limited to the elementwise-ReLU case and is restrictive in that removed weight columns must be approximately redundant. Thus requiring near-collinearity with \textit{a} preserved column vector. However, viewed retrospectively and alongside Net2WiderNet's work on neurogenesis, one could argue that this combination enables a form of neuroadaptability in networks described by scaled permutation equivariance (ReLU multilayer perceptrons). 

    The works of ``Elastic Weight Consolidation'' \cite{Kirkpatrick2017} (EWC), ``Progressive Neural Networks'' \cite{Ruso2016} (PNN)  and ``Dynamically Expandable Networks'' \cite{Yoon2018} (DEN) can also be contextualised in reference to the presented approach. The former is motivated by mammalian dendritic spine preservation during learning of subsequent tasks, this is emulated by `anchoring' parameters to a low-loss region for the previous task through use of a quadratic loss term multiplied by the diagonal of the Fisher matrix --- analogous to a physical elastic-spring system with parameter-wise varying stiffness coefficients. Principally, this approach addresses this different `catastrophic forgetting problem' \cite{McCloskey1989, Ratcliff1990, French1999}, which is not the focus of our work. However, the EWC approach could be generalised to isotropic networks with relative ease, since the underlying Fisher-quadratic term treatment remains applicable, but it is basis-dependent, requiring consideration and at least appropriate transforms. The diagonal Fisher approximation prevents the latter exactly, but may be approximated via \textit{App.}~\ref{App:GradientCoupling}'s approximate rank-2 basis-change, or transforming the full Fisher prior to diagonal approximation, or a full recomputation of approximate Fisher diagonal after a basis-change. Progressive neural networks approach continual learning by adding subnetworks for new tasks (a neurogenesis-like approach) while freezing already converged parameters. Lateral `adapters' are used to permit transfer learning from prior trained subnetworks to the newly appended ones if useful. This differs considerably from the isotropic approach, as PNNs presuppose elementwise independence of the activations when preserving function. Isotropy would re-couple these frozen columns through the non-linearity, as affine layers are no longer the only neuron-mixing operations present; thus, function-independence is infeasible without separately reapplying isotropic non-linearities with blockwise-orthogonal equivariance, preventing full SVD diagonalisations. The blockwise isotropic application also reduces to odd-symmetric elementwise activation functions in a single-column-width case.

    Dynamically expandable networks consider these various approaches and generalise from them: sparse connectivity through regularisation; preserving existing network functionality through selective retraining and frozen parameters alongside an EWC-like regularisation term (an $L_2$ rather than Fisher-scaled quadratic term); adding a variable number of neurons based on a loss-based criterion; or split-duplication where drift exceeds a threshold. Together, this ensemble of actions produces a dynamic programme for neurogenesis, mitigating perceived oversights in prior approaches. Similar considerations regarding presupposition of neuron-independence and functionality freezing prevent a naive isotropic generalisation. Thus, isotropic-adaptive and dynamically expandable networks are contingent on their respective activation forms for intended function, with neurogenesis mechanisms that trigger on considerably different metrics. Thresholded split-duplication cannot be performed on diagonalised neurons, because the SVD likely produces different orthogonal transforms as training evolves, making differently timed diagonalised neurons incomparable. This largely stems from neuron indistinguishability and basis freedom. However, group sparsity (not a group-theoretic approach, despite $g\in\mathcal{G}$ notation; more set-theoretic) is drastically simpler in the diagonalised isotropic picture, as the one-to-one connectivity makes classic $L_1$ (or $L_2$) per singular value equivalent to the neuron-wise group sparsity used in \citet{Yoon2018} (and nuclear norm) --- thus, is comparable to the decay we used. 

    Overall, comparing these continual learning mechanisms directly requires isolating each mechanism from significant confounders, but fundamental differences prevent this. Substituting isotropic forms into either of the latter methods breaks their respective function-preserving mechanisms, similar for Net2Net, Network Morphism and Neuron Merging --- while EWC addresses a distinct problem. Each method implicitly relies on elementwise formulae, or specific instances thereof, and/or the independence brought by individuated neurons to restructure the architecture without function interference; however, our reframing removes this requirement and replaces it with isotropic forms. In the contrary case: running each method with its own native activation form avoids this failure, but introduces a significant confound: even superficially similar functions like standard- and isotropic-tanh learn in inherently disparate ways, even in static networks shown by diverse spectral evolutions of \textit{App.}~\ref{App:ExperimentTwo}. This activation form confound is inseparable from the continual learning algorithms built upon it. Thus, taken together, a suitable ablation strategy is needed to isolate continual learning algorithms from the activation forms on which they depend, but it is unclear how this can be undertaken whilst faithfully presenting each method within its intended regime. Thus, this discussion provides structural and mechanistic comparisons between the various approaches. 

    Finally, what we term `isotropic activation functions' has precedent under the `radial rescaling activations' \cite{Ganev2022} term, which should not be conflated with the historical and contemporary prevalence of `radial basis functions' \cite{Broomhead1988} or $\text{U}\left(1\right)^n\rtimes S_n$ (elementwise complex-norm) permutation and independent-phase (unitary) equivariant C-ReLU functions \cite{Worrall2017} (acting more like complexified elementwise activation functions). \citet{Ganev2022} independently \cite{Bird2026Provenance, Bird2025a, Bird2025b, Bird2025c} recognised this form, and analogously its basis-freedom, under a rather different QR-decomposition-based approach. This compression approach leverages the distinct QR decomposition to reduce the width of subsequent layers, by also exploiting the orthogonal reparameterisation. Hence, the network is analytically function invariant to this as well. For a preceding width $n$, this decomposition effectively finds a transform that embeds the subspace activations can reach under the affine map's image into the first $n+1$ neurons of the subsequent layer---thus the network is restructured to grow layer-by-layer by at most one neuron. The additional $+1$ arises from considering the bias-term $(\mathbf{W}, \vec{b})\times(\vec{x},1)^T$. Comparable compression remains possible via our distinct SVD-based approach; however, this was strictly not undertaken due to gradient inequivalence and the importance of the buffer of so-called `scaffold neurons' we wished to retain with small (potentially zero or non-zero) singular values. The derivation of \textit{projected} gradient descent \cite{Ganev2022} equivalence does not alleviate concerns about scaffold neurons, as projected formulations prevent their update and specialisation. Near-but-not-zero singular values are also not directly exposed through QR decomposition either; they cannot be identified solely from this decomposition. This inhibits our neurodegenerative approach, which requires these. However, because they have distinct remits, we highlight an important consequence of combining both independent pursuits: QR compression followed by SVD sparsification---these are not mutually exclusive actions. 
    
    One can consider an amplified inference-time compression by diagonalising the compressed model. For $m+1<n$, weights of $\mathbf{W}\in\mathbb{R}^{n\times m}$ then QR-compression yields a matrix of $\mathbf{W}\in\mathbb{R}^{m+1 \times m}$, which SVD-sparsification (on alternating layers) then sparsifies to $\vec{W}\in\mathbb{R}^m$, thus yielding a connectivity reduction of $1-n^{-1}$ instead of $1-m/n-1/n$ or $1-\min(m^{-1},n^{-1})$ for QR-compression or SVD-diagonalisation respectively. However, since the $m+1<n$ case yields $n^{-1}=\min(m^{-1},n^{-1})$, this offers no improvement over SVD alone when the criteria for QR compression apply. Intuitively, and simplifying to the biasless case, this is because SVD's rectangular diagonal matrix zeros out rows beyond the square diagonal, which QR compression explicitly removes anyway; hence, the main difference may be the full-matrix footprint when used in combination. However, we note that one could store only the SVD's diagonal vector rather than the full matrix, which reduces the matrix footprint further and more than QR compression. SVD-diagonalisation does not have such criteria either, so it may be applied regardless on alternating layers; yet, QR-compression could meaningfully reduce footprint and computation on the intervening layers left dense by the alternating SVD-diagonalisation procedure.

    The authors also present a special-case, bounded-width universal approximation theorem to accompany this result, which they explicitly acknowledge as restricted to the Step-ReLU case. Despite this special-case regime, it nevertheless provides important insights into the expressibility of such forms and potentially broader generalisations. 
    
    Overall, alongside distinctly motivated specific cases \cite{Sabour2017}, instances of mutual convergence on this functional form have occurred. Yet across these, distinct considerations, consequences, architectures \cite{Sabour2017} and behaviours have arisen. These fruitful directions encourage analysis of this form in particular and the broader, symmetry-led primitives and possibilities available, without presupposing neuron-like historical constructions or global model adherence. This need not be limited to activation functions either, but can span normalisers, optimisers, initialisers, convolution, operations and may extend to residuals, attention, and more. It is this symmetry-prescriptive use at the level of individual primitives, extending beyond compositional model-scale or reparameterisation considerations, to induce useful or potentially confounding \cite{Bird2025c} biases, behaviours and benefits, that this approach and taxonomy emphasise as a potentially productive avenue for generalisation. Such symmetry consequences may arise individually or compositionally (symmetry making and breaking), while the taxonomy may helpfully organise these consequences and associate them with their underlying symmetry roots.

    Returning to neuroadaptation, the isotropic approach enables both considerable connectivity pruning through sparsification and neurogenesis, and is complemented by full neurodegeneration. Thus, the three typically separated constructions: connectivity-pruning, neurogenesis and neurodegeneration, are unusually unified in this approach, unlike the discussed works. Prior work recovers only aspects of this full regime, limited to neurogenesis, neurodegeneration or connectivity sparsity, rather than a fully integrated and cohesive approach. Moreover, these prior neuroadaptive works are limited by neuron individuation, identified as a significant hurdle to network adaptation due to the high interconnectivity among individuated neurons, which is not present among diagonalised neurons. This paper's approach to reformulating primitives through orthogonal, equivariant prescriptions to deindividuate neurons enables a broader use of decomposition tools, most notably singular-value decomposition, to construct function-invariant diagonalised representations. This yields a basis for treating neurogenesis, connectivity sparsification, and neurodegeneration within a common framework, rather than as separate heuristic procedures. The latter two become equivalent for the diagonalised neurons. The various thresholdings enable arbitrarily well-approximated function invariances during neuroadaptation by exploiting their symmetry-principled design.

    Consequently, direct comparison with elementwise approaches is limited: they operate within the inherited symmetry structure of conventional neuron-indexed networks, whereas the present work alters the primitive symmetry class itself, making adaptive topology a consequence of the network’s reparameterisation structure --- and incurring considerably different computational approaches, shown in \textit{App.}~\ref{App:ExperimentTwo}, further limiting comparison.


    
\newpage
\section{Experimental Details\label{App:Experiments}}

    This section covers all experimental details used in the production of the plots and tables in \textit{Sec.}~\ref{Sec:Empirical}. It also includes additional verification results, such as sparsification invariance and spectra evolution of standard and isotropic-tanh networks of various widths. This paper primarily presents the theoretical developments underpinning such architectures, and the experiments are intended to substantiate those theoretical claims.

    Generic hyperparameters were used across all experiments to ensure consistency and that observed effects were attributable to the general theoretical formulation rather than to carefully selected parameter settings. These hyperparameters are as follows: learning-rate $\eta=1\times10^{-3}$, batch-size, $B_s=48$, orthogonal initialisation, all networks are trained with ADAM-w \cite{Loshchilov2017} with $L_2$ weight decay, $\lambda=1\times10^{-3}$ (otherwise PyTorch \cite{Paszke2019} defaults were used), additional psi-vector $L_2$ decay $\lambda_{\vec{psi}}=0.1$ (due to the fast decay to zero being an important consideration for the compensation mechanism), and elementwise standardisation of the input. We chose these hyperparameters as standard values and kept them fixed across training to ensure maximal comparability. Standardisation ensures that, elementwise, the samples have zero mean and a standard deviation of 1 across the dataset, unless all samples have an element equal to 0, in which case it remains 0. This approach was used across all experiments, and default train-test data splits were utilised ($80:20$ for Caltech256). Caltech256 also had all samples resized to $64\times64$ pixels across all experiments, which we judged a good balance: large enough to distinguish classes while remaining reasonably minimal. Cross-entropy loss was used across all classification experiments alongside isotropic-tanh, $\mathbf{f}\left(\vec{x}\right)=\tanh\left(\left\|\vec{x}\right\|_2\right)\hat{x}$, as the activation function in every case. All experiments report KL-divergence with natural-base logarithms, often termed `nats'.
    
    For neurogenesis, the appended scaffold neurons were initialised with outgoing weights as $\vec{W}^{+}\sim\mathcal{N}(\vec{0},0.02\times\mathbf{I})$, intended to provide a small symmetry-breaking perturbation as needed (\textit{App.}~\ref{App:NullLike}), while being sufficiently small such that this initial biased direction can be potentially overridden by optimisations. Other approaches to non-zero initialisation could have been explored, such as a linear combination of existing columns, but this was considered a stronger inductive bias. The number of independent repeats varied per experiment. All code used to produce the plots is available at \texttt{\url{https://github.com/GeorgeBird1/Isotropic-Adaptive-MLPs}} to enable reproduction. The code provides base seeds for many runs; however, we ran concurrent instances in tandem, each with distinct seeds to prevent duplicate samples. Experiments were conducted on a single NVIDIA RTX 3080 10GB GPU with 128 GB of DDR3 RAM.

    Architecture, number of epochs and dataset varied by experiment to show a breadth of networks in which the theoretical claims can be verified, and epochs varied due to the individual nature of the experiments.
    
\subsection{Experiment One: Function-Invariance to Neuroadaptation \label{App:ExperimentOne}}

    The objective of this experiment was to verify that adapting the width of a trained network can be achieved with minimal functional change, owing to the symmetry-led network construction. In particular, this invariance should hold without subsequent retraining following the architectural adaptation. The experiment explored several different approaches for the compensatory parameters of intrinsic length, $o$, and $\vec{\psi}$.
    
    For this verification experiment, we used Caltech256 classifiers on an isotropic multilayer perceptron architecture with the structure $\left[12288, 100, 100, 256\right]$. We trained each network for $50$ epochs across $20$ independent repeats per configuration. This section also provides identical experimental results but for CIFAR-10 and CIFAR-100, or network size $\left[3072, 100, 100, 10\right]$ and $\left[3072, 100, 100, 100\right]$, respectively.
    
    The columnwise configurations are as follows:

    \begin{enumerate}
        \item ``$o_{\text{none}}$, $\vec{\psi}_{\text{decay}}$'' consists of networks with no intrinsic length parameters, but they do include a trainable psi-vector compensatory parameter, which does have an $L_2$ decay term such that it tends to the zero vector. The stated algebra is used to transform psi under neuroadaptation.
        \item ``$o_{\text{constant}}$, $\vec{\psi}_{\text{decay}}$'' consists of networks with a constant, non-trainable, intrinsic length parameter, and they include a trainable psi-vector compensatory parameter, decaying to the zero vector. The stated algebra is used to transform intrinsic length and psi under neuroadaptation.
        \item ``$o_{\text{trainable}}$, $\vec{\psi}_{\text{decay}}$'' consists of networks with a trainable intrinsic length parameter and a trainable psi-vector compensatory parameter that decays to the zero vector. The stated algebra is used to transform intrinsic length and psi under neuroadaptation.
        \item ``$o_{\text{trainable}}$, $\vec{\psi}_{\text{none}}$'' consists of networks with a trainable intrinsic length parameter and no psi-vector compensatory parameter. The stated algebra is used to transform the intrinsic length.
        \item ``$o_{\text{trainable}}$, $\vec{\psi}_{\text{constant}}$'' consists of networks with a trainable intrinsic length parameter but a constant psi-vector compensatory parameter. The stated algebra is used to transform intrinsic length and psi under neuroadaptation.
        \item ``$o_{\text{trainable}}$, $\vec{\psi}_{\text{trainable}}$'' consists of networks with a trainable intrinsic length parameter and a trainable psi-vector compensatory parameter that does not decay to zero. The stated algebra is used to transform intrinsic length and psi under neuroadaptation.
        \item ``$o_{\text{trainable}}$, $\vec{b}_{\text{adjusted}}$'' consists of networks with a trainable intrinsic length parameter and no psi-vector compensatory parameter. The stated algebra is used to transform intrinsic length, and it uses the $\vec{b}'_2=\vec{b}_2+\mathbb{E}\left[\bar{\sigma}\left(\cdot\right)\right]\vec{W}_{\times}b_{\times}$ to transform the bias vector.
    \end{enumerate}

    Exploring all permutations of approaches to the psi-vector and intrinsic length would have required extensive compute, without yielding further meaningful insight or interpretation. Therefore, comparing intrinsic length approaches against a trainable psi-vector and comparing psi-vector approaches against a trainable intrinsic length provided the most informative combinations for practical implementation of this neuroadaptive methodology.

    These trained networks then undergo adaptation of their $\mathbb{R}^{100}\rightarrow\mathbb{R}^{100}$ affine map, \textit{in $k$ sequential steps} increasing or decreasing the width to the stated quantity, such as $+75$, yielding a map $\mathbb{R}^{100}\rightarrow\mathbb{R}^{175}$ or $-75$ yielding a map $\mathbb{R}^{100}\rightarrow\mathbb{R}^{25}$ --- no retraining is then undertaken, such that the invariance of the adaptation alone can be analysed. The row stating $0$ refers to just a layerwise partial diagonalisation, where the map retains the $\mathbb{R}^{100}\rightarrow\mathbb{R}^{100}$ dimensionality but is reparameterised accordingly.

    From this, we compute various metrics to compare before-and-after distributions of the final-layer activations. These are then averaged over all samples in the dataset, and then the final quoted mean and standard error were computed repeat-wise. The metrics utilised were relative-change (\textit{Eqn.}~\ref{Eqn:RelativeChange}), KL-divergence (\textit{Eqn.}~\ref{Eqn:KLmetric}), sum of singular values (\textit{Eqn.}~\ref{Eqn:SVchange}) and change in test-set accuracy. Relative change applies to the logits before softmax, $\vec{z}$, while KL-divergence is computed from elements $p^{(i)}$ of the vector after softmax, $\vec{p}$.

    \begin{equation}
        \epsilon_{\text{rel}}=\mathbb{E}_{(x,y)\sim D_\text{test}}\frac{\left\|\vec{z}_{\text{post}}-\vec{z}_{\text{pre}}\right\|_2}{\left\|\vec{z}_\text{pre}\right\|_2}\quad\times100[\%]
        \label{Eqn:RelativeChange}
    \end{equation}

    \begin{equation}
        D_{KL}=\mathbb{E}_{(x,y)\sim D_\text{test}}\sum_i p^{(i)}_{\text{pre}}\ln\left(\frac{p^{(i)}_{\text{pre}}}{p^{(i)}_{\text{post}}}\right)\quad[\text{nats}]
        \label{Eqn:KLmetric}
    \end{equation}

    \begin{equation}
        \Sigma=\sum_i^{\min(m,n)}\mathbf{\Lambda}_{ii}\quad\text{for}\quad \mathbf{U},\mathbf{\Lambda},\mathbf{V}=\operatorname{SVD}\left(\mathbf{W}\right)
        \label{Eqn:SVchange}
    \end{equation}

    These metrics were chosen as informative, robust measures of function invariance during neuroadaptation. KL-divergence indicates the overall distribution shift after adaptation compared to before, whilst relative-change indicates how much the logits shift as a percentage relative to their original scale. Singular values are non-negative, and their sums thus quantitatively measure how neuroadaptation changes the underlying affine transform's linear scaling through the weights. Test-set accuracy, also recorded, is likely the least informative, as more substantial shifts in logits may be required to cross a `decision boundary', but it still indicates model performance. This overlaps somewhat with relative-change as a coarser metric: sufficiently small $\epsilon_{\text{rel}}$ results in a negligible change in test accuracy, which may vary more discretely in steps.

    Baseline accuracies indicate how the various approaches to psi-vector and intrinsic-length compensatory parameters affected training over the $50$ epochs preceding neuroadaptation. They show only a marginal effect, suggesting their role is largely limited to the neuroadaptive compensations for which they were derived, rather than broader utility as training parameters.

    This section now presents the full tabulated results of experiment one, across all metrics and datasets (CIFAR-10, CIFAR-100, Caltech256) tested.

    Across the results (\textit{Tab.}~\ref{Tab:Caltech256_Erel}~to~\ref{Tab:CIFAR10_KL}), the broad pattern is consistent: the network is exactly function invariant under neurogenesis and diagonalisation/sparsification, whilst invariance is well approximated under modest neurodegeneration, which inevitably degrades with more substantial pruning due to the significant reduction in singular values available to manipulate the data usefully. Despite this, accuracy, particularly for CIFAR-10 (\textit{Tab.}~\ref{Tab:CIFAR10_KL}), remains more consistent.

    \begin{table}[htb]
    \caption{Experiment One (Caltech-256): singular-value structure and relative function change under neuroadaptation at layer $L=1$ of the $[12288, 100, 100, 256]$ classifier. Results are mean $\pm$ standard error over 20 independent repeats. Row $0$ applies diagonalisation only; $+k$ subsequently applies $k$ neurogenesis operations in a single step and $-k$ applies $k$ neurodegeneration operations in a single step. $\Sigma_0$ and $\Sigma_1$ denote the sums of singular values before and after neuroadaptation, respectively. Singular-value standard errors are omitted for legibility; the largest singular-value standard error is 0.38. $\epsilon_{\mathrm{rel}}$ is the \textit{Eqn.}~\ref{Eqn:RelativeChange}'s metric, bold indicates the smallest value within each neuroadaptation condition.}
    \label{Tab:Caltech256_Erel}
    \centering
    \scriptsize
    \setlength{\tabcolsep}{2.2pt}
    \renewcommand{\arraystretch}{1.12}
    \resizebox{\textwidth}{!}{%
    \begin{tabular}{r*{14}{c}}
    \toprule
    \text{Neuroadaptation} & \multicolumn{2}{c}{\shortstack{$o_{\mathrm{none}},\ \vec{\psi}_{\mathrm{decay}}$\\$\Sigma_0=240$}} & \multicolumn{2}{c}{\shortstack{$o_{\mathrm{constant}},\ \vec{\psi}_{\mathrm{decay}}$\\$\Sigma_0=240$}} & \multicolumn{2}{c}{\shortstack{$o_{\mathrm{trainable}},\ \vec{\psi}_{\mathrm{decay}}$\\$\Sigma_0=241$}} & \multicolumn{2}{c}{\shortstack{$o_{\mathrm{trainable}},\ \vec{\psi}_{\mathrm{none}}$\\$\Sigma_0=240$}} & \multicolumn{2}{c}{\shortstack{$o_{\mathrm{trainable}},\ \vec{\psi}_{\mathrm{constant}}$\\$\Sigma_0=237$}} & \multicolumn{2}{c}{\shortstack{$o_{\mathrm{trainable}},\ \vec{\psi}_{\mathrm{trainable}}$\\$\Sigma_0=237$}} & \multicolumn{2}{c}{\shortstack{$o_{\mathrm{trainable}},\ \vec{b}_{\mathrm{adjusted}}$\\$\Sigma_0=240$}} \\
    \cmidrule(lr){2-3}
    \cmidrule(lr){4-5}
    \cmidrule(lr){6-7}
    \cmidrule(lr){8-9}
    \cmidrule(lr){10-11}
    \cmidrule(lr){12-13}
    \cmidrule(lr){14-15}
     & $\Sigma_1$ & $\epsilon_{\mathrm{rel}}$ (\%) & $\Sigma_1$ & $\epsilon_{\mathrm{rel}}$ (\%) & $\Sigma_1$ & $\epsilon_{\mathrm{rel}}$ (\%) & $\Sigma_1$ & $\epsilon_{\mathrm{rel}}$ (\%) & $\Sigma_1$ & $\epsilon_{\mathrm{rel}}$ (\%) & $\Sigma_1$ & $\epsilon_{\mathrm{rel}}$ (\%) & $\Sigma_1$ & $\epsilon_{\mathrm{rel}}$ (\%) \\
    \midrule
    $+75$ & $240$ & $(8.4\!\pm\!0.2)\!\times\!10^{-4}$ & $240$ & $(8.5\!\pm\!0.2)\!\times\!10^{-4}$ & $241$ & $\mathbf{(8.2\!\pm\!0.2)\!\times\!10^{-4}}$ & $240$ & $(8.6\!\pm\!0.1)\!\times\!10^{-4}$ & $237$ & $(8.5\!\pm\!0.2)\!\times\!10^{-4}$ & $237$ & $(8.3\!\pm\!0.2)\!\times\!10^{-4}$ & $240$ & $(8.3\!\pm\!0.1)\!\times\!10^{-4}$ \\
    $+50$ & $240$ & $(8.4\!\pm\!0.2)\!\times\!10^{-4}$ & $240$ & $(8.5\!\pm\!0.2)\!\times\!10^{-4}$ & $241$ & $\mathbf{(8.2\!\pm\!0.2)\!\times\!10^{-4}}$ & $240$ & $(8.6\!\pm\!0.1)\!\times\!10^{-4}$ & $237$ & $(8.5\!\pm\!0.2)\!\times\!10^{-4}$ & $237$ & $(8.3\!\pm\!0.2)\!\times\!10^{-4}$ & $240$ & $(8.3\!\pm\!0.1)\!\times\!10^{-4}$ \\
    $+25$ & $240$ & $(8.4\!\pm\!0.2)\!\times\!10^{-4}$ & $240$ & $(8.5\!\pm\!0.2)\!\times\!10^{-4}$ & $241$ & $\mathbf{(8.2\!\pm\!0.2)\!\times\!10^{-4}}$ & $240$ & $(8.6\!\pm\!0.1)\!\times\!10^{-4}$ & $237$ & $(8.5\!\pm\!0.2)\!\times\!10^{-4}$ & $237$ & $(8.3\!\pm\!0.2)\!\times\!10^{-4}$ & $240$ & $(8.3\!\pm\!0.1)\!\times\!10^{-4}$ \\
    $+5$ & $240$ & $(8.4\!\pm\!0.2)\!\times\!10^{-4}$ & $240$ & $(8.5\!\pm\!0.2)\!\times\!10^{-4}$ & $241$ & $\mathbf{(8.2\!\pm\!0.2)\!\times\!10^{-4}}$ & $240$ & $(8.6\!\pm\!0.1)\!\times\!10^{-4}$ & $237$ & $(8.5\!\pm\!0.2)\!\times\!10^{-4}$ & $237$ & $(8.3\!\pm\!0.2)\!\times\!10^{-4}$ & $240$ & $(8.3\!\pm\!0.1)\!\times\!10^{-4}$ \\
    $+1$ & $240$ & $(8.4\!\pm\!0.2)\!\times\!10^{-4}$ & $240$ & $(8.5\!\pm\!0.2)\!\times\!10^{-4}$ & $241$ & $\mathbf{(8.2\!\pm\!0.2)\!\times\!10^{-4}}$ & $240$ & $(8.6\!\pm\!0.1)\!\times\!10^{-4}$ & $237$ & $(8.5\!\pm\!0.2)\!\times\!10^{-4}$ & $237$ & $(8.3\!\pm\!0.2)\!\times\!10^{-4}$ & $240$ & $(8.3\!\pm\!0.1)\!\times\!10^{-4}$ \\
    \addlinespace[1pt]
    $0$ & $240$ & $(8.4\!\pm\!0.2)\!\times\!10^{-4}$ & $240$ & $(8.5\!\pm\!0.2)\!\times\!10^{-4}$ & $241$ & $\mathbf{(8.2\!\pm\!0.2)\!\times\!10^{-4}}$ & $240$ & $(8.6\!\pm\!0.1)\!\times\!10^{-4}$ & $237$ & $(8.5\!\pm\!0.2)\!\times\!10^{-4}$ & $237$ & $(8.3\!\pm\!0.2)\!\times\!10^{-4}$ & $240$ & $(8.3\!\pm\!0.1)\!\times\!10^{-4}$ \\
    \addlinespace[1pt]
    $-1$ & $240$ & $(2.4\!\pm\!0.1)\!\times\!10^{-1}$ & $240$ & $(2.4\!\pm\!0.1)\!\times\!10^{-1}$ & $240$ & $(2.6\!\pm\!0.2)\!\times\!10^{-1}$ & $240$ & $\mathbf{(2.2\!\pm\!0.1)\!\times\!10^{-1}}$ & $237$ & $(2.7\!\pm\!0.1)\!\times\!10^{-1}$ & $236$ & $(2.6\!\pm\!0.1)\!\times\!10^{-1}$ & $240$ & $(2.3\!\pm\!0.2)\!\times\!10^{-1}$ \\
    $-5$ & $236$ & $(3.5\!\pm\!0.2)\!\times\!10^{-1}$ & $236$ & $(3.7\!\pm\!0.2)\!\times\!10^{-1}$ & $236$ & $(3.8\!\pm\!0.3)\!\times\!10^{-1}$ & $236$ & $\mathbf{(3.4\!\pm\!0.2)\!\times\!10^{-1}}$ & $233$ & $(6.7\!\pm\!0.2)\!\times\!10^{-1}$ & $233$ & $(6.4\!\pm\!0.2)\!\times\!10^{-1}$ & $236$ & $(3.7\!\pm\!0.3)\!\times\!10^{-1}$ \\
    $-25$ & $212$ & $(7.2\!\pm\!0.2)\!\times\!10^{-1}$ & $212$ & $(7.3\!\pm\!0.2)\!\times\!10^{-1}$ & $212$ & $(7.5\!\pm\!0.3)\!\times\!10^{-1}$ & $212$ & $(7.1\!\pm\!0.2)\!\times\!10^{-1}$ & $209$ & $(1.0\!\pm\!0.0)\!\times\!10^{0}$ & $209$ & $(9.9\!\pm\!0.2)\!\times\!10^{-1}$ & $212$ & $\mathbf{(7.1\!\pm\!0.3)\!\times\!10^{-1}}$ \\
    $-50$ & $170$ & $(3.7\!\pm\!0.1)\!\times\!10^{0}$ & $170$ & $(3.6\!\pm\!0.1)\!\times\!10^{0}$ & $170$ & $(3.9\!\pm\!0.2)\!\times\!10^{0}$ & $170$ & $\mathbf{(3.6\!\pm\!0.2)\!\times\!10^{0}}$ & $167$ & $(4.2\!\pm\!0.1)\!\times\!10^{0}$ & $167$ & $(4.1\!\pm\!0.2)\!\times\!10^{0}$ & $170$ & $(3.7\!\pm\!0.1)\!\times\!10^{0}$ \\
    $-75$ & $99$ & $(6.3\!\pm\!0.0)\!\times\!10^{+1}$ & $99$ & $(6.3\!\pm\!0.0)\!\times\!10^{+1}$ & $99$ & $(6.2\!\pm\!0.1)\!\times\!10^{+1}$ & $99$ & $(6.3\!\pm\!0.0)\!\times\!10^{+1}$ & $98$ & $(6.3\!\pm\!0.0)\!\times\!10^{+1}$ & $97$ & $\mathbf{(6.2\!\pm\!0.0)\!\times\!10^{+1}}$ & $99$ & $(6.3\!\pm\!0.0)\!\times\!10^{+1}$ \\
    \bottomrule
    \end{tabular}%
    }
\end{table}

    \begin{table}[htb]
    \caption{Experiment One (Caltech-256): additional function-change measures for the same neuroadaptation conditions. Results are mean $\pm$ standard error over 20 independent repeats. $D_{\mathrm{KL}}$ is provided by \textit{Eqn.}~\ref{Eqn:KLmetric} and $\Delta a=a_{\mathrm{post}}-a_{\mathrm{pre}}$ is the change in test accuracy in percentage points; positive values indicate increased accuracy. The baseline row reports the unadapted test accuracy $a_0$ as a percentage. Bold values indicate the smallest mean $D_{\mathrm{KL}}$ and the accuracy change closest to zero within each neuroadaptation condition.}
    \label{Tab:Caltech256_KL}
    \centering
    \scriptsize
    \setlength{\tabcolsep}{2.2pt}
    \renewcommand{\arraystretch}{1.12}
    \resizebox{\textwidth}{!}{%
    \begin{tabular}{r*{14}{c}}
    \toprule
    \text{Neuroadaptation} & \multicolumn{2}{c}{$o_{\mathrm{none}},\ \vec{\psi}_{\mathrm{decay}}$} & \multicolumn{2}{c}{$o_{\mathrm{constant}},\ \vec{\psi}_{\mathrm{decay}}$} & \multicolumn{2}{c}{$o_{\mathrm{trainable}},\ \vec{\psi}_{\mathrm{decay}}$} & \multicolumn{2}{c}{$o_{\mathrm{trainable}},\ \vec{\psi}_{\mathrm{none}}$} & \multicolumn{2}{c}{$o_{\mathrm{trainable}},\ \vec{\psi}_{\mathrm{constant}}$} & \multicolumn{2}{c}{$o_{\mathrm{trainable}},\ \vec{\psi}_{\mathrm{trainable}}$} & \multicolumn{2}{c}{$o_{\mathrm{trainable}},\ \vec{b}_{\mathrm{adjusted}}$} \\
    \cmidrule(lr){2-3}
    \cmidrule(lr){4-5}
    \cmidrule(lr){6-7}
    \cmidrule(lr){8-9}
    \cmidrule(lr){10-11}
    \cmidrule(lr){12-13}
    \cmidrule(lr){14-15}
     & $D_{\mathrm{KL}}$ & $\Delta a$ (pp) & $D_{\mathrm{KL}}$ & $\Delta a$ (pp) & $D_{\mathrm{KL}}$ & $\Delta a$ (pp) & $D_{\mathrm{KL}}$ & $\Delta a$ (pp) & $D_{\mathrm{KL}}$ & $\Delta a$ (pp) & $D_{\mathrm{KL}}$ & $\Delta a$ (pp) & $D_{\mathrm{KL}}$ & $\Delta a$ (pp) \\
    \midrule
    $+75$ & $\mathbf{(1.0\!\pm\!0.3)\!\times\!10^{-9}}$ & $\mathbf{0.0\!\pm\!0.0}$ & $(1.4\!\pm\!0.2)\!\times\!10^{-9}$ & $\mathbf{0.0\!\pm\!0.0}$ & $(1.4\!\pm\!0.3)\!\times\!10^{-9}$ & $\mathbf{0.0\!\pm\!0.0}$ & $(1.2\!\pm\!0.3)\!\times\!10^{-9}$ & $\mathbf{0.0\!\pm\!0.0}$ & $(1.2\!\pm\!0.3)\!\times\!10^{-9}$ & $\mathbf{0.0\!\pm\!0.0}$ & $(1.4\!\pm\!0.3)\!\times\!10^{-9}$ & $\mathbf{0.0\!\pm\!0.0}$ & $(1.5\!\pm\!0.3)\!\times\!10^{-9}$ & $\mathbf{0.0\!\pm\!0.0}$ \\
    $+50$ & $(1.3\!\pm\!0.3)\!\times\!10^{-9}$ & $\mathbf{0.0\!\pm\!0.0}$ & $\mathbf{(9.0\!\pm\!3.2)\!\times\!10^{-10}}$ & $\mathbf{0.0\!\pm\!0.0}$ & $(1.3\!\pm\!0.3)\!\times\!10^{-9}$ & $\mathbf{0.0\!\pm\!0.0}$ & $(1.3\!\pm\!0.3)\!\times\!10^{-9}$ & $\mathbf{0.0\!\pm\!0.0}$ & $(1.5\!\pm\!0.3)\!\times\!10^{-9}$ & $\mathbf{0.0\!\pm\!0.0}$ & $(1.2\!\pm\!0.3)\!\times\!10^{-9}$ & $\mathbf{0.0\!\pm\!0.0}$ & $(1.5\!\pm\!0.2)\!\times\!10^{-9}$ & $\mathbf{0.0\!\pm\!0.0}$ \\
    $+25$ & $(1.3\!\pm\!0.3)\!\times\!10^{-9}$ & $\mathbf{0.0\!\pm\!0.0}$ & $\mathbf{(1.1\!\pm\!0.3)\!\times\!10^{-9}}$ & $\mathbf{0.0\!\pm\!0.0}$ & $(1.3\!\pm\!0.3)\!\times\!10^{-9}$ & $\mathbf{0.0\!\pm\!0.0}$ & $(1.3\!\pm\!0.3)\!\times\!10^{-9}$ & $\mathbf{0.0\!\pm\!0.0}$ & $(1.5\!\pm\!0.3)\!\times\!10^{-9}$ & $\mathbf{0.0\!\pm\!0.0}$ & $(1.2\!\pm\!0.3)\!\times\!10^{-9}$ & $\mathbf{0.0\!\pm\!0.0}$ & $(1.5\!\pm\!0.2)\!\times\!10^{-9}$ & $\mathbf{0.0\!\pm\!0.0}$ \\
    $+5$ & $(1.3\!\pm\!0.3)\!\times\!10^{-9}$ & $\mathbf{0.0\!\pm\!0.0}$ & $\mathbf{(1.1\!\pm\!0.3)\!\times\!10^{-9}}$ & $\mathbf{0.0\!\pm\!0.0}$ & $(1.3\!\pm\!0.3)\!\times\!10^{-9}$ & $\mathbf{0.0\!\pm\!0.0}$ & $(1.3\!\pm\!0.3)\!\times\!10^{-9}$ & $\mathbf{0.0\!\pm\!0.0}$ & $(1.5\!\pm\!0.3)\!\times\!10^{-9}$ & $\mathbf{0.0\!\pm\!0.0}$ & $(1.2\!\pm\!0.3)\!\times\!10^{-9}$ & $\mathbf{0.0\!\pm\!0.0}$ & $(1.5\!\pm\!0.2)\!\times\!10^{-9}$ & $\mathbf{0.0\!\pm\!0.0}$ \\
    $+1$ & $(1.3\!\pm\!0.3)\!\times\!10^{-9}$ & $\mathbf{0.0\!\pm\!0.0}$ & $\mathbf{(1.1\!\pm\!0.3)\!\times\!10^{-9}}$ & $\mathbf{0.0\!\pm\!0.0}$ & $(1.2\!\pm\!0.3)\!\times\!10^{-9}$ & $\mathbf{0.0\!\pm\!0.0}$ & $(1.3\!\pm\!0.3)\!\times\!10^{-9}$ & $\mathbf{0.0\!\pm\!0.0}$ & $(1.7\!\pm\!0.3)\!\times\!10^{-9}$ & $\mathbf{0.0\!\pm\!0.0}$ & $(1.2\!\pm\!0.3)\!\times\!10^{-9}$ & $\mathbf{0.0\!\pm\!0.0}$ & $(1.5\!\pm\!0.2)\!\times\!10^{-9}$ & $\mathbf{0.0\!\pm\!0.0}$ \\
    \addlinespace[1pt]
    $0$ & $\mathbf{(6.9\!\pm\!2.9)\!\times\!10^{-10}}$ & $\mathbf{0.0\!\pm\!0.0}$ & $(1.2\!\pm\!0.2)\!\times\!10^{-9}$ & $\mathbf{0.0\!\pm\!0.0}$ & $(1.9\!\pm\!0.3)\!\times\!10^{-9}$ & $\mathbf{0.0\!\pm\!0.0}$ & $(8.0\!\pm\!3.0)\!\times\!10^{-10}$ & $\mathbf{0.0\!\pm\!0.0}$ & $(1.2\!\pm\!0.3)\!\times\!10^{-9}$ & $\mathbf{0.0\!\pm\!0.0}$ & $(1.5\!\pm\!0.3)\!\times\!10^{-9}$ & $\mathbf{0.0\!\pm\!0.0}$ & $(9.0\!\pm\!3.0)\!\times\!10^{-10}$ & $\mathbf{0.0\!\pm\!0.0}$ \\
    \addlinespace[1pt]
    $-1$ & $(1.6\!\pm\!0.2)\!\times\!10^{-5}$ & $\mathbf{0.0\!\pm\!0.0}$ & $(1.6\!\pm\!0.2)\!\times\!10^{-5}$ & $+0.0\!\pm\!0.0$ & $(2.0\!\pm\!0.4)\!\times\!10^{-5}$ & $+0.0\!\pm\!0.0$ & $\mathbf{(1.3\!\pm\!0.1)\!\times\!10^{-5}}$ & $-0.0\!\pm\!0.0$ & $(2.0\!\pm\!0.2)\!\times\!10^{-5}$ & $-0.0\!\pm\!0.0$ & $(1.9\!\pm\!0.2)\!\times\!10^{-5}$ & $-0.0\!\pm\!0.0$ & $(1.5\!\pm\!0.2)\!\times\!10^{-5}$ & $+0.0\!\pm\!0.0$ \\
    $-5$ & $(2.9\!\pm\!0.3)\!\times\!10^{-5}$ & $-0.0\!\pm\!0.0$ & $(3.3\!\pm\!0.4)\!\times\!10^{-5}$ & $-0.0\!\pm\!0.0$ & $(3.7\!\pm\!0.7)\!\times\!10^{-5}$ & $\mathbf{+0.0\!\pm\!0.0}$ & $\mathbf{(2.8\!\pm\!0.5)\!\times\!10^{-5}}$ & $-0.0\!\pm\!0.0$ & $(1.1\!\pm\!0.1)\!\times\!10^{-4}$ & $-0.0\!\pm\!0.0$ & $(10.0\!\pm\!0.6)\!\times\!10^{-5}$ & $-0.0\!\pm\!0.0$ & $(3.5\!\pm\!0.7)\!\times\!10^{-5}$ & $+0.0\!\pm\!0.0$ \\
    $-25$ & $(9.9\!\pm\!0.5)\!\times\!10^{-5}$ & $-0.0\!\pm\!0.0$ & $(1.0\!\pm\!0.1)\!\times\!10^{-4}$ & $-0.0\!\pm\!0.0$ & $(1.1\!\pm\!0.1)\!\times\!10^{-4}$ & $\mathbf{-0.0\!\pm\!0.0}$ & $\mathbf{(9.8\!\pm\!0.6)\!\times\!10^{-5}}$ & $-0.0\!\pm\!0.0$ & $(2.2\!\pm\!0.1)\!\times\!10^{-4}$ & $-0.1\!\pm\!0.0$ & $(2.0\!\pm\!0.1)\!\times\!10^{-4}$ & $-0.1\!\pm\!0.0$ & $(9.8\!\pm\!0.9)\!\times\!10^{-5}$ & $-0.0\!\pm\!0.0$ \\
    $-50$ & $\mathbf{(2.8\!\pm\!0.2)\!\times\!10^{-3}}$ & $-0.3\!\pm\!0.0$ & $(2.8\!\pm\!0.2)\!\times\!10^{-3}$ & $-0.3\!\pm\!0.0$ & $(3.4\!\pm\!0.4)\!\times\!10^{-3}$ & $-0.3\!\pm\!0.0$ & $(2.8\!\pm\!0.3)\!\times\!10^{-3}$ & $-0.3\!\pm\!0.0$ & $(3.8\!\pm\!0.3)\!\times\!10^{-3}$ & $\mathbf{-0.3\!\pm\!0.0}$ & $(3.6\!\pm\!0.4)\!\times\!10^{-3}$ & $-0.3\!\pm\!0.1$ & $(2.8\!\pm\!0.2)\!\times\!10^{-3}$ & $-0.3\!\pm\!0.0$ \\
    $-75$ & $(9.0\!\pm\!0.2)\!\times\!10^{-1}$ & $-11.9\!\pm\!1.0$ & $(9.0\!\pm\!0.1)\!\times\!10^{-1}$ & $-11.2\!\pm\!0.9$ & $(9.0\!\pm\!0.1)\!\times\!10^{-1}$ & $-11.4\!\pm\!1.0$ & $\mathbf{(8.6\!\pm\!0.1)\!\times\!10^{-1}}$ & $\mathbf{-11.0\!\pm\!0.9}$ & $(9.7\!\pm\!0.1)\!\times\!10^{-1}$ & $-11.8\!\pm\!1.1$ & $(9.4\!\pm\!0.1)\!\times\!10^{-1}$ & $-11.5\!\pm\!1.0$ & $(8.8\!\pm\!0.1)\!\times\!10^{-1}$ & $-11.8\!\pm\!1.0$ \\
    \midrule
    \text{Baseline} & -- & $20.6\!\pm\!2.2$ & -- & $19.4\!\pm\!2.1$ & -- & $19.4\!\pm\!2.1$ & -- & $19.4\!\pm\!2.1$ & -- & $19.7\!\pm\!2.2$ & -- & $19.7\!\pm\!2.2$ & -- & $20.4\!\pm\!2.2$ \\
    \bottomrule
    \end{tabular}%
    }
    \end{table}
    
    The Caltech256 neurogenesis KL-divergence is particularly informative, as shown in \textit{Tab.}~\ref{Tab:Caltech256_KL}, indicating values on order of $10^{-9}$ nats, which is a practically negligible change in function, confirming the exact function invariance desired. The comparability of these values to the diagonalisation row may suggest they reflect overall precision rather than a small error associated with neurogenesis. These small errors may reflect amplification of numerical errors by subsequent layers, or may stem from the precision of the SVD decomposition itself. At $-50$ neurons degenerated, the network shows a more modest change in function while retaining its classification accuracy, suggesting robustness to moderate degeneration even for Caltech256's many classifications. At a $75\%$ reduction in layer width, the invariance drops more drastically, due to the aggressive loss in singular values and their transforms critical to expressivity. However, even with aggressive neurodegeneration, the network continues to perform better than chance, suggesting the remaining learned transforms are still persistently useful rather than fully compromising utility. Separately, the column-wise comparisons indicate that while intrinsic-length helps a small but inconsistent amount in neurodegeneration, $\vec{\psi}$-corrections are modestly detrimental in this example.
    
    The results (\textit{Tab.}~\ref{Tab:CIFAR10_Erel}~and~\ref{Tab:CIFAR10_KL}) show these metrics on the CIFAR-10 experimental setup, documenting similar trends but greater indifference to test accuracy in neurodegeneration, likely due to the smaller number of classes and wider margins.

    \begin{table}[htb]
    \caption{Experiment One (CIFAR-10): singular-value structure and relative function change under neuroadaptation at layer $L=1$ of the $[3072, 100, 100, 10]$ classifier.  Results are mean $\pm$ standard error over 20 independent repeats. Row $0$ applies diagonalisation only; $+k$ subsequently applies $k$ neurogenesis operations in a single step and $-k$ applies $k$ neurodegeneration operations in a single step. $\Sigma_0$ and $\Sigma_1$ denote the sums of singular values before and after neuroadaptation, respectively. Singular-value standard errors are omitted for legibility; the largest singular-value standard error is 0.32. $\epsilon_{\mathrm{rel}}$ is the \textit{Eqn.}~\ref{Eqn:RelativeChange}'s metric, bold indicates the smallest value within each neuroadaptation condition.}
    \label{Tab:CIFAR10_Erel}
    \centering
    \scriptsize
    \setlength{\tabcolsep}{2.2pt}
    \renewcommand{\arraystretch}{1.12}
    \resizebox{\textwidth}{!}{%
    \begin{tabular}{r*{14}{c}}
    \toprule
    \text{Neuroadaptation} & \multicolumn{2}{c}{\shortstack{$o_{\mathrm{none}},\ \vec{\psi}_{\mathrm{decay}}$\\$\Sigma_0=174$}} & \multicolumn{2}{c}{\shortstack{$o_{\mathrm{constant}},\ \vec{\psi}_{\mathrm{decay}}$\\$\Sigma_0=174$}} & \multicolumn{2}{c}{\shortstack{$o_{\mathrm{trainable}},\ \vec{\psi}_{\mathrm{decay}}$\\$\Sigma_0=174$}} & \multicolumn{2}{c}{\shortstack{$o_{\mathrm{trainable}},\ \vec{\psi}_{\mathrm{none}}$\\$\Sigma_0=174$}} & \multicolumn{2}{c}{\shortstack{$o_{\mathrm{trainable}},\ \vec{\psi}_{\mathrm{constant}}$\\$\Sigma_0=164$}} & \multicolumn{2}{c}{\shortstack{$o_{\mathrm{trainable}},\ \vec{\psi}_{\mathrm{trainable}}$\\$\Sigma_0=164$}} & \multicolumn{2}{c}{\shortstack{$o_{\mathrm{trainable}},\ \vec{b}_{\mathrm{adjusted}}$\\$\Sigma_0=174$}} \\
    \cmidrule(lr){2-3}
    \cmidrule(lr){4-5}
    \cmidrule(lr){6-7}
    \cmidrule(lr){8-9}
    \cmidrule(lr){10-11}
    \cmidrule(lr){12-13}
    \cmidrule(lr){14-15}
     & $\Sigma_1$ & $\epsilon_{\mathrm{rel}}$ (\%) & $\Sigma_1$ & $\epsilon_{\mathrm{rel}}$ (\%) & $\Sigma_1$ & $\epsilon_{\mathrm{rel}}$ (\%) & $\Sigma_1$ & $\epsilon_{\mathrm{rel}}$ (\%) & $\Sigma_1$ & $\epsilon_{\mathrm{rel}}$ (\%) & $\Sigma_1$ & $\epsilon_{\mathrm{rel}}$ (\%) & $\Sigma_1$ & $\epsilon_{\mathrm{rel}}$ (\%) \\
    \midrule
    $+75$ & $174$ & $(3.1\!\pm\!0.1)\!\times\!10^{-3}$ & $174$ & $(2.9\!\pm\!0.1)\!\times\!10^{-3}$ & $174$ & $(2.9\!\pm\!0.1)\!\times\!10^{-3}$ & $174$ & $(3.0\!\pm\!0.1)\!\times\!10^{-3}$ & $164$ & $\mathbf{(2.5\!\pm\!0.1)\!\times\!10^{-3}}$ & $164$ & $(2.7\!\pm\!0.1)\!\times\!10^{-3}$ & $174$ & $(3.1\!\pm\!0.2)\!\times\!10^{-3}$ \\
    $+50$ & $174$ & $(3.1\!\pm\!0.1)\!\times\!10^{-3}$ & $174$ & $(2.9\!\pm\!0.1)\!\times\!10^{-3}$ & $174$ & $(2.9\!\pm\!0.1)\!\times\!10^{-3}$ & $174$ & $(3.0\!\pm\!0.1)\!\times\!10^{-3}$ & $164$ & $\mathbf{(2.5\!\pm\!0.1)\!\times\!10^{-3}}$ & $164$ & $(2.7\!\pm\!0.1)\!\times\!10^{-3}$ & $174$ & $(3.1\!\pm\!0.2)\!\times\!10^{-3}$ \\
    $+25$ & $174$ & $(3.1\!\pm\!0.1)\!\times\!10^{-3}$ & $174$ & $(2.9\!\pm\!0.1)\!\times\!10^{-3}$ & $174$ & $(2.9\!\pm\!0.1)\!\times\!10^{-3}$ & $174$ & $(3.0\!\pm\!0.1)\!\times\!10^{-3}$ & $164$ & $\mathbf{(2.5\!\pm\!0.1)\!\times\!10^{-3}}$ & $164$ & $(2.7\!\pm\!0.1)\!\times\!10^{-3}$ & $174$ & $(3.1\!\pm\!0.2)\!\times\!10^{-3}$ \\
    $+5$ & $174$ & $(3.1\!\pm\!0.1)\!\times\!10^{-3}$ & $174$ & $(2.9\!\pm\!0.1)\!\times\!10^{-3}$ & $174$ & $(2.9\!\pm\!0.1)\!\times\!10^{-3}$ & $174$ & $(3.0\!\pm\!0.1)\!\times\!10^{-3}$ & $164$ & $\mathbf{(2.5\!\pm\!0.1)\!\times\!10^{-3}}$ & $164$ & $(2.7\!\pm\!0.1)\!\times\!10^{-3}$ & $174$ & $(3.1\!\pm\!0.2)\!\times\!10^{-3}$ \\
    $+1$ & $174$ & $(3.1\!\pm\!0.1)\!\times\!10^{-3}$ & $174$ & $(2.9\!\pm\!0.1)\!\times\!10^{-3}$ & $174$ & $(2.9\!\pm\!0.1)\!\times\!10^{-3}$ & $174$ & $(3.0\!\pm\!0.1)\!\times\!10^{-3}$ & $164$ & $\mathbf{(2.5\!\pm\!0.1)\!\times\!10^{-3}}$ & $164$ & $(2.7\!\pm\!0.1)\!\times\!10^{-3}$ & $174$ & $(3.1\!\pm\!0.2)\!\times\!10^{-3}$ \\
    \addlinespace[1pt]
    $0$ & $174$ & $(3.1\!\pm\!0.1)\!\times\!10^{-3}$ & $174$ & $(2.9\!\pm\!0.1)\!\times\!10^{-3}$ & $174$ & $(2.9\!\pm\!0.1)\!\times\!10^{-3}$ & $174$ & $(3.0\!\pm\!0.1)\!\times\!10^{-3}$ & $164$ & $\mathbf{(2.5\!\pm\!0.1)\!\times\!10^{-3}}$ & $164$ & $(2.7\!\pm\!0.1)\!\times\!10^{-3}$ & $174$ & $(3.1\!\pm\!0.2)\!\times\!10^{-3}$ \\
    \addlinespace[1pt]
    $-1$ & $174$ & $(6.4\!\pm\!1.8)\!\times\!10^{-3}$ & $174$ & $(3.0\!\pm\!0.1)\!\times\!10^{-3}$ & $174$ & $\mathbf{(2.9\!\pm\!0.1)\!\times\!10^{-3}}$ & $174$ & $(1.9\!\pm\!0.5)\!\times\!10^{-2}$ & $164$ & $(3.9\!\pm\!0.4)\!\times\!10^{-3}$ & $164$ & $(3.7\!\pm\!0.3)\!\times\!10^{-3}$ & $174$ & $(7.9\!\pm\!1.7)\!\times\!10^{-3}$ \\
    $-5$ & $174$ & $(1.2\!\pm\!0.1)\!\times\!10^{-1}$ & $174$ & $(6.8\!\pm\!0.7)\!\times\!10^{-2}$ & $174$ & $(5.7\!\pm\!0.6)\!\times\!10^{-2}$ & $174$ & $(6.3\!\pm\!0.9)\!\times\!10^{-1}$ & $164$ & $(5.3\!\pm\!0.6)\!\times\!10^{-2}$ & $164$ & $\mathbf{(5.3\!\pm\!0.6)\!\times\!10^{-2}}$ & $174$ & $(1.7\!\pm\!0.2)\!\times\!10^{-1}$ \\
    $-25$ & $162$ & $(1.4\!\pm\!0.0)\!\times\!10^{0}$ & $162$ & $\mathbf{(1.2\!\pm\!0.0)\!\times\!10^{0}}$ & $162$ & $(1.2\!\pm\!0.0)\!\times\!10^{0}$ & $162$ & $(5.3\!\pm\!0.2)\!\times\!10^{0}$ & $152$ & $(1.4\!\pm\!0.0)\!\times\!10^{0}$ & $152$ & $(1.4\!\pm\!0.0)\!\times\!10^{0}$ & $162$ & $(1.8\!\pm\!0.0)\!\times\!10^{0}$ \\
    $-50$ & $139$ & $(1.7\!\pm\!0.0)\!\times\!10^{0}$ & $139$ & $\mathbf{(1.3\!\pm\!0.0)\!\times\!10^{0}}$ & $139$ & $(1.4\!\pm\!0.0)\!\times\!10^{0}$ & $139$ & $(5.9\!\pm\!0.2)\!\times\!10^{0}$ & $130$ & $(1.6\!\pm\!0.0)\!\times\!10^{0}$ & $130$ & $(1.5\!\pm\!0.0)\!\times\!10^{0}$ & $139$ & $(2.0\!\pm\!0.0)\!\times\!10^{0}$ \\
    $-75$ & $115$ & $(1.9\!\pm\!0.0)\!\times\!10^{0}$ & $114$ & $\mathbf{(1.5\!\pm\!0.0)\!\times\!10^{0}}$ & $115$ & $(1.6\!\pm\!0.0)\!\times\!10^{0}$ & $115$ & $(6.8\!\pm\!0.3)\!\times\!10^{0}$ & $105$ & $(1.8\!\pm\!0.0)\!\times\!10^{0}$ & $105$ & $(1.7\!\pm\!0.0)\!\times\!10^{0}$ & $114$ & $(2.3\!\pm\!0.0)\!\times\!10^{0}$ \\
    \bottomrule
    \end{tabular}%
    }
    \end{table}
    
    \begin{table}[htb]
    \caption{Experiment One (CIFAR-10): additional function-change measures for the same neuroadaptation conditions. Results are mean $\pm$ standard error over 20 independent repeats. $D_{\mathrm{KL}}$ is provided by \textit{Eqn.}~\ref{Eqn:KLmetric} and $\Delta a=a_{\mathrm{post}}-a_{\mathrm{pre}}$ is the change in test accuracy in percentage points; positive values indicate increased accuracy. The baseline row reports the unadapted test accuracy $a_0$ as a percentage. Bold values indicate the smallest mean $D_{\mathrm{KL}}$ and the accuracy change closest to zero within each neuroadaptation condition.}
    \label{Tab:CIFAR10_KL}
    \centering
    \scriptsize
    \setlength{\tabcolsep}{2.2pt}
    \renewcommand{\arraystretch}{1.12}
    \resizebox{\textwidth}{!}{%
    \begin{tabular}{r*{14}{c}}
    \toprule
    \text{Neuroadaptation} & \multicolumn{2}{c}{$o_{\mathrm{none}},\ \vec{\psi}_{\mathrm{decay}}$} & \multicolumn{2}{c}{$o_{\mathrm{constant}},\ \vec{\psi}_{\mathrm{decay}}$} & \multicolumn{2}{c}{$o_{\mathrm{trainable}},\ \vec{\psi}_{\mathrm{decay}}$} & \multicolumn{2}{c}{$o_{\mathrm{trainable}},\ \vec{\psi}_{\mathrm{none}}$} & \multicolumn{2}{c}{$o_{\mathrm{trainable}},\ \vec{\psi}_{\mathrm{constant}}$} & \multicolumn{2}{c}{$o_{\mathrm{trainable}},\ \vec{\psi}_{\mathrm{trainable}}$} & \multicolumn{2}{c}{$o_{\mathrm{trainable}},\ \vec{b}_{\mathrm{adjusted}}$} \\
    \cmidrule(lr){2-3}
    \cmidrule(lr){4-5}
    \cmidrule(lr){6-7}
    \cmidrule(lr){8-9}
    \cmidrule(lr){10-11}
    \cmidrule(lr){12-13}
    \cmidrule(lr){14-15}
     & $D_{\mathrm{KL}}$ & $\Delta a$ (pp) & $D_{\mathrm{KL}}$ & $\Delta a$ (pp) & $D_{\mathrm{KL}}$ & $\Delta a$ (pp) & $D_{\mathrm{KL}}$ & $\Delta a$ (pp) & $D_{\mathrm{KL}}$ & $\Delta a$ (pp) & $D_{\mathrm{KL}}$ & $\Delta a$ (pp) & $D_{\mathrm{KL}}$ & $\Delta a$ (pp) \\
    \midrule
    $+75$ & $(1.3\!\pm\!0.2)\!\times\!10^{-9}$ & $-0.0\!\pm\!0.0$ & $(1.5\!\pm\!0.2)\!\times\!10^{-9}$ & $\mathbf{0.0\!\pm\!0.0}$ & $(1.2\!\pm\!0.1)\!\times\!10^{-9}$ & $-0.0\!\pm\!0.0$ & $(1.3\!\pm\!0.2)\!\times\!10^{-9}$ & $+0.0\!\pm\!0.0$ & $\mathbf{(1.1\!\pm\!0.2)\!\times\!10^{-9}}$ & $\mathbf{0.0\!\pm\!0.0}$ & $(1.2\!\pm\!0.1)\!\times\!10^{-9}$ & $-0.0\!\pm\!0.0$ & $(1.6\!\pm\!0.2)\!\times\!10^{-9}$ & $+0.0\!\pm\!0.0$ \\
    $+50$ & $(1.4\!\pm\!0.1)\!\times\!10^{-9}$ & $-0.0\!\pm\!0.0$ & $(1.5\!\pm\!0.2)\!\times\!10^{-9}$ & $\mathbf{0.0\!\pm\!0.0}$ & $(1.5\!\pm\!0.2)\!\times\!10^{-9}$ & $-0.0\!\pm\!0.0$ & $(1.4\!\pm\!0.2)\!\times\!10^{-9}$ & $+0.0\!\pm\!0.0$ & $\mathbf{(1.1\!\pm\!0.2)\!\times\!10^{-9}}$ & $\mathbf{0.0\!\pm\!0.0}$ & $(1.2\!\pm\!0.1)\!\times\!10^{-9}$ & $-0.0\!\pm\!0.0$ & $(1.7\!\pm\!0.2)\!\times\!10^{-9}$ & $+0.0\!\pm\!0.0$ \\
    $+25$ & $(1.4\!\pm\!0.1)\!\times\!10^{-9}$ & $-0.0\!\pm\!0.0$ & $(1.4\!\pm\!0.2)\!\times\!10^{-9}$ & $\mathbf{0.0\!\pm\!0.0}$ & $(1.3\!\pm\!0.2)\!\times\!10^{-9}$ & $-0.0\!\pm\!0.0$ & $(1.2\!\pm\!0.1)\!\times\!10^{-9}$ & $+0.0\!\pm\!0.0$ & $\mathbf{(1.1\!\pm\!0.2)\!\times\!10^{-9}}$ & $\mathbf{0.0\!\pm\!0.0}$ & $(1.3\!\pm\!0.2)\!\times\!10^{-9}$ & $-0.0\!\pm\!0.0$ & $(1.6\!\pm\!0.2)\!\times\!10^{-9}$ & $+0.0\!\pm\!0.0$ \\
    $+5$ & $(1.4\!\pm\!0.2)\!\times\!10^{-9}$ & $-0.0\!\pm\!0.0$ & $(1.3\!\pm\!0.2)\!\times\!10^{-9}$ & $\mathbf{0.0\!\pm\!0.0}$ & $(1.3\!\pm\!0.2)\!\times\!10^{-9}$ & $-0.0\!\pm\!0.0$ & $(1.3\!\pm\!0.1)\!\times\!10^{-9}$ & $+0.0\!\pm\!0.0$ & $(1.1\!\pm\!0.2)\!\times\!10^{-9}$ & $\mathbf{0.0\!\pm\!0.0}$ & $\mathbf{(1.0\!\pm\!0.2)\!\times\!10^{-9}}$ & $-0.0\!\pm\!0.0$ & $(1.6\!\pm\!0.2)\!\times\!10^{-9}$ & $+0.0\!\pm\!0.0$ \\
    $+1$ & $(1.3\!\pm\!0.2)\!\times\!10^{-9}$ & $-0.0\!\pm\!0.0$ & $(1.4\!\pm\!0.3)\!\times\!10^{-9}$ & $\mathbf{0.0\!\pm\!0.0}$ & $(1.3\!\pm\!0.2)\!\times\!10^{-9}$ & $-0.0\!\pm\!0.0$ & $(1.1\!\pm\!0.2)\!\times\!10^{-9}$ & $+0.0\!\pm\!0.0$ & $\mathbf{(1.0\!\pm\!0.1)\!\times\!10^{-9}}$ & $\mathbf{0.0\!\pm\!0.0}$ & $(1.1\!\pm\!0.1)\!\times\!10^{-9}$ & $-0.0\!\pm\!0.0$ & $(1.7\!\pm\!0.2)\!\times\!10^{-9}$ & $+0.0\!\pm\!0.0$ \\
    \addlinespace[1pt]
    $0$ & $(1.3\!\pm\!0.1)\!\times\!10^{-9}$ & $-0.0\!\pm\!0.0$ & $(1.5\!\pm\!0.2)\!\times\!10^{-9}$ & $\mathbf{0.0\!\pm\!0.0}$ & $(1.6\!\pm\!0.1)\!\times\!10^{-9}$ & $-0.0\!\pm\!0.0$ & $(1.2\!\pm\!0.1)\!\times\!10^{-9}$ & $+0.0\!\pm\!0.0$ & $\mathbf{(1.0\!\pm\!0.2)\!\times\!10^{-9}}$ & $\mathbf{0.0\!\pm\!0.0}$ & $(1.2\!\pm\!0.2)\!\times\!10^{-9}$ & $-0.0\!\pm\!0.0$ & $(1.7\!\pm\!0.2)\!\times\!10^{-9}$ & $+0.0\!\pm\!0.0$ \\
    \addlinespace[1pt]
    $-1$ & $(8.1\!\pm\!4.1)\!\times\!10^{-9}$ & $-0.0\!\pm\!0.0$ & $(1.5\!\pm\!0.2)\!\times\!10^{-9}$ & $\mathbf{0.0\!\pm\!0.0}$ & $\mathbf{(1.2\!\pm\!0.1)\!\times\!10^{-9}}$ & $-0.0\!\pm\!0.0$ & $(5.3\!\pm\!2.4)\!\times\!10^{-8}$ & $+0.0\!\pm\!0.0$ & $(1.7\!\pm\!0.3)\!\times\!10^{-9}$ & $-0.0\!\pm\!0.0$ & $(1.4\!\pm\!0.2)\!\times\!10^{-9}$ & $-0.0\!\pm\!0.0$ & $(1.1\!\pm\!0.5)\!\times\!10^{-8}$ & $-0.0\!\pm\!0.0$ \\
    $-5$ & $(1.3\!\pm\!0.2)\!\times\!10^{-6}$ & $\mathbf{+0.0\!\pm\!0.0}$ & $(6.3\!\pm\!1.3)\!\times\!10^{-7}$ & $-0.0\!\pm\!0.0$ & $(4.5\!\pm\!0.9)\!\times\!10^{-7}$ & $-0.0\!\pm\!0.0$ & $(4.1\!\pm\!0.9)\!\times\!10^{-5}$ & $+0.0\!\pm\!0.0$ & $\mathbf{(3.7\!\pm\!0.8)\!\times\!10^{-7}}$ & $+0.0\!\pm\!0.0$ & $(3.8\!\pm\!0.7)\!\times\!10^{-7}$ & $-0.0\!\pm\!0.0$ & $(3.3\!\pm\!0.7)\!\times\!10^{-6}$ & $\mathbf{-0.0\!\pm\!0.0}$ \\
    $-25$ & $(2.0\!\pm\!0.1)\!\times\!10^{-4}$ & $-0.0\!\pm\!0.0$ & $\mathbf{(1.4\!\pm\!0.0)\!\times\!10^{-4}}$ & $-0.0\!\pm\!0.0$ & $(1.5\!\pm\!0.1)\!\times\!10^{-4}$ & $-0.0\!\pm\!0.0$ & $(2.2\!\pm\!0.2)\!\times\!10^{-3}$ & $-0.1\!\pm\!0.0$ & $(2.0\!\pm\!0.1)\!\times\!10^{-4}$ & $\mathbf{-0.0\!\pm\!0.0}$ & $(2.0\!\pm\!0.1)\!\times\!10^{-4}$ & $-0.0\!\pm\!0.0$ & $(3.0\!\pm\!0.2)\!\times\!10^{-4}$ & $-0.1\!\pm\!0.0$ \\
    $-50$ & $(2.6\!\pm\!0.1)\!\times\!10^{-4}$ & $-0.0\!\pm\!0.0$ & $\mathbf{(1.8\!\pm\!0.1)\!\times\!10^{-4}}$ & $-0.0\!\pm\!0.0$ & $(1.9\!\pm\!0.1)\!\times\!10^{-4}$ & $-0.0\!\pm\!0.0$ & $(2.7\!\pm\!0.2)\!\times\!10^{-3}$ & $-0.1\!\pm\!0.0$ & $(2.5\!\pm\!0.1)\!\times\!10^{-4}$ & $\mathbf{-0.0\!\pm\!0.0}$ & $(2.5\!\pm\!0.1)\!\times\!10^{-4}$ & $-0.0\!\pm\!0.0$ & $(3.9\!\pm\!0.2)\!\times\!10^{-4}$ & $-0.1\!\pm\!0.0$ \\
    $-75$ & $(3.4\!\pm\!0.1)\!\times\!10^{-4}$ & $\mathbf{-0.0\!\pm\!0.0}$ & $\mathbf{(2.3\!\pm\!0.1)\!\times\!10^{-4}}$ & $-0.1\!\pm\!0.0$ & $(2.5\!\pm\!0.1)\!\times\!10^{-4}$ & $-0.1\!\pm\!0.0$ & $(3.5\!\pm\!0.2)\!\times\!10^{-3}$ & $-0.1\!\pm\!0.0$ & $(3.3\!\pm\!0.1)\!\times\!10^{-4}$ & $-0.0\!\pm\!0.0$ & $(3.2\!\pm\!0.1)\!\times\!10^{-4}$ & $-0.1\!\pm\!0.0$ & $(5.2\!\pm\!0.2)\!\times\!10^{-4}$ & $-0.1\!\pm\!0.0$ \\
    \midrule
    \text{Baseline} & -- & $44.2\!\pm\!0.1$ & -- & $44.1\!\pm\!0.1$ & -- & $44.1\!\pm\!0.1$ & -- & $44.0\!\pm\!0.1$ & -- & $43.9\!\pm\!0.1$ & -- & $44.1\!\pm\!0.1$ & -- & $44.0\!\pm\!0.1$ \\
    \bottomrule
    \end{tabular}%
    }
    \end{table}
    
    Furthermore, CIFAR-10 results (\textit{Tab.}~\ref{Tab:CIFAR10_Erel}~and~\ref{Tab:CIFAR10_KL}) show a much less dramatic rise in function drift, even with a significant number of degenerated neurons. This plateauing is reflected in KL divergence, relative epsilon and test-set accuracy, demonstrating robustness despite a large decrease in the summed singular-value count. This greater tolerance to neurodegeneration suggests a dataset-dependent effect and may intuitively indicate a greater redundancy in networks at constant width for simpler datasets. Additionally, the constant intrinsic length appears more impactful, and, by contrast, the decaying $\vec{\psi}$ approach performed better on CIFAR-10 than on Caltech256. Constant versus trainable may show comparable performance because all cases lack post-neuroadaptive retraining, but these approaches do differ in initial training pre-neuroadaptation regardless, so they have the potential to act as generally useful parameters for the model.

    The CIFAR-100 results of \textit{Tab.}~\ref{Tab:CIFAR100_Erel}~and~\ref{Tab:CIFAR100_KL} remain more comparable to Caltech256, likely due to their many-class labellings.

    \begin{table}[htb]
    \caption{Experiment One (CIFAR-100): singular-value structure and relative function change under neuroadaptation at layer $L=1$ of the $[3072, 100, 100, 100]$ classifier. Results are mean $\pm$ standard error over 20 independent repeats. Row $0$ applies diagonalisation only; $+k$ subsequently applies $k$ neurogenesis operations in a single step and $-k$ applies $k$ neurodegeneration operations in a single step. $\Sigma_0$ and $\Sigma_1$ denote the sums of singular values before and after neuroadaptation, respectively. Singular-value standard errors are omitted for legibility; the largest singular-value standard error is 0.23. $\epsilon_{\mathrm{rel}}$ is the \textit{Eqn.}~\ref{Eqn:RelativeChange}'s metric, bold indicates the smallest value within each neuroadaptation condition.}
    \label{Tab:CIFAR100_Erel}
    \centering
    \scriptsize
    \setlength{\tabcolsep}{2.2pt}
    \renewcommand{\arraystretch}{1.12}
    \resizebox{\textwidth}{!}{%
    \begin{tabular}{r*{14}{c}}
    \toprule
    \text{Neuroadaptation} & \multicolumn{2}{c}{\shortstack{$o_{\mathrm{none}},\ \vec{\psi}_{\mathrm{decay}}$\\$\Sigma_0=260$}} & \multicolumn{2}{c}{\shortstack{$o_{\mathrm{constant}},\ \vec{\psi}_{\mathrm{decay}}$\\$\Sigma_0=260$}} & \multicolumn{2}{c}{\shortstack{$o_{\mathrm{trainable}},\ \vec{\psi}_{\mathrm{decay}}$\\$\Sigma_0=260$}} & \multicolumn{2}{c}{\shortstack{$o_{\mathrm{trainable}},\ \vec{\psi}_{\mathrm{none}}$\\$\Sigma_0=260$}} & \multicolumn{2}{c}{\shortstack{$o_{\mathrm{trainable}},\ \vec{\psi}_{\mathrm{constant}}$\\$\Sigma_0=252$}} & \multicolumn{2}{c}{\shortstack{$o_{\mathrm{trainable}},\ \vec{\psi}_{\mathrm{trainable}}$\\$\Sigma_0=252$}} & \multicolumn{2}{c}{\shortstack{$o_{\mathrm{trainable}},\ \vec{b}_{\mathrm{adjusted}}$\\$\Sigma_0=260$}} \\
    \cmidrule(lr){2-3}
    \cmidrule(lr){4-5}
    \cmidrule(lr){6-7}
    \cmidrule(lr){8-9}
    \cmidrule(lr){10-11}
    \cmidrule(lr){12-13}
    \cmidrule(lr){14-15}
     & $\Sigma_1$ & $\epsilon_{\mathrm{rel}}$ (\%) & $\Sigma_1$ & $\epsilon_{\mathrm{rel}}$ (\%) & $\Sigma_1$ & $\epsilon_{\mathrm{rel}}$ (\%) & $\Sigma_1$ & $\epsilon_{\mathrm{rel}}$ (\%) & $\Sigma_1$ & $\epsilon_{\mathrm{rel}}$ (\%) & $\Sigma_1$ & $\epsilon_{\mathrm{rel}}$ (\%) & $\Sigma_1$ & $\epsilon_{\mathrm{rel}}$ (\%) \\
    \midrule
    $+75$ & $260$ & $(1.7\!\pm\!0.1)\!\times\!10^{-3}$ & $260$ & $(1.7\!\pm\!0.1)\!\times\!10^{-3}$ & $260$ & $(1.6\!\pm\!0.1)\!\times\!10^{-3}$ & $260$ & $(1.6\!\pm\!0.0)\!\times\!10^{-3}$ & $252$ & $\mathbf{(1.4\!\pm\!0.1)\!\times\!10^{-3}}$ & $252$ & $(1.5\!\pm\!0.1)\!\times\!10^{-3}$ & $260$ & $(1.7\!\pm\!0.1)\!\times\!10^{-3}$ \\
    $+50$ & $260$ & $(1.7\!\pm\!0.1)\!\times\!10^{-3}$ & $260$ & $(1.7\!\pm\!0.1)\!\times\!10^{-3}$ & $260$ & $(1.6\!\pm\!0.1)\!\times\!10^{-3}$ & $260$ & $(1.6\!\pm\!0.0)\!\times\!10^{-3}$ & $252$ & $\mathbf{(1.4\!\pm\!0.1)\!\times\!10^{-3}}$ & $252$ & $(1.5\!\pm\!0.1)\!\times\!10^{-3}$ & $260$ & $(1.7\!\pm\!0.1)\!\times\!10^{-3}$ \\
    $+25$ & $260$ & $(1.7\!\pm\!0.1)\!\times\!10^{-3}$ & $260$ & $(1.7\!\pm\!0.1)\!\times\!10^{-3}$ & $260$ & $(1.6\!\pm\!0.1)\!\times\!10^{-3}$ & $260$ & $(1.6\!\pm\!0.0)\!\times\!10^{-3}$ & $252$ & $\mathbf{(1.4\!\pm\!0.1)\!\times\!10^{-3}}$ & $252$ & $(1.5\!\pm\!0.1)\!\times\!10^{-3}$ & $260$ & $(1.7\!\pm\!0.1)\!\times\!10^{-3}$ \\
    $+5$ & $260$ & $(1.7\!\pm\!0.1)\!\times\!10^{-3}$ & $260$ & $(1.7\!\pm\!0.1)\!\times\!10^{-3}$ & $260$ & $(1.6\!\pm\!0.1)\!\times\!10^{-3}$ & $260$ & $(1.6\!\pm\!0.0)\!\times\!10^{-3}$ & $252$ & $\mathbf{(1.4\!\pm\!0.1)\!\times\!10^{-3}}$ & $252$ & $(1.5\!\pm\!0.1)\!\times\!10^{-3}$ & $260$ & $(1.7\!\pm\!0.1)\!\times\!10^{-3}$ \\
    $+1$ & $260$ & $(1.7\!\pm\!0.1)\!\times\!10^{-3}$ & $260$ & $(1.7\!\pm\!0.1)\!\times\!10^{-3}$ & $260$ & $(1.6\!\pm\!0.1)\!\times\!10^{-3}$ & $260$ & $(1.6\!\pm\!0.0)\!\times\!10^{-3}$ & $252$ & $\mathbf{(1.4\!\pm\!0.1)\!\times\!10^{-3}}$ & $252$ & $(1.5\!\pm\!0.1)\!\times\!10^{-3}$ & $260$ & $(1.7\!\pm\!0.1)\!\times\!10^{-3}$ \\
    \addlinespace[1pt]
    $0$ & $260$ & $(1.7\!\pm\!0.1)\!\times\!10^{-3}$ & $260$ & $(1.7\!\pm\!0.1)\!\times\!10^{-3}$ & $260$ & $(1.6\!\pm\!0.1)\!\times\!10^{-3}$ & $260$ & $(1.6\!\pm\!0.0)\!\times\!10^{-3}$ & $252$ & $\mathbf{(1.4\!\pm\!0.1)\!\times\!10^{-3}}$ & $252$ & $(1.5\!\pm\!0.1)\!\times\!10^{-3}$ & $260$ & $(1.7\!\pm\!0.1)\!\times\!10^{-3}$ \\
    \addlinespace[1pt]
    $-1$ & $260$ & $(4.4\!\pm\!0.1)\!\times\!10^{-1}$ & $260$ & $(4.4\!\pm\!0.1)\!\times\!10^{-1}$ & $259$ & $(4.5\!\pm\!0.2)\!\times\!10^{-1}$ & $259$ & $(4.6\!\pm\!0.1)\!\times\!10^{-1}$ & $252$ & $\mathbf{(3.5\!\pm\!0.1)\!\times\!10^{-1}}$ & $252$ & $(3.6\!\pm\!0.2)\!\times\!10^{-1}$ & $259$ & $(4.5\!\pm\!0.1)\!\times\!10^{-1}$ \\
    $-5$ & $257$ & $(2.0\!\pm\!0.0)\!\times\!10^{0}$ & $257$ & $\mathbf{(1.9\!\pm\!0.0)\!\times\!10^{0}}$ & $257$ & $(2.0\!\pm\!0.0)\!\times\!10^{0}$ & $256$ & $(2.1\!\pm\!0.0)\!\times\!10^{0}$ & $249$ & $(2.0\!\pm\!0.0)\!\times\!10^{0}$ & $249$ & $(2.1\!\pm\!0.0)\!\times\!10^{0}$ & $257$ & $(2.0\!\pm\!0.0)\!\times\!10^{0}$ \\
    $-25$ & $237$ & $(2.6\!\pm\!0.0)\!\times\!10^{0}$ & $237$ & $\mathbf{(2.5\!\pm\!0.0)\!\times\!10^{0}}$ & $237$ & $(2.6\!\pm\!0.0)\!\times\!10^{0}$ & $237$ & $(2.7\!\pm\!0.0)\!\times\!10^{0}$ & $229$ & $(2.7\!\pm\!0.0)\!\times\!10^{0}$ & $229$ & $(2.7\!\pm\!0.0)\!\times\!10^{0}$ & $237$ & $(2.6\!\pm\!0.0)\!\times\!10^{0}$ \\
    $-50$ & $197$ & $(1.3\!\pm\!0.0)\!\times\!10^{+1}$ & $197$ & $(1.4\!\pm\!0.0)\!\times\!10^{+1}$ & $197$ & $(1.4\!\pm\!0.0)\!\times\!10^{+1}$ & $197$ & $\mathbf{(1.3\!\pm\!0.0)\!\times\!10^{+1}}$ & $188$ & $(1.6\!\pm\!0.0)\!\times\!10^{+1}$ & $188$ & $(1.7\!\pm\!0.0)\!\times\!10^{+1}$ & $197$ & $(1.4\!\pm\!0.1)\!\times\!10^{+1}$ \\
    $-75$ & $117$ & $(6.2\!\pm\!0.0)\!\times\!10^{+1}$ & $117$ & $(6.1\!\pm\!0.0)\!\times\!10^{+1}$ & $117$ & $(6.1\!\pm\!0.0)\!\times\!10^{+1}$ & $117$ & $(6.1\!\pm\!0.0)\!\times\!10^{+1}$ & $111$ & $(6.2\!\pm\!0.0)\!\times\!10^{+1}$ & $111$ & $(6.2\!\pm\!0.0)\!\times\!10^{+1}$ & $117$ & $\mathbf{(6.1\!\pm\!0.0)\!\times\!10^{+1}}$ \\
    \bottomrule
    \end{tabular}%
    }
    \end{table}
    
    \begin{table}[htb]
    \caption{Experiment One (CIFAR-100): additional function-change measures for the same neuroadaptation conditions. Results are mean $\pm$ standard error over 20 independent repeats. $D_{\mathrm{KL}}$ is provided by \textit{Eqn.}~\ref{Eqn:KLmetric} and $\Delta a=a_{\mathrm{post}}-a_{\mathrm{pre}}$ is the change in test accuracy in percentage points; positive values indicate increased accuracy. The baseline row reports the unadapted test accuracy $a_0$ as a percentage. Bold values indicate the smallest mean $D_{\mathrm{KL}}$ and the accuracy change closest to zero within each neuroadaptation condition.}
    \label{Tab:CIFAR100_KL}
    \centering
    \scriptsize
    \setlength{\tabcolsep}{2.2pt}
    \renewcommand{\arraystretch}{1.12}
    \resizebox{\textwidth}{!}{%
    \begin{tabular}{r*{14}{c}}
    \toprule
    \text{Neuroadaptation} & \multicolumn{2}{c}{$o_{\mathrm{none}},\ \vec{\psi}_{\mathrm{decay}}$} & \multicolumn{2}{c}{$o_{\mathrm{constant}},\ \vec{\psi}_{\mathrm{decay}}$} & \multicolumn{2}{c}{$o_{\mathrm{trainable}},\ \vec{\psi}_{\mathrm{decay}}$} & \multicolumn{2}{c}{$o_{\mathrm{trainable}},\ \vec{\psi}_{\mathrm{none}}$} & \multicolumn{2}{c}{$o_{\mathrm{trainable}},\ \vec{\psi}_{\mathrm{constant}}$} & \multicolumn{2}{c}{$o_{\mathrm{trainable}},\ \vec{\psi}_{\mathrm{trainable}}$} & \multicolumn{2}{c}{$o_{\mathrm{trainable}},\ \vec{b}_{\mathrm{adjusted}}$} \\
    \cmidrule(lr){2-3}
    \cmidrule(lr){4-5}
    \cmidrule(lr){6-7}
    \cmidrule(lr){8-9}
    \cmidrule(lr){10-11}
    \cmidrule(lr){12-13}
    \cmidrule(lr){14-15}
     & $D_{\mathrm{KL}}$ & $\Delta a$ (pp) & $D_{\mathrm{KL}}$ & $\Delta a$ (pp) & $D_{\mathrm{KL}}$ & $\Delta a$ (pp) & $D_{\mathrm{KL}}$ & $\Delta a$ (pp) & $D_{\mathrm{KL}}$ & $\Delta a$ (pp) & $D_{\mathrm{KL}}$ & $\Delta a$ (pp) & $D_{\mathrm{KL}}$ & $\Delta a$ (pp) \\
    \midrule
    $+75$ & $(1.6\!\pm\!0.2)\!\times\!10^{-9}$ & $+0.0\!\pm\!0.0$ & $(1.5\!\pm\!0.2)\!\times\!10^{-9}$ & $+0.0\!\pm\!0.0$ & $\mathbf{(1.2\!\pm\!0.2)\!\times\!10^{-9}}$ & $-0.0\!\pm\!0.0$ & $(1.6\!\pm\!0.2)\!\times\!10^{-9}$ & $\mathbf{0.0\!\pm\!0.0}$ & $(1.2\!\pm\!0.1)\!\times\!10^{-9}$ & $+0.0\!\pm\!0.0$ & $(1.6\!\pm\!0.2)\!\times\!10^{-9}$ & $\mathbf{0.0\!\pm\!0.0}$ & $(1.4\!\pm\!0.2)\!\times\!10^{-9}$ & $+0.0\!\pm\!0.0$ \\
    $+50$ & $(1.6\!\pm\!0.2)\!\times\!10^{-9}$ & $+0.0\!\pm\!0.0$ & $(1.6\!\pm\!0.2)\!\times\!10^{-9}$ & $+0.0\!\pm\!0.0$ & $(1.2\!\pm\!0.2)\!\times\!10^{-9}$ & $-0.0\!\pm\!0.0$ & $(1.6\!\pm\!0.2)\!\times\!10^{-9}$ & $\mathbf{0.0\!\pm\!0.0}$ & $\mathbf{(1.1\!\pm\!0.2)\!\times\!10^{-9}}$ & $+0.0\!\pm\!0.0$ & $(1.6\!\pm\!0.2)\!\times\!10^{-9}$ & $\mathbf{0.0\!\pm\!0.0}$ & $(1.4\!\pm\!0.2)\!\times\!10^{-9}$ & $+0.0\!\pm\!0.0$ \\
    $+25$ & $(1.6\!\pm\!0.2)\!\times\!10^{-9}$ & $+0.0\!\pm\!0.0$ & $(1.5\!\pm\!0.2)\!\times\!10^{-9}$ & $+0.0\!\pm\!0.0$ & $\mathbf{(1.2\!\pm\!0.2)\!\times\!10^{-9}}$ & $-0.0\!\pm\!0.0$ & $(1.6\!\pm\!0.2)\!\times\!10^{-9}$ & $\mathbf{0.0\!\pm\!0.0}$ & $(1.2\!\pm\!0.1)\!\times\!10^{-9}$ & $+0.0\!\pm\!0.0$ & $(1.6\!\pm\!0.2)\!\times\!10^{-9}$ & $\mathbf{0.0\!\pm\!0.0}$ & $(1.4\!\pm\!0.2)\!\times\!10^{-9}$ & $+0.0\!\pm\!0.0$ \\
    $+5$ & $(1.5\!\pm\!0.1)\!\times\!10^{-9}$ & $+0.0\!\pm\!0.0$ & $(1.4\!\pm\!0.2)\!\times\!10^{-9}$ & $+0.0\!\pm\!0.0$ & $(1.2\!\pm\!0.2)\!\times\!10^{-9}$ & $-0.0\!\pm\!0.0$ & $(1.5\!\pm\!0.2)\!\times\!10^{-9}$ & $\mathbf{0.0\!\pm\!0.0}$ & $\mathbf{(1.2\!\pm\!0.1)\!\times\!10^{-9}}$ & $+0.0\!\pm\!0.0$ & $(1.6\!\pm\!0.2)\!\times\!10^{-9}$ & $\mathbf{0.0\!\pm\!0.0}$ & $(1.5\!\pm\!0.2)\!\times\!10^{-9}$ & $+0.0\!\pm\!0.0$ \\
    $+1$ & $(1.6\!\pm\!0.2)\!\times\!10^{-9}$ & $+0.0\!\pm\!0.0$ & $(1.5\!\pm\!0.2)\!\times\!10^{-9}$ & $+0.0\!\pm\!0.0$ & $\mathbf{(1.2\!\pm\!0.2)\!\times\!10^{-9}}$ & $-0.0\!\pm\!0.0$ & $(1.9\!\pm\!0.2)\!\times\!10^{-9}$ & $\mathbf{0.0\!\pm\!0.0}$ & $(1.2\!\pm\!0.2)\!\times\!10^{-9}$ & $+0.0\!\pm\!0.0$ & $(1.4\!\pm\!0.2)\!\times\!10^{-9}$ & $\mathbf{0.0\!\pm\!0.0}$ & $(1.4\!\pm\!0.2)\!\times\!10^{-9}$ & $+0.0\!\pm\!0.0$ \\
    \addlinespace[1pt]
    $0$ & $(1.7\!\pm\!0.2)\!\times\!10^{-9}$ & $+0.0\!\pm\!0.0$ & $(1.4\!\pm\!0.2)\!\times\!10^{-9}$ & $+0.0\!\pm\!0.0$ & $(1.2\!\pm\!0.1)\!\times\!10^{-9}$ & $-0.0\!\pm\!0.0$ & $(1.7\!\pm\!0.2)\!\times\!10^{-9}$ & $\mathbf{0.0\!\pm\!0.0}$ & $\mathbf{(1.2\!\pm\!0.2)\!\times\!10^{-9}}$ & $+0.0\!\pm\!0.0$ & $(1.4\!\pm\!0.2)\!\times\!10^{-9}$ & $\mathbf{0.0\!\pm\!0.0}$ & $(1.5\!\pm\!0.2)\!\times\!10^{-9}$ & $+0.0\!\pm\!0.0$ \\
    \addlinespace[1pt]
    $-1$ & $(7.8\!\pm\!0.4)\!\times\!10^{-5}$ & $-0.0\!\pm\!0.0$ & $(7.6\!\pm\!0.5)\!\times\!10^{-5}$ & $-0.0\!\pm\!0.0$ & $(8.3\!\pm\!0.6)\!\times\!10^{-5}$ & $-0.0\!\pm\!0.0$ & $(8.9\!\pm\!0.5)\!\times\!10^{-5}$ & $-0.0\!\pm\!0.0$ & $\mathbf{(4.9\!\pm\!0.4)\!\times\!10^{-5}}$ & $\mathbf{-0.0\!\pm\!0.0}$ & $(5.2\!\pm\!0.5)\!\times\!10^{-5}$ & $-0.0\!\pm\!0.0$ & $(7.9\!\pm\!0.4)\!\times\!10^{-5}$ & $-0.0\!\pm\!0.0$ \\
    $-5$ & $(8.9\!\pm\!0.3)\!\times\!10^{-4}$ & $-0.1\!\pm\!0.0$ & $\mathbf{(8.1\!\pm\!0.2)\!\times\!10^{-4}}$ & $-0.1\!\pm\!0.0$ & $(8.8\!\pm\!0.3)\!\times\!10^{-4}$ & $-0.1\!\pm\!0.0$ & $(1.0\!\pm\!0.0)\!\times\!10^{-3}$ & $-0.1\!\pm\!0.0$ & $(9.8\!\pm\!0.3)\!\times\!10^{-4}$ & $-0.1\!\pm\!0.0$ & $(10.0\!\pm\!0.3)\!\times\!10^{-4}$ & $\mathbf{-0.1\!\pm\!0.0}$ & $(9.3\!\pm\!0.3)\!\times\!10^{-4}$ & $-0.1\!\pm\!0.0$ \\
    $-25$ & $(1.5\!\pm\!0.1)\!\times\!10^{-3}$ & $-0.1\!\pm\!0.0$ & $\mathbf{(1.4\!\pm\!0.0)\!\times\!10^{-3}}$ & $-0.1\!\pm\!0.0$ & $(1.5\!\pm\!0.0)\!\times\!10^{-3}$ & $-0.1\!\pm\!0.0$ & $(1.6\!\pm\!0.1)\!\times\!10^{-3}$ & $\mathbf{-0.1\!\pm\!0.0}$ & $(1.7\!\pm\!0.1)\!\times\!10^{-3}$ & $-0.1\!\pm\!0.0$ & $(1.7\!\pm\!0.1)\!\times\!10^{-3}$ & $-0.1\!\pm\!0.0$ & $(1.6\!\pm\!0.0)\!\times\!10^{-3}$ & $-0.1\!\pm\!0.0$ \\
    $-50$ & $(4.7\!\pm\!0.4)\!\times\!10^{-2}$ & $-1.1\!\pm\!0.1$ & $(5.4\!\pm\!0.4)\!\times\!10^{-2}$ & $-1.2\!\pm\!0.1$ & $(5.0\!\pm\!0.4)\!\times\!10^{-2}$ & $-1.2\!\pm\!0.1$ & $\mathbf{(4.5\!\pm\!0.4)\!\times\!10^{-2}}$ & $\mathbf{-1.1\!\pm\!0.1}$ & $(7.1\!\pm\!0.3)\!\times\!10^{-2}$ & $-1.5\!\pm\!0.1$ & $(7.3\!\pm\!0.3)\!\times\!10^{-2}$ & $-1.5\!\pm\!0.1$ & $(5.2\!\pm\!0.4)\!\times\!10^{-2}$ & $-1.2\!\pm\!0.1$ \\
    $-75$ & $(8.1\!\pm\!0.1)\!\times\!10^{-1}$ & $-11.5\!\pm\!0.1$ & $(8.2\!\pm\!0.1)\!\times\!10^{-1}$ & $-11.6\!\pm\!0.1$ & $(7.9\!\pm\!0.1)\!\times\!10^{-1}$ & $-11.4\!\pm\!0.1$ & $\mathbf{(7.6\!\pm\!0.1)\!\times\!10^{-1}}$ & $\mathbf{-11.2\!\pm\!0.1}$ & $(8.5\!\pm\!0.1)\!\times\!10^{-1}$ & $-11.7\!\pm\!0.1$ & $(8.5\!\pm\!0.1)\!\times\!10^{-1}$ & $-11.8\!\pm\!0.1$ & $(7.7\!\pm\!0.1)\!\times\!10^{-1}$ & $-11.3\!\pm\!0.1$ \\
    \midrule
    \text{Baseline} & -- & $18.6\!\pm\!0.1$ & -- & $18.6\!\pm\!0.0$ & -- & $18.6\!\pm\!0.0$ & -- & $18.6\!\pm\!0.1$ & -- & $18.6\!\pm\!0.0$ & -- & $18.7\!\pm\!0.0$ & -- & $18.7\!\pm\!0.0$ \\
    \bottomrule
    \end{tabular}%
    }
    \end{table}

    Overall, neurogenesis and diagonalisation appear inconsequential to performance or even logit distributions. This directly confirms the theoretical exact function invariance enabled through the approach up to numerical precision. Moreover, the sparsified networks reduce from $100201$ parameters in this layer to just $301$, whilst remarkably retaining exact performance. This is a $99.7\%$ reduction, with no measurable loss in classification performance across all datasets tested --- this is without any retraining or fine-tuning. Moreover, this reaches $99.98\%$ sparsification for no intrinsic length or $\vec{\psi}$. 
    
    Neurodegeneration also approximates function well for most networks up to around $50\%$ width reduction, while deterioration occurs more sharply when many singular values are removed. It appears that CIFAR-10 is more resilient to this process, likely due to the substantially fewer output classes, that may require fewer functionally important directions to maintain its prediction. Many-class datasets showed greater deterioration in relative logit change and KL divergence, not just test accuracy, suggesting less redundancy in their fixed-width layers, where more complex transforms are required of the weights to separate classes. This is expected for a more demanding dataset.
    
    Furthermore, we observed a negligible difference between constant and trainable $\vec{\psi}$ or $o$ in the table. This is directly because these networks did not undergo retraining, where the constant or trained values would exhibit static or changing values. Therefore, the nature of these parameters did not have the opportunity to influence the later invariance, leading to comparable results. Intrinsic length appeared to improve neurodegeneration invariance, while $\vec{\psi}$-compensations appeared less influential and produced more comparable performance; however, the preference for using it was clearer on CIFAR-10, where its absence resulted in poorer performance at $-75$ neurons degenerated.

\subsection{Experiment Two: Singular Value and Bias Spectra Evolutions\label{App:ExperimentTwo}}

    The experiment displayed in \textit{Fig.}~\ref{Fig:IsotropicSpectra} shows the evolution of the layerwise singular value and bias spectra of a static (non-neuroadaptive) isotropic multilayer perceptron over $40$ repeats for the architecture $\left[3072, 100, 100, 10\right]$, trained on CIFAR-10. This experiment examines spectral and bias shifts during training and how they may affect thresholding choices. Although they have little bearing on thresholding, the bias elements are included for completeness and to allow a more holistic comparison of the training dynamics that differ between anisotropic and isotropic networks.
    
    The spectra are pooled across all independent repeats at each epoch snapshot, enabling an evolving joyplot of the results over 40 epochs. We performed the aforementioned standard training approach as described above, including input standardisation. As with all experiments, we consistently used orthogonal initialisation to ensure all initial singular values were equal to 1. This reduces ambiguity about whether a singular value is increasingly significant to the function or decaying to become negligible by providing a known initial value of 1. This contrasts with random-normal approaches, which typically have a more distributed spectrum, obfuscating such trends. 
    
    We used a slightly boosted learning rate of $\eta=0.05$ to accentuate long-term trend evolution over the $40$-epoch window, allowing more repeats to be taken. This made it easier to observe the increase, decay and separation of singular values in this experiment. This was preferred because final classification accuracy was not important for this experiment; rather, we needed a clearer diagnosis of whether singular-value thresholding is experimentally plausible. Together with experiment three, we deemed this thresholding appropriate.
    
    The plots use log-scaled heights and signed-log ($\operatorname{sign}{l}\ln\left(1+|l|\right)$) x-axis scaling, with the x-axis range custom-chosen to include all notable details.
    
    To complement the CIFAR-10, 100-width plot presented in \textit{Sec.}~\ref{Sec:Empirical}, we provide an array of additional spectra plots. Particularly isotropic (isotropic-tanh) and anisotropic (standard, elementwise tanh) are trained under identical conditions, allowing us to study how these superficially similar activation functions actually lead to drastically different spectra through learning and thus approach the problem very differently. Furthermore, a broad sweep of widths for both of these is provided: $\left[3072, m, m, 10\right]$, for $m\in\left\{100, 500, 1000\right\}$, whose purpose is to examine how the spectra change across a range of widths (accessible by neuroadaptation), and how this may also impact thresholding. Finally, motivated by the analyses of \textit{App.}~\ref{App:ExperimentOne}, where neurodegeneration appeared less impactful for easier datasets such as CIFAR-10, but had a sharper degradation in Caltech256, we repeat the isotropic width-sweeps for a Caltech256 network to examine how the spectra differ in such datasets. These Caltech256 networks are trained otherwise identically to CIFAR-10, but for an architecture of $[12288, m, m, 256]$, for $m\in\left\{100, 500, 1000\right\}$.

\subsubsection{Comparison to Identical Architecture with Standard Tanh}

\begin{figure}[htb]
        \centering
        \includegraphics[width=\textwidth]{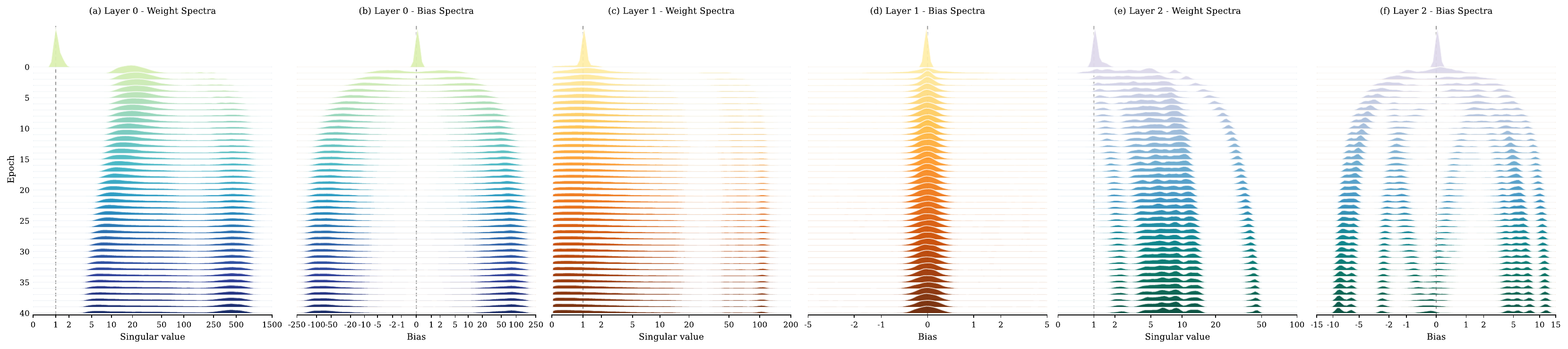}
        \caption{Experiment Two (\textit{App.}~\ref{App:ExperimentTwo}) evolving layerwise singular value and bias spectra for a $\left[3072, 100, 100, 10\right]$ \textbf{isotropic-tanh} CIFAR-10 classifier.}
        \label{Fig:IsoSpectra100}
    \end{figure}

\begin{figure}[htb]
        \centering
        \includegraphics[width=\textwidth]{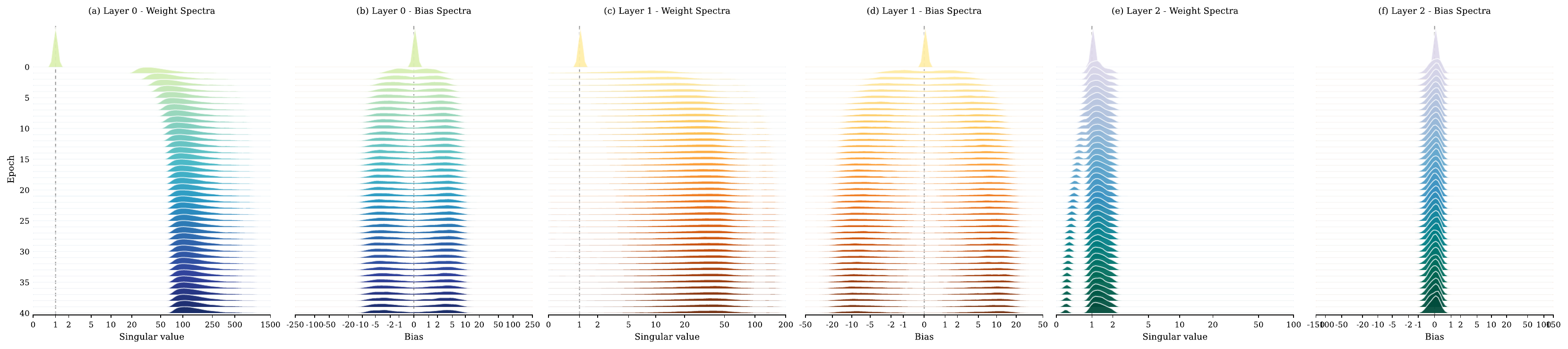}
        \caption{Experiment Two (\textit{App.}~\ref{App:ExperimentTwo}) evolving layerwise singular value and bias spectra for a $\left[3072, 100, 100, 10\right]$ \textbf{standard-tanh} CIFAR-10 classifier.}
        \label{Fig:AnisoSpectra100}
    \end{figure}

    The experiments in this section are trained identically, with plots showing evolving layerwise singular value and bias spectra for a non-neuroadaptive isotropic-tanh or standard-tanh MLP trained on CIFAR-10 classification. We collate the spectra over $40$ independent repeats, with an emerging strong consistency. Singular values begin at one, indicated by the dashed line, due to orthogonal initialisation, and biases at zero due to zero initialisation. The plots of \textit{Fig.}~\ref{Fig:IsoSpectra100}~and~\ref{Fig:AnisoSpectra100} show an isotropic-tanh and standard-tanh 100-hidden width CIFAR-10 classification networks respectively. \textit{Fig.}~\ref{Fig:IsoSpectra100} is identical to \textit{Fig.}~\ref{Fig:IsotropicSpectra} of \textit{Sec.}~\ref{Sec:Empirical}.

    The isotropic network shows key diversification across all layers, but minimal spread around zero for layer-one biases. Both layer-zero and layer-one singular value spectra show a general bulk distribution that initially increases with training before separating into two distinct clusters, the smallest of which appears to decay toward zero, while the higher cluster remains stable. The split from this unimodal to bimodal cluster occurs rapidly within the first epoch. It is this descending tail that thresholding may begin to progressively neurodegenerate if applied. The most striking characteristic is arguably the layer-two ($\mathbb{R}^{100}\rightarrow\mathbb{R}^{10}$ output-layer), which rapidly separates into several distinct singular-value clusters with clear gaps between the dense bands. These likely reflect strong consistency over the $40$ repeats for the $10$ singular values in this layer, showing that isotropic-tanh reliably finds a comparable solution. This may result from the flattened loss around orthogonal-group orbits, where similar trajectories are produced despite differing initialisations, compared with the more nuanced loss structure of discrete-group-defined activations, which may cause more unpredictable deflections. The layer-two biases also show a rich structure of distinct bands (some progressively decaying to zero), unlike the simple bimodal structure of layer-zero biases or the unimodal, zero-centred cluster of layer-one. The decaying band crossovers which occur in layer-two biases suggest these correspond to distinct elementwise outputs, since the crossovers do not appear to induce any interference in the decaying trajectories. 
    
\begin{figure}[htb]
        \centering
        \includegraphics[width=\textwidth]{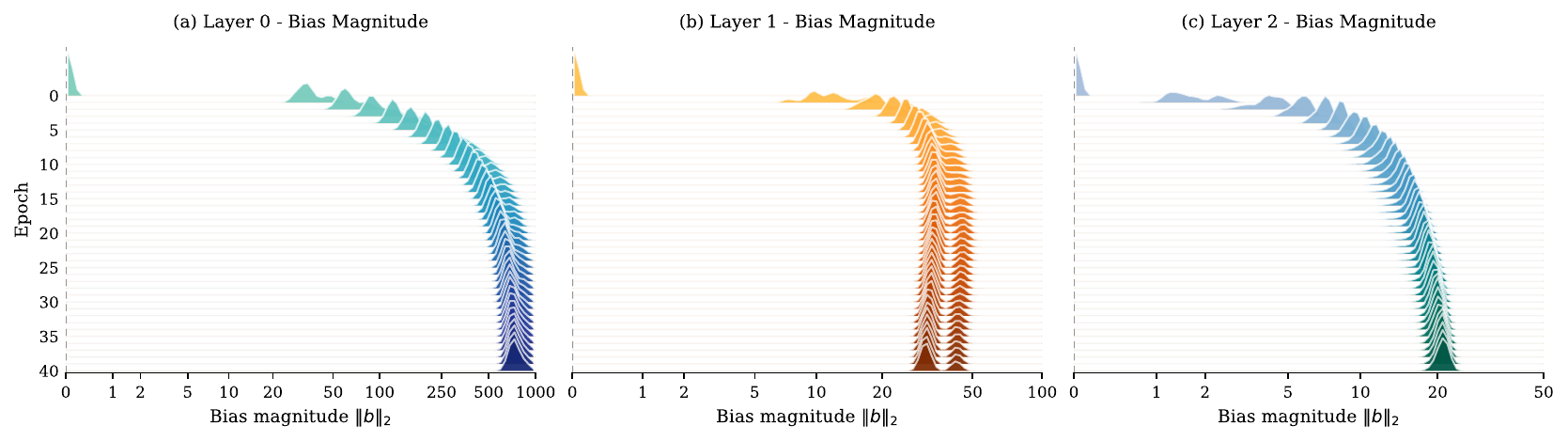}
        \caption{Experiment Two (\textit{App.}~\ref{App:ExperimentTwo}) evolving layerwise bias-magnitude spectra for a $\left[3072, 100, 100, 10\right]$ \textbf{isotropic-tanh} CIFAR-10 classifier.}
        \label{Fig:BiasIso100}
    \end{figure}

    Practically, because isotropic-tanh is spherically symmetric and we use signed-log x-axis scaling, the bimodal distribution of layer-zero biases instead indicates a unimodal, non-zero distribution of magnitudes shown in \textit{Fig.}~\ref{Fig:BiasIso100} --- this basis-freedom is constrained by the lack of two-sided orthogonal symmetry of layer-zero and layer-two and the dataset's symmetry breaking. This also clarifies a bifurcation in layer-one biases.
    
    Overall, this suggests that isotropic networks are particularly anisotropic in their transform: some directions are amplified considerably more than others, and this is consistent between initialisations. Layers zero and two appear to produce the dominant transforms in this case, while layer-one has only a few singular values that remain stably separated from the general bulk distribution, implying that only a few directions are anisotropically amplified. All layers also appear to have large-magnitude singular values, suggesting the network amplifies scale considerably, perhaps using saturation of the isotropic-tanh for non-linear reshaping.

    \begin{figure}[htb]
        \centering
        \includegraphics[width=\textwidth]{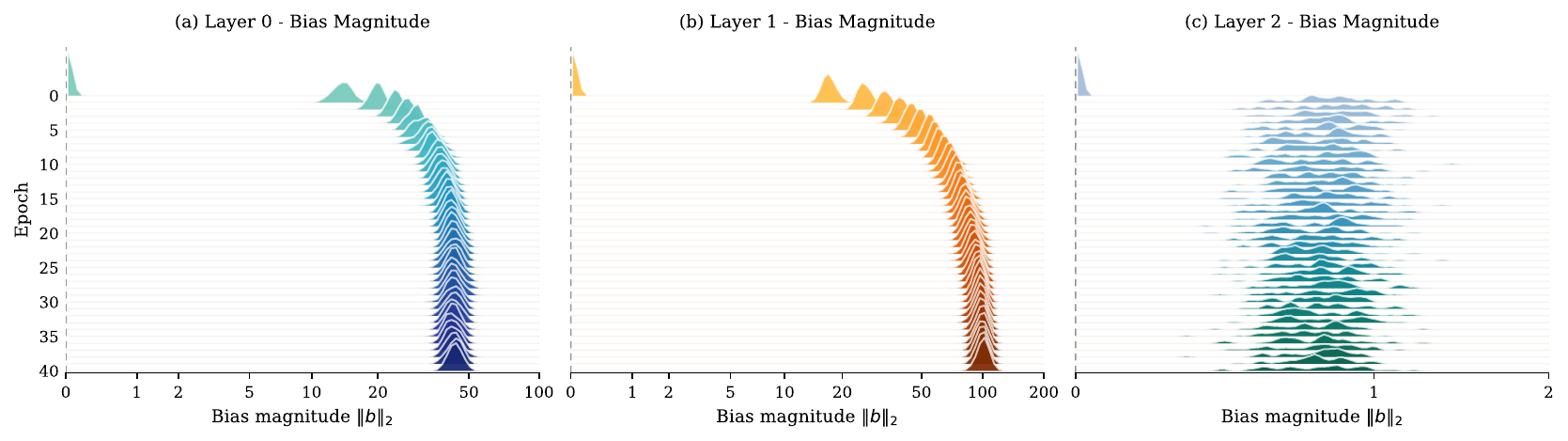}
        \caption{Experiment Two (\textit{App.}~\ref{App:ExperimentTwo}) evolving layerwise bias-magnitude spectra for a $\left[3072, 100, 100, 10\right]$ \textbf{standard-tanh} CIFAR-10 classifier.}
        \label{Fig:BiasAniso100}
    \end{figure}

    Overall, this is significantly dissimilar to the standard-tanh case. The standard-tanh case shows a more pronounced, non-decaying spectrum in layers zero and one, suggesting a largely coherent bulk evolution that may contribute to pruning difficulties. The final layer shows little diversification, with training staying peaked around the initialised values and only a small decay in some singular values --- suggesting it may act largely as a semi-orthogonal transform. Thus, anisotropic functions (standard-tanh) conversely have a more isotropic-like affine transform, with all directions scaled similarly, markedly different from isotropic-tanh's behaviour. Besides layer-two, they have somewhat comparable amplification, suggesting the networks prefer a specific activation scale, indicative of exploiting the saturating non-linear regime. Standard-tanh also appears to have distinctly different layers dominating the transform; layer-two shows little affine specialisation, unlike the rich structure in isotropic-tanh's last layer. However, layer-two's bias-magnitude structure is more nuanced in standard-tanh, as shown in \textit{Fig.}~\ref{Fig:BiasAniso100}. Unlike isotropic-tanh, care is needed when interpreting bias-magnitude over elementwise distribution in standard-tanh because it lacks rotational symmetry.

    Considered together, isotropic-tanh plots consistently show stronger stratification than in \textit{Fig.}~\ref{Fig:AnisoSpectra100}, suggesting parameters become highly anisotropic despite the isotropic activation form. The data and task are intrinsically anisotropic due to the image-to-classification objective, and the emergence of anisotropic parameters in isotropic networks indicates a clear, spontaneous symmetry breaking necessitated to achieve this, with a few singular directions becoming specialised while many decay. This contrasts with standard tanh networks, which tend to distribute their spectra as a broad bulk, aligning with more classical interpretations of them as an assortment of somewhat comparably acting individual feature detectors.

    Additionally, isotropic-tanh networks appear to have a greater utilisation of the final layer compared to standard-tanh networks, whilst standard-tanh shows a smoother zeroth-layer spectrum, consistent with many directions being amplified to comparable magnitudes. Thus, they appear to exhibit very different computational regimes despite superficial similarities of the function (even identical along the axes). Thus, functional form has a marked effect on training and should not be treated as comparable---each induces different biases in optimisation. Collectively, there appears to be an inverse tradeoff between the an/isotropy of the activation and the an/isotropy of the affine map.
    
    Isotropic tanh exhibits a distinct, well-separated ordering of its singular values, particularly in the last layer, supporting the paper's motivation for thresholded neurodegeneration. Additionally, this supports interpreting singular values as having a perturbative impact on the function --- isotropic networks arrange into distinct bands of singular values, indicating increasingly amplified anisotropic scaling of activations. The strong similarity across repeats may also stem from the flat-loss orbit induced by continuous orthogonal equivariance per layer, where many optimisation trajectories may follow comparable paths up to rotation because they are not deflected by the loss structure that emerges along those orbits for permutation-derived primitives.

    Overall, both demonstrate qualitatively different modes of learning, indicating a lack of true comparability between the forms despite the superficial use of tanh. Caution around classical interpretations of spectra is warranted, as they may be predicated on elementwise activation geometry, thereby limiting their direct reapplication to isotropic networks.

    These comparisons also continue into the subsequent subsection's hidden-width sweeps.

\subsubsection{Comparison Over Hidden-Layer Widths}

    Comparisons across hidden-layer widths reveal how the spectra generalise and differ between architectures. Moreover, neuroadaptation can explore a broad range of hidden widths, so establishing whether thresholding remains robust across these changes is essential, particularly if the spectrum densifies and removes clear band-gaps between singular values in the isotropic case, which some thresholding approaches may rely upon.

    We use the results from the prior subsection as examples with a hidden width of $100$, while this section also includes hidden widths of $500$ and $1000$ for the CIFAR-10 classifiers. All other training conditions are identical.

\begin{figure}[htb]
        \centering
        \includegraphics[width=\textwidth]{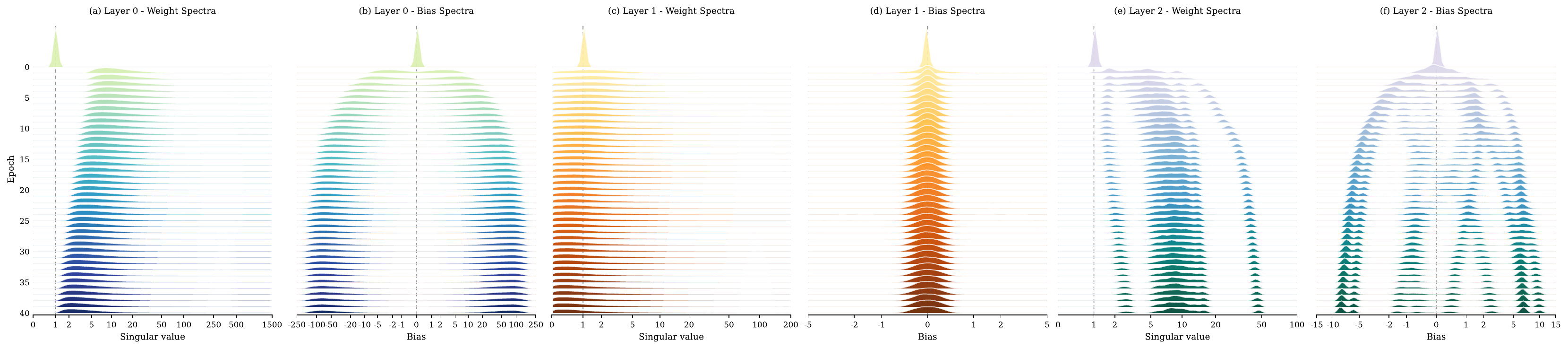}
        \caption{Experiment Two (\textit{App.}~\ref{App:ExperimentTwo}) evolving layerwise singular value and bias spectra for a $\left[3072, 500, 500, 10\right]$ \textbf{isotropic-tanh} CIFAR-10 classifier.}
        \label{Fig:IsoSpectra500}
    \end{figure}

\begin{figure}[htb]
        \centering
        \includegraphics[width=\textwidth]{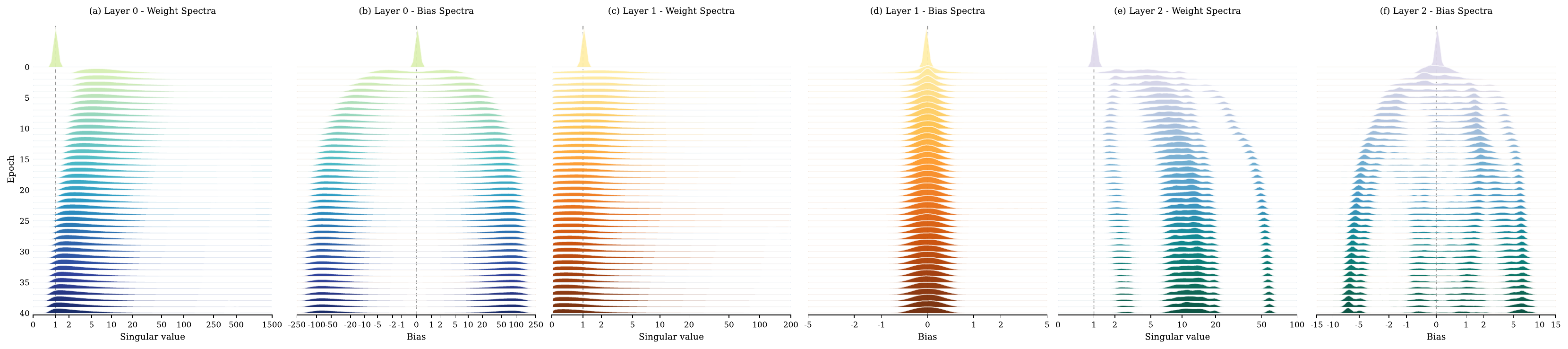}
        \caption{Experiment Two (\textit{App.}~\ref{App:ExperimentTwo}) evolving layerwise singular value and bias spectra for a $\left[3072, 1000, 1000, 10\right]$ \textbf{isotropic-tanh} CIFAR-10 classifier.}
        \label{Fig:IsoSpectra1000}
    \end{figure}

    For the isotropic case, comparing across \textit{Figs.}~\ref{Fig:IsoSpectra100},~\ref{Fig:IsoSpectra500}~and~\ref{Fig:IsoSpectra1000}, the overarching trends are stable and consistent with those observed for \textit{Fig.}~\ref{Fig:IsoSpectra100}. The distinct band behaviour remains pronounced, with comparable specialisation in layer-zero and two's singular values and biases. What differs is the proportion of the spectrum in the singular value bulk of layers zero and one. This suggests that as layer width increases, most new singular values join the main bulk, which shows more pronounced decay-to-zero behaviour with greater width. The epoch-wise scaling then suppresses the relative height of the peak of stable singular values in the more-positive cluster, despite a comparable number of values comprising it.
        
    This is a reassuring result, suggesting that the more-positive cluster is important to the classification task, while the lower-decaying bulk is redundant width that can be neurodegenerated. That lower bulk scales proportionally with width, as one might expect from this hypothesis, as more redundant neurons are introduced. This suggests that regardless of width, isotropic-tanh appears to maintain a comparable solution to the task. The clear band-gap and decaying spectra therefore encourage continued use of thresholding to trigger neurodegeneration in the overly large architecture. This also raises questions regarding the nature of overparameterisation behaviours in isotropic networks.

\begin{figure}[htb]
        \centering
        \includegraphics[width=\textwidth]{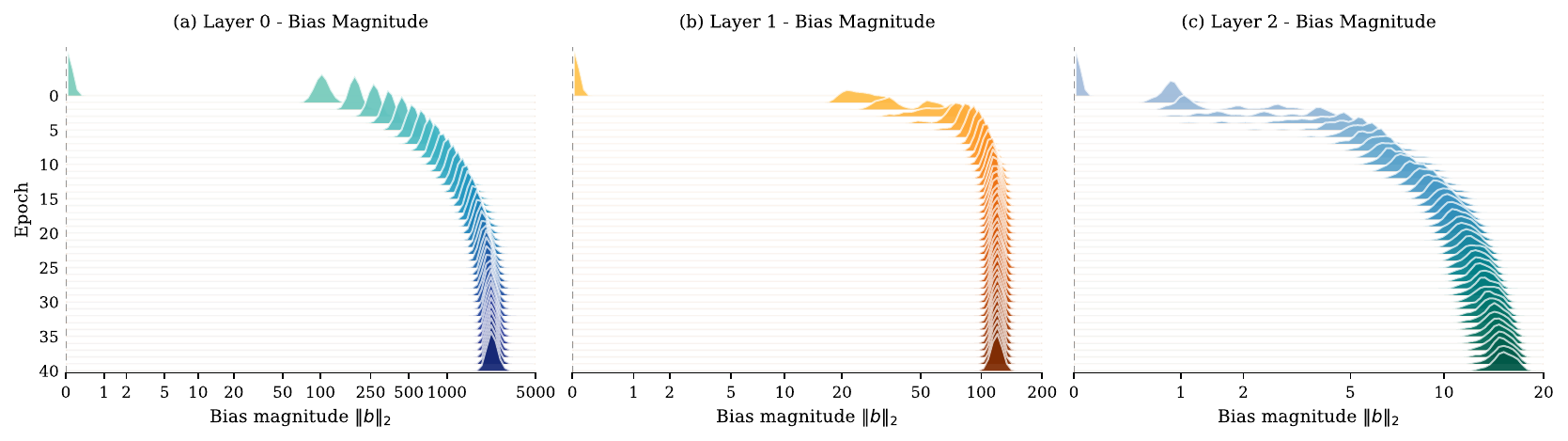}
        \caption{Experiment Two (\textit{App.}~\ref{App:ExperimentTwo}) evolving layerwise bias-magnitude spectra for a $\left[3072, 1000, 1000, 10\right]$ \textbf{isotropic-tanh} CIFAR-10 classifier.}
        \label{Fig:BiasIso1000}
    \end{figure}

    The overall scale of singular values also remains consistent, as does the elementwise histogram of biases; however, bias magnitudes scale with width, indicating that most biases are consistently non-zero and scale roughly as $\approx\sqrt{m}$, as indicated by \textit{Fig.}~\ref{Fig:BiasIso1000} (500-width case comparably interpolates this trend).

\begin{figure}[htb]
        \centering
        \includegraphics[width=\textwidth]{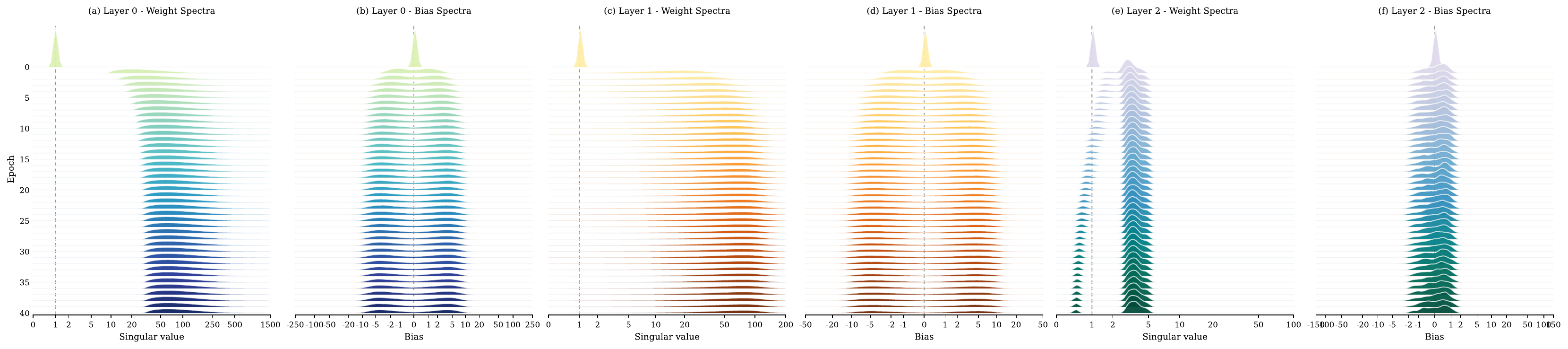}
        \caption{Experiment Two (\textit{App.}~\ref{App:ExperimentTwo}) evolving layerwise singular value and bias spectra for a $\left[3072, 500, 500, 10\right]$ \textbf{standard-tanh} CIFAR-10 classifier.}
        \label{Fig:AnisoSpectra500}
    \end{figure}

\begin{figure}[htb]
        \centering
        \includegraphics[width=\textwidth]{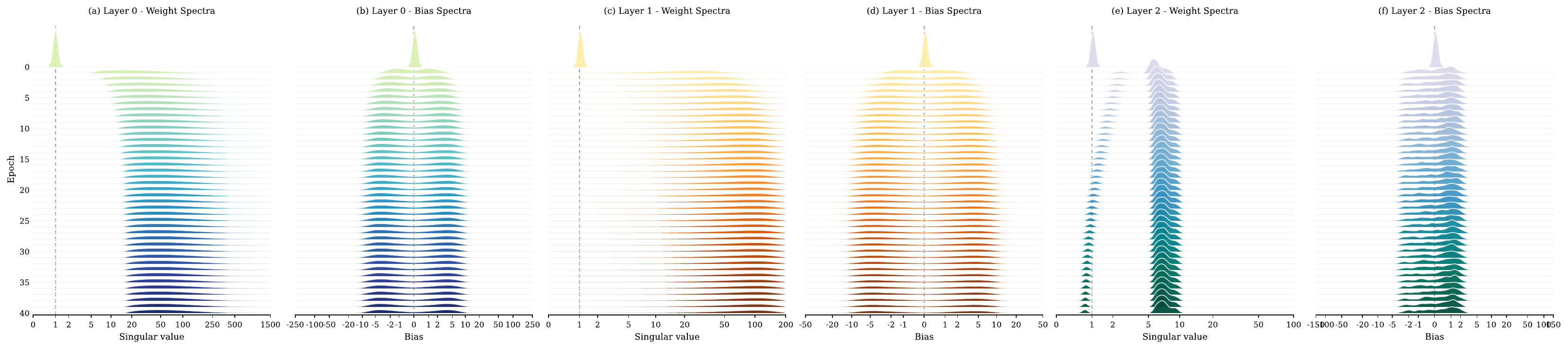}
        \caption{Experiment Two (\textit{App.}~\ref{App:ExperimentTwo}) evolving layerwise singular value and bias spectra for a $\left[3072, 1000, 1000, 10\right]$ \textbf{standard-tanh} CIFAR-10 classifier.}
        \label{Fig:AnisoSpectra1000}
    \end{figure}

    The standard-tanh case (\textit{Figs.}~\ref{Fig:AnisoSpectra100},~\ref{Fig:AnisoSpectra500}~and~\ref{Fig:AnisoSpectra1000}) also reproduces comparable trends for larger widths with some distinctions. Layer-zero's main singular-value bulk coherently decreases in scale with width, while the primary peak of layer-one's and two's bulk increases with scale, with layer-two showing greater separation from its decaying tail. That decaying tail is not proportionally smaller despite increasing width, suggesting that proportions of singular values in each cluster are maintained despite increasing width, unlike isotropic-tanh's enlarging lower-magnitude cluster. This suggests a continued distinction between standard- and isotropic-tanh learning even across differing widths: the latter appears to maintain a similarly functioning subset of singular values, while the former approaches the task via an increasing number of distinct but comparably scaled feature detectors, rather than retaining just the sufficient subset. Additionally, the bias structure becomes increasingly nuanced for layer-two standard-tanh networks.
    
    Overall, these results are encouraging for the use of thresholding, and are used as motivation for its various potential forms (\textit{App.}~\ref{App:Thresholding}). Moreover, they reinforce the distinct incomparability between elementwise and isotropic functional forms, even in the seemingly most similar instances of two tanh-based functions (with identical non-linear maps along basis directions). This encourages a deeper exploration into the biases these symmetry-led primitives may bring to general networks.

    Finally, the next section investigates how the spectra differ for the more complicated Caltech256 dataset across different widths when using isotropic-tanh, with this subsection's results on CIFAR-10 offering an illuminating comparison.

\subsubsection{Comparison to Caltech256 Across Various Hidden-Layer Widths}

\begin{figure}[htb]
        \centering
        \includegraphics[width=\textwidth]{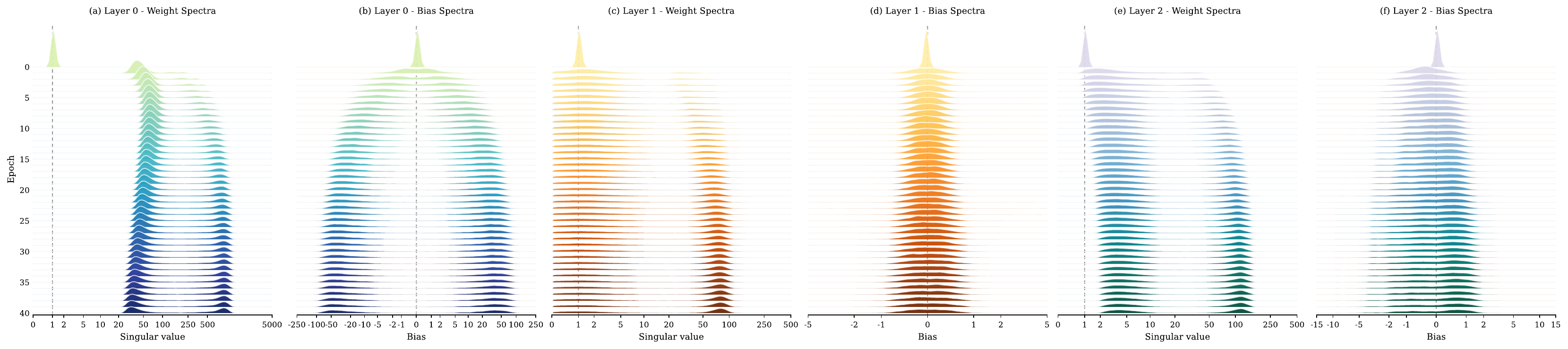}
        \caption{Experiment Two (\textit{App.}~\ref{App:ExperimentTwo}) evolving layerwise singular value and bias spectra for a $\left[12288, 100, 100, 256\right]$ \textbf{isotropic-tanh} Caltech256 classifier.}
        \label{Fig:Caltech256IsoSpectra100}
    \end{figure}

\begin{figure}[htb]
        \centering
        \includegraphics[width=\textwidth]{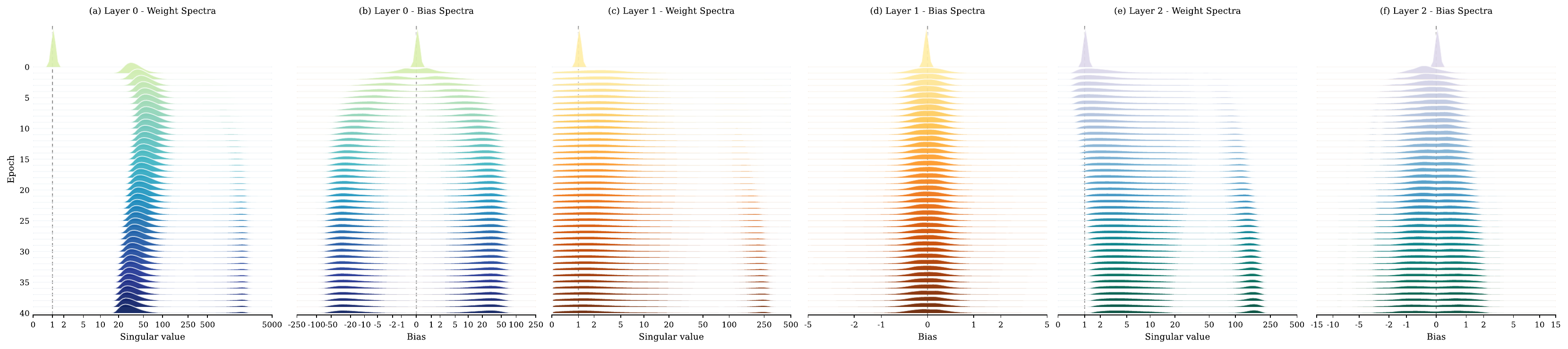}
        \caption{Experiment Two (\textit{App.}~\ref{App:ExperimentTwo}) evolving layerwise singular value and bias spectra for a $\left[12288, 500, 500, 256\right]$ \textbf{isotropic-tanh} Caltech256 classifier.}
        \label{Fig:Caltech256IsoSpectra500}
    \end{figure}

\begin{figure}[htb]
        \centering
        \includegraphics[width=\textwidth]{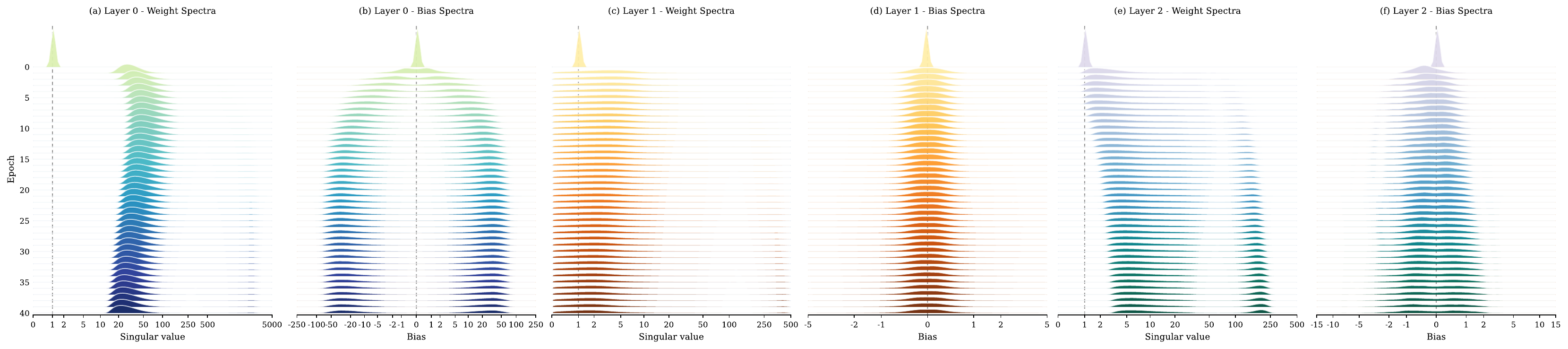}
        \caption{Experiment Two (\textit{App.}~\ref{App:ExperimentTwo}) evolving layerwise singular value and bias spectra for a $\left[12288, 1000, 1000, 256\right]$ \textbf{isotropic-tanh} Caltech256 classifier.}
        \label{Fig:Caltech256IsoSpectra1000}
    \end{figure}

    These Caltech256 networks are trained identically to the CIFAR-10 cases, except for the architectural differences required for the increased number of classes and $64\times64$ pixel RGB inputs instead of CIFAR-10's native $32\times 32$. We also standardised the inputs consistently, so the elementwise scale is identical across the dataset, as in CIFAR-10. Results over three hidden-layer widths are also provided: $\left[12288, m, m, 256\right]$ for $m\in\left\{100, 500, 1000\right\}$ for \textit{Figs.}~\ref{Fig:Caltech256IsoSpectra100}, \ref{Fig:Caltech256IsoSpectra500}, and \ref{Fig:Caltech256IsoSpectra1000} respectively.

    Surprisingly, Caltech256 shows very comparable overarching behaviour to CIFAR-10: layer-zero singular values have a decaying bulk with a distinct, stable, more-positive cluster remaining (albeit with somewhat sharper clusters for Caltech256); the same holds for layer-one and layer-two singular values. Crucially, this structure is now evident across all layers, whereas the small layer-two 10-singular value distribution of CIFAR-10 obscured this repeated structure, which the 256-singular value distribution doesn't. Thus, isotropic-tanh appears to be characterised by a bimodal behaviour wholly distinct from standard-tanh, and it occurs reliably across all layers --- suggesting each reproduces an analogous anisotropic affine scaling for its sub-task. Seemingly, a comparable subset of singular values is reused regardless of initialisation across differing widths for the same task.

\begin{figure}[htb]
        \centering
        \includegraphics[width=\textwidth]{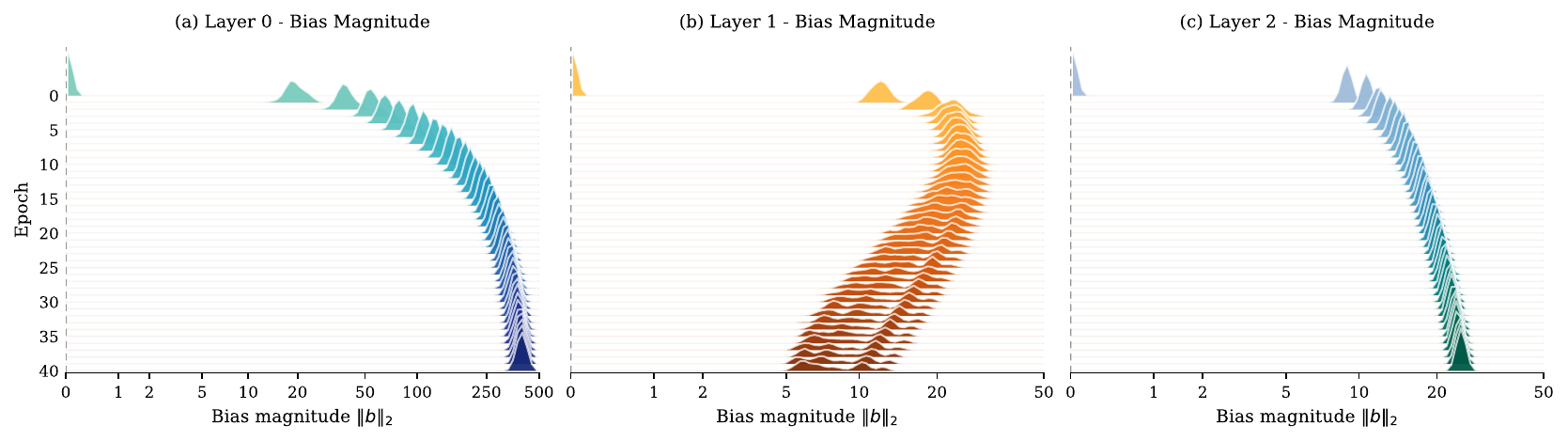}
        \caption{Experiment Two (\textit{App.}~\ref{App:ExperimentTwo}) evolving layerwise bias-magnitude spectra for a $\left[12288, 100, 100, 256\right]$ \textbf{isotropic-tanh} Caltech256 classifier.}
        \label{Fig:Caltech256IsoSpectra100Bias}
    \end{figure}

\begin{figure}[htb]
        \centering
        \includegraphics[width=\textwidth]{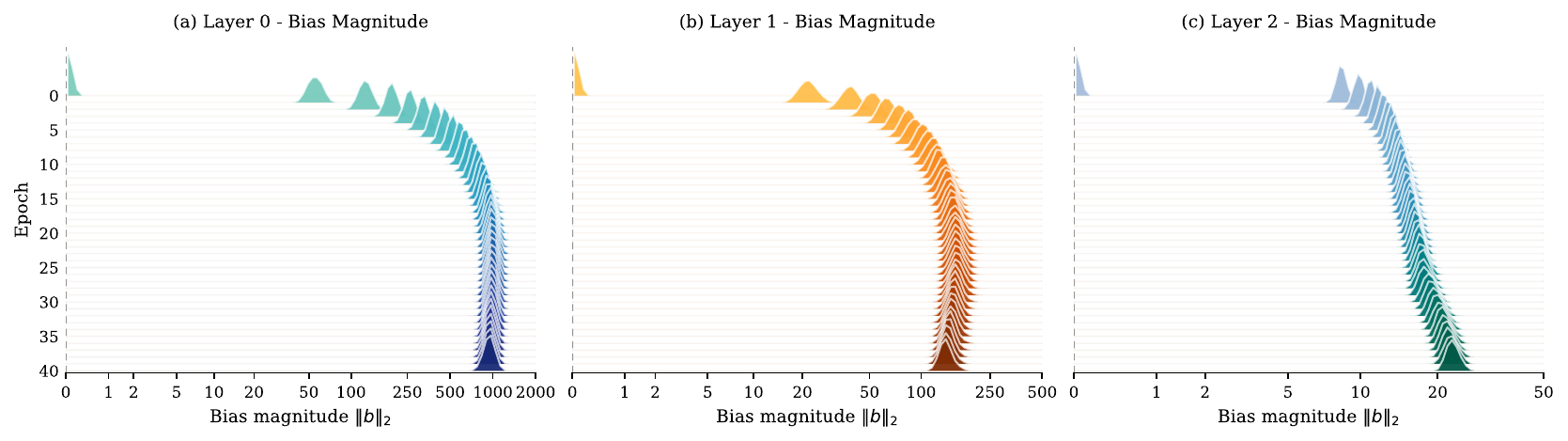}
        \caption{Experiment Two (\textit{App.}~\ref{App:ExperimentTwo}) evolving layerwise bias-magnitude spectra for a $\left[12288, 1000, 1000, 256\right]$ \textbf{isotropic-tanh} Caltech256 classifier.}
        \label{Fig:Caltech256IsoSpectra1000Bias}
    \end{figure}

    Bias structure also appears more nuanced in the Caltech256 case, as shown in \textit{Fig.}~\ref{Fig:Caltech256IsoSpectra100Bias} and \ref{Fig:Caltech256IsoSpectra1000Bias}. The richer structure for the 100-width case may be due to the signed-log x-axis scaling. Curiously, \textit{Fig.}~\ref{Fig:Caltech256IsoSpectra100Bias}'s bias first appears to increase to a stable value, but then eventually begins to increase again, suggesting an evolving metastable solution.

    From this, we conclude an important finding: differences in function form produce fundamentally different modes of learning (even with identical non-linear maps along the axes). Standard-tanh networks are characterised by a single coherent bulk of similarly scaled so-called `feature detectors', whereas, at sufficient width, isotropic-tanh is characterised by a bimodal distribution. This appears to be a reproducible finding across classification networks for different widths and datasets, suggesting an inherent incomparability of how the two functional forms bias a network in its optimised solutions. Thus, with increasing confidence, one can assume that even superficially similar instances of isotropic and elementwise activations behave in fundamentally distinct ways during training and their resultant approach to classification. 
    
    We argue this continues to motivate a broader taxonomic exploration of symmetry-defined primitives and how they bias, helpfully, detrimentally or in a poorly understood but importantly interpretative way (e.g. grandmother neurons \cite{Bird2025c}) general networks, regardless of overall model symmetries or available reparameterisations. Primitive symmetry alone may offer an informative avenue for further research, and these results demonstrate a clear influence on learned representations.

\subsection{Experiment Three: Thresholded Neuroadaptation to Changing Task Schedule\label{App:ExperimentThree}}

    This experiment directly tests the threshold-based approach to neuroadaptation by observing how the isotropic multilayer perceptron adapts to a changing task scheduler. Unlike experiment one (\textit{App.}~\ref{App:ExperimentOne}), neuroadaptation is possible after every epoch, using thresholding to automatically adapt the widths of three different layers. This enables us to determine whether this approach adapts as expected to changing task demands while maintaining competitive task performance relative to a non-adaptive control network.

    Three distinct image-classification datasets are utilised: MNIST \cite{LeCun1998}, Fashion-MNIST (abbreviated to FMNIST) \cite{Xiao2017}, and the `A' to `J' character samples derived from EMNIST \cite{Cohen2017}. During training, all samples are provided as input; the scheduler only determines whether each dataset's cross-entropy term contributes to the overall loss, which the network is optimised against. Hence, the architecture is described by the structure $\left[2352(\!=\!28\!\times\!28\!\times\!3), \text{H1}, \text{H2}, \text{H3}, 30\right]$, where $\text{H1}, \text{H2}, \text{H3}$ are initialised uniformly: $\text{H1}, \text{H2}, \text{H3}\sim U\left(500, 2500\right)$. This was determined to be a reasonable range, producing adaptations for both initial under- and overabundance of neurons. 
    
    It was noted that when there were fewer than $\approx500$ initial neurons, the growth permitted by the $\Xi=5$ scaffold neurons was often insufficient to meet demand, repeatedly triggering neurogenesis with too little time to grow before the next schedule. We chose the singular value thresholding at $\vartheta=0.95$ to quickly neurodegenerate unused parameters, decaying immediately from an initial value of $1$ due to the orthogonal initialisers. Using this, $5$ `scaffold' neurons are maintained as a buffer below this threshold. The $30$-width output corresponds to three disjoint ten-class label blocks, each softmaxed separately as a ten-class block. This structure permits the scheduler to activate a task via a Boolean switch, $S_{\text{dataset}}$, and the total loss is averaged to reduce discontinuity upon task-switching, as shown by \textit{Eqn.}~\ref{Eqn:LossSwitcher}. The plot of \textit{Fig.}~\ref{Fig:DifferentDatasets} utilised the same learning rate as \textit{App.}~\ref{App:ExperimentTwo}, $\eta=5\!\times\!10^{-2}$, such that $150$-epochs per task-schedule was sufficient for parameters to change enough to trigger adaptation.

    \begin{equation}
        \mathcal{L}_{\text{total}} = \frac{S_{\text{MNIST}}\mathcal{L}_{\text{MNIST}}+S_{\text{FMNIST}}\mathcal{L}_{\text{FMNIST}}+S_{\text{EMNIST}}\mathcal{L}_{\text{EMNIST}}}{S_{\text{MNIST}}+S_{\text{FMNIST}}+S_{\text{EMNIST}}}
        \label{Eqn:LossSwitcher}
    \end{equation}

    Using this setup, $25$ independent repeats were taken to produce \textit{Fig.}~\ref{Fig:DifferentDatasets} with a schedule described by \textit{Tab.}~\ref{Tab:Schedule}. Accuracies were still gathered for all datasets, regardless of whether the corresponding loss contributed to the total loss. A static model with identical initialisation (and the same starting random architecture $\left[2352, \text{H1}, \text{H2}, \text{H3}, 30\right]$ ) was used as a control across all repeats and allowed the stated adaptive-versus-control accuracies to be determined at the T3 schedule boundary.

    \begin{table}[htbp]
        \centering
        \caption{Task schedule used in Experiment Three.}
        \label{Tab:Schedule}
        \begin{tabular}{cl}
            \toprule
            Epoch range & Active dataset(s) \\
            \midrule
            $0$--$149$     & MNIST \\
            $150$--$299$   & MNIST + Fashion-MNIST \\
            $300$--$449$   & MNIST + Fashion-MNIST + EMNIST A--J \\
            $450$--$599$   & Fashion-MNIST \\
            \bottomrule
        \end{tabular}
    \end{table}

\begin{figure}[htb]
        \centering
        \includegraphics[width=\textwidth]{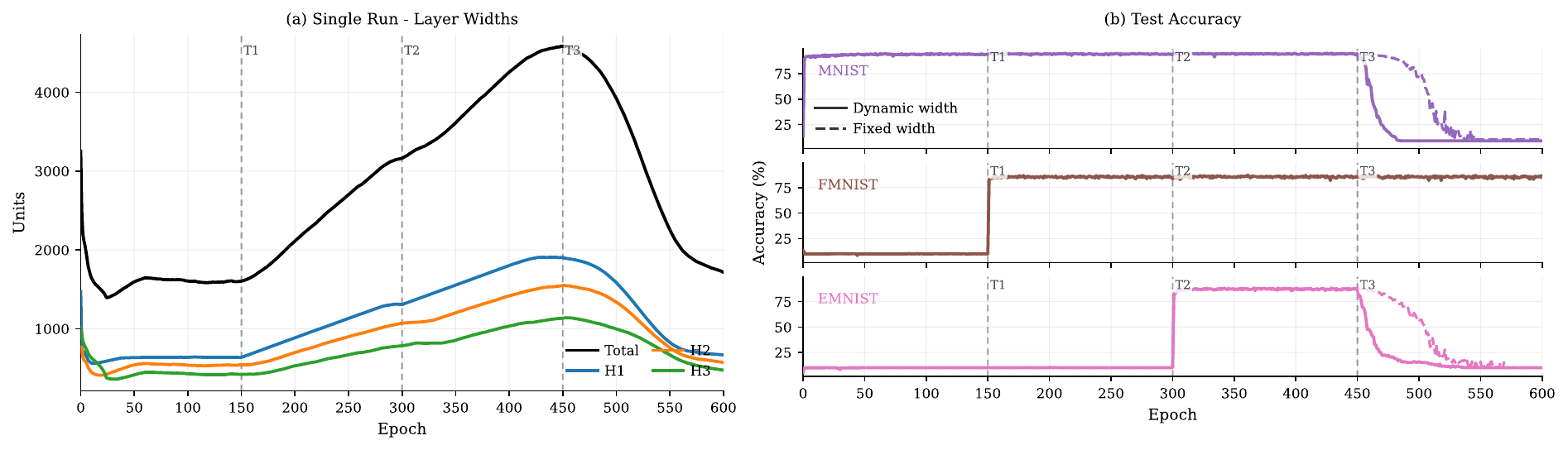}
        \caption{Experiment Three (\textit{App.}~\ref{App:ExperimentThree}) of a single idealistic example}
        \label{Fig:Idealistic1}
    \end{figure}

    \textit{Fig.}~\ref{Fig:Idealistic1} shows an idealistic example: neurogenesis triggers for each successive dataset added, while neurodegeneration occurs when a simpler task is provided. One can observe that neurodegeneration rapidly reduces the unused dataset's accuracies after the scheduler removes them; this suggests it quickly removes the corresponding neurons for those tasks, with test accuracy falls faster than the non-neuroadaptive (fixed) counterpart.

\begin{figure}[htb]
        \centering
        \includegraphics[width=\textwidth]{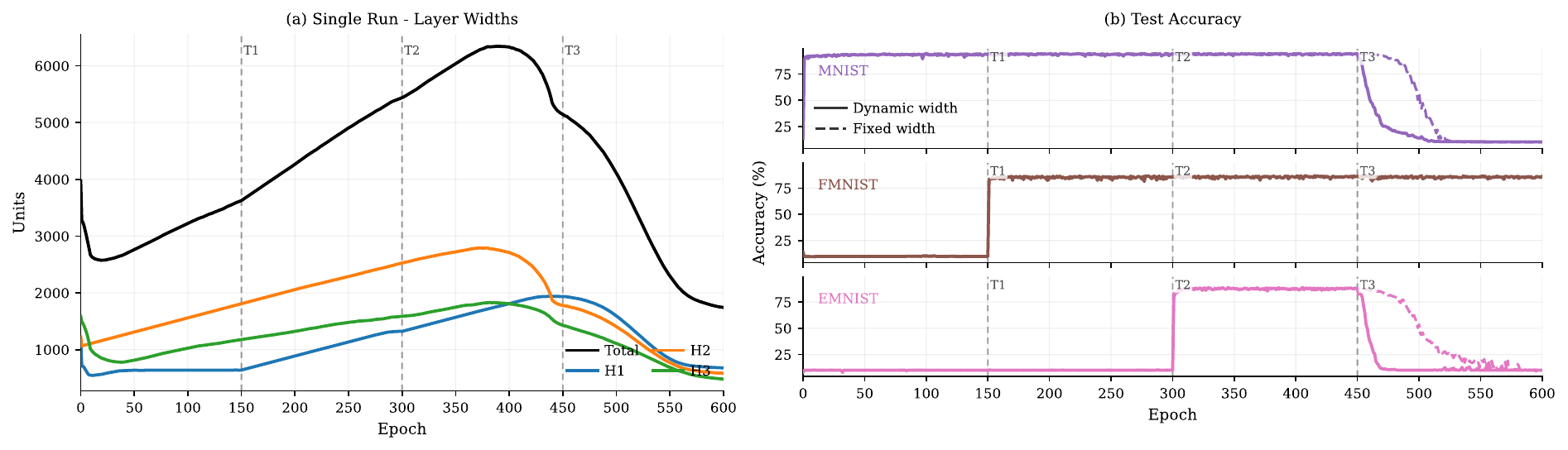}
        \caption{Experiment Three (\textit{App.}~\ref{App:ExperimentThree}) of a single non-idealistic example}
        \label{Fig:Unidealistic1}
    \end{figure}

\begin{figure}[htb]
        \centering
        \includegraphics[width=\textwidth]{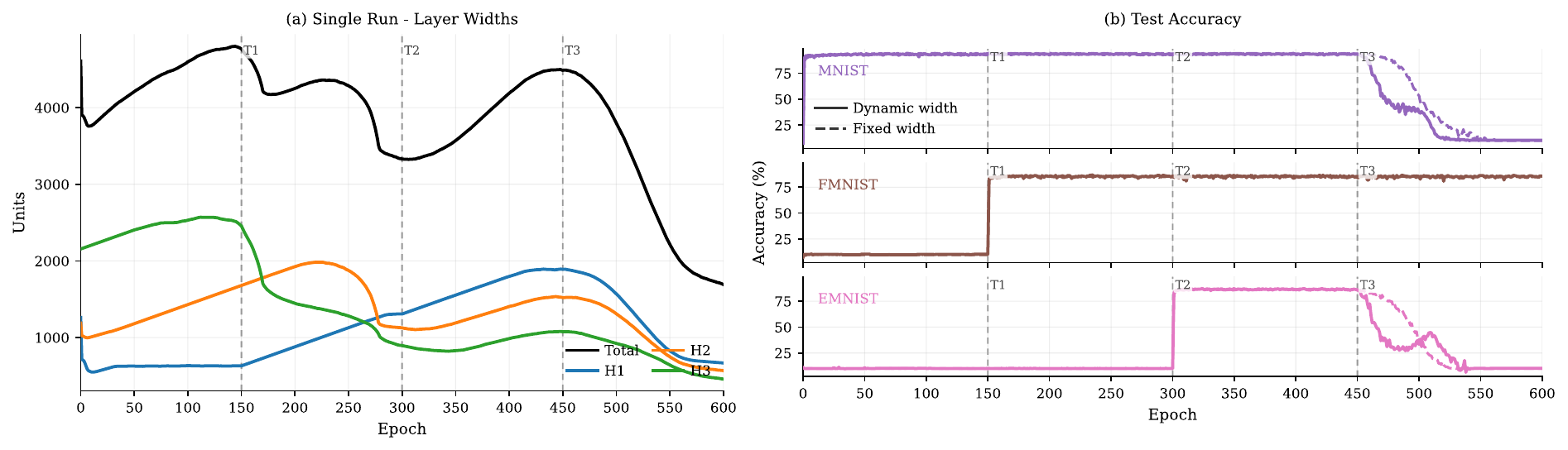}
        \caption{Experiment Three (\textit{App.}~\ref{App:ExperimentThree}) of a single non-idealistic example}
        \label{Fig:Unidealistic2}
    \end{figure}

    Non-prototypical examples also occur, as shown in \textit{Figs.}~\ref{Fig:Unidealistic1}~and~\ref{Fig:Unidealistic2}.

    The example shown in \textit{Fig.}~\ref{Fig:Unidealistic1} shows a layer-wdth rearrangement which occured around epoch 400, causing a $H2>H3>H1$ network to become a classical contracting $H1>H2>H3$ network rapidly. This could have been forced by QR-compression \cite{Ganev2022} (\textit{App.}~\ref{App:Related}), but it is especially interesting that this occurs naturally through the neuroadaptive procedure, showing how it tends toward the more classical construction when architecture flexibility is introduced. \textit{Fig.}~\ref{Fig:Unidealistic2} is another interesting case; although convoluted, it can be understood as transitioning from $H3>H2>H1$ to a temporary $H2>H3>H1$ intermediate, and finally to the classical contracting $H1>H2>H3$. Thus, it also demonstrates this meta-stable architecture behaviour. Its test-accuracy response is also unusual after T3, perhaps suggesting the network's internal representations are not well-organised or separable due to the frequent and considerable architectural rearrangements in its recent history. This is suggestive that internal learned representations may have a `memory-like' property; if its earlier optimisation trajectories became particularly unusual, this disorganisation may persist.

    Similarly, several instances had a consistently growing $H2$ or $H3$ layer, but rarely both concurrently. This suggests that thresholding may be repeatedly triggering neurogenesis under some conditions, such as insufficiently decaying singular values or optimisation increasing the scale of singular values. This motivates exploration of more thresholding schemes, such as those discussed in \textit{App.}~\ref{App:Thresholding}.

\subsection{Experiment Four: Initial Under-to-Over Neuron Abundance\label{App:ExperimentFour}}

    This experiment aimed to determine whether initial neuron under- or over-abundance benefits test accuracy. It did so by training the starting networks for $15$ epochs at the initial width, then gradually (at a linear rate) neuroadapting for $75$ epochs to the desired width, followed by a final $10$ epochs of post-training at the final width. This was undertaken for four repeats per permutation, with all permutations of $n=\left\{7, 10, 25, 50, 75, 100\right\}$ starting and ending widths trialled for a CIFAR-10 classification network with architecture $\left[3072, n, n, 10\right]$. This gives $144$ independent networks trained for this experiment; we report the mean $\pm$ standard error. We added $n=7$ permutations to investigate insufficient-width networks for CIFAR-10. All networks across these experiments use trainable positive intrinsic-length and trainable decaying $\vec{\psi}$ compensations. The CIFAR-10 results are displayed in \textit{Tab.}~\ref{Tab:CIFAR10Neuroabundance}.

    \begin{table}[H]
      \centering
      \caption{Experiment Four (\textit{App.}~\ref{App:ExperimentFour}) shows CIFAR-10 test accuracy for scheduled neuroadaptation between two set widths, $m\!\rightarrow\!n$. }
      \label{Tab:CIFAR10Neuroabundance}
    
      \scriptsize
      \setlength{\tabcolsep}{9pt} 
      \renewcommand{\arraystretch}{1.0} 
    
      \resizebox{\textwidth}{!}{%
      \begin{tabular}{rcccccc}
        \toprule
        Start width $\downarrow$ $\backslash$ End width $\rightarrow$
        & $7$ & $10$ & $25$ & $50$ & $75$ & $100$ \\
        \midrule
        $7$
        & $\best{39.88{\pm}0.36}$
        & $41.65{\pm}0.12$
        & $\best{43.83{\pm}0.02}$
        & $\best{44.37{\pm}0.15}$
        & $43.95{\pm}0.42$
        & $43.91{\pm}0.36$ \\
        $10$
        & $39.64{\pm}0.25$
        & $\best{41.88{\pm}0.36}$
        & $43.59{\pm}0.44$
        & $43.96{\pm}0.19$
        & $\best{44.18{\pm}0.26}$
        & $43.90{\pm}0.47$ \\
        $25$
        & $38.94{\pm}0.30$
        & $41.30{\pm}0.50$
        & $43.41{\pm}0.51$
        & $43.50{\pm}0.34$
        & $43.57{\pm}0.07$
        & $43.74{\pm}0.53$ \\
        $50$
        & $38.84{\pm}0.44$
        & $41.14{\pm}0.41$
        & $43.05{\pm}0.32$
        & $43.43{\pm}0.29$
        & $43.74{\pm}0.25$
        & $43.59{\pm}0.34$ \\
        $75$
        & $38.48{\pm}0.53$
        & $41.32{\pm}0.59$
        & $43.19{\pm}0.43$
        & $43.70{\pm}0.23$
        & $43.12{\pm}0.46$
        & $\best{43.95{\pm}0.17}$ \\
        $100$
        & $39.27{\pm}0.09$
        & $41.19{\pm}0.62$
        & $43.33{\pm}0.13$
        & $43.75{\pm}0.37$
        & $43.68{\pm}0.25$
        & $43.24{\pm}0.54$ \\
        \bottomrule
      \end{tabular}%
      }
    \end{table}

    The CIFAR-10 results show little clear trend, aside from a slight advantage to initial underabundance; however, their significance relative to standard error is poor. Therefore, no particular starting width shows a clear advantage. Despite this, and perhaps more importantly, there is neither disadvantage to any particular starting width either, indicating that neuroadaptation performs adequately with little function loss or impedance to learning. 
    
    The results for Caltech256 (\textit{Tab.}~\ref{Tab:Caltech256Neuroabundance}) show similar inconclusiveness, but with a different distribution of top accuracies and increased significance. Due to the increased 256-class dataset, the network widths were adjusted for all permutations of $n=\left\{32, 128, 256, 384, 512, 640\right\}$ for the architecture $\left[12288, n, n, 256\right]$, trained identically to the CIFAR-10 case, with $15$ initial width epochs, then $75$ epochs for neuroadaption, then $10$ epochs for training at the final width. We also ran four repeats and report the mean $\pm$ standard error. Unlike the CIFAR-10 case, which varied by at most one neuron width per epoch, these Caltech256 networks could vary by up to 9 neurons per epoch. This experiment was designed with widths of $32$ and $128$ estimated to be roughly underabundant, with $512$ and $640$ being roughly overabundant; despite this, the final widths of $32$, $128$ and $640$ neurons reported the highest accuracies.
    
    \begin{table}[H]
      \centering
      \caption{Experiment Four (\textit{App.}~\ref{App:ExperimentFour}) shows Caltech256 64$\times$64 test accuracy for scheduled neuroadaptation between two set widths, $m\!\rightarrow\!n$. This generally indicates an adaption benefit when $m\!\neq\!n$.}
      \label{Tab:Caltech256Neuroabundance}
    
      \scriptsize
      \setlength{\tabcolsep}{9pt}
      \renewcommand{\arraystretch}{1.0} 
    
      \resizebox{\textwidth}{!}{%
      \begin{tabular}{rcccccc}
        \toprule
        Start width $\downarrow$ $\backslash$ End width $\rightarrow$
        & $32$ & $128$ & $256$ & $384$ & $512$ & $640$ \\
        \midrule
        $32$
        & $\best{12.42{\pm}0.15}$
        & $\best{11.85{\pm}0.16}$
        & $11.44{\pm}0.11$
        & $11.53{\pm}0.20$
        & $11.37{\pm}0.28$
        & $11.26{\pm}0.10$ \\
        $128$
        & $10.26{\pm}0.13$
        & $11.25{\pm}0.28$
        & $11.09{\pm}0.12$
        & $11.35{\pm}0.14$
        & $11.31{\pm}0.13$
        & $11.29{\pm}0.17$ \\
        $256$
        & $10.17{\pm}0.18$
        & $11.18{\pm}0.03$
        & $10.68{\pm}0.05$
        & $11.45{\pm}0.05$
        & $\best{11.47{\pm}0.31}$
        & $11.35{\pm}0.26$ \\
        $384$
        & $10.27{\pm}0.18$
        & $11.02{\pm}0.19$
        & $11.40{\pm}0.26$
        & $10.76{\pm}0.13$
        & $11.30{\pm}0.23$
        & $11.57{\pm}0.10$ \\
        $512$
        & $10.33{\pm}0.07$
        & $11.11{\pm}0.10$
        & $\best{11.49{\pm}0.17}$
        & $11.33{\pm}0.05$
        & $10.81{\pm}0.18$
        & $\best{11.87{\pm}0.09}$ \\
        $640$
        & $10.69{\pm}0.18$
        & $11.42{\pm}0.16$
        & $11.33{\pm}0.20$
        & $\best{11.61{\pm}0.16}$
        & $11.45{\pm}0.09$
        & $11.06{\pm}0.20$ \\
        \bottomrule
      \end{tabular}%
      }
    \end{table}

    These results demonstrate a more complicated relationship between starting and ending width, with no clear preference for neuro under- or overabundance. This is somewhat surprising given the mammalian approach to initial overabundance \cite{Scholl2021}, which is not reflected in the artificial neural network case---at least for isotropic adaptive MLPs. Curiously, results of $256$, $384$, $512$ and $640$ all show significant underperformance when there is no neuroadaptation (start width equal to end width). This intriguing result suggests a benefit of neuroadaptation during training and reinforces that these processes are not implicitly detrimental. However, the highest results were distributed evenly between $\texttt{initial-width}<\texttt{final-width}$ and $\texttt{initial-width}>\texttt{final-width}$, leading to an inconclusive preference for the direction in which adaptation helps. Overall, this is an encouraging result for neuroadaptation, particularly neurodegeneration, which is unavoidably a lossy process, as the network demonstrates resilience and even a benefit under such adaptations. 
    
    In conclusion, this experiment was less revealing than others, showing less clear or statistically significant trends. However, this suggests that the neuroadaptive methodology successfully preserved function path-independently, since dynamic networks achieved similar or higher performance than the fixed-width controls. This remains a positive indication of successful neuroadaptation, despite not replicating the overabundance benefit observed in mammalian neural networks.

\subsection{Experiment Five: Sparsification\label{App:ExperimentFive}}

    In \textit{Sec.}~\ref{Sec:Sparsification}, we detail parameter reduction using the full-diagonalisation approach. This sparsification approach is analytically function-preserving; we demonstrate this function retention numerically in this subsection alongside its parameter reductions. This complements the $k=0$ results of \textit{App.}~\ref{App:ExperimentOne}.
    
    These results verify the theoretical claims, showing that a substantial parameter reduction can be achieved with negligible function loss. The derived formulae show that the reduction can asymptotically achieve $50\%$ sparsity at infinite width and depth, while this section empirically details various finite-width and depth architectures and the corresponding practical sparsification fractions achieved.

    To ensure initial functionality, these networks are trained with the aforementioned hyperparameters stated at the beginning of \textit{App.}~\ref{App:Experiments}. Results span CIFAR-10, CIFAR-100 and Caltech256 classification datasets over $5$ independent repeats (mean and standard error reported), each trained for $25$ epochs using trainable positive intrinsic length and trainable decaying $\vec{\psi}$ compensation. Good test accuracy does not matter to show sparsification and function preservation beyond a reasonable output sensitivity to the input samples. We provide a variety of width and depth architectures.

    It is also important to clarify that \textit{Sec.}~\ref{Sec:Sparsification}'s \textit{Eqns.}~\ref{Eqn:OddLength}~and~\ref{Eqn:EvenLength} count just the weight and bias parameters; however, the networks below utilise intrinsic length and $\vec{\psi}$ additionally. Hence, we detail a slight modification to account for these additional parameters in \textit{Eqn.}~\ref{Eqn:OddLength2} for odd-length networks and \textit{Eqn.}~\ref{Eqn:EvenLength2} for even-length networks. These additional parameters contribute $+N+1$ parameters per layer, which cannot be diagonalised and therefore persist in the numerator and denominator. These do not affect the stated asymptotic results and apply to constant-width networks, as $D, N\rightarrow\infty$ then $S_{\text{odd}}, S_{\text{even}}\rightarrow0.5$. 

    \begin{align}
        S_{\text{odd}} &= \frac{D\left(3N+1\right)+\left(D+1\right)\left(N^2+2N+1\right)}{\left(2D+1\right)\left(N^2+2N+1\right)}&&=\frac{1+D^{-1}}{2+D^{-1}}+
        \frac{D\left(3N+1\right)}{\left(2D+1\right)\left(N+1\right)^2}
        \label{Eqn:OddLength2}\\
        S_{\text{even}} &= \frac{\left(D-1\right)\left(3N+1\right)+\left(D+1\right)\left(N^2+2N+1\right)}{2D\left(N^2+2N+1\right)}&&=\frac{1+D^{-1}}{2}+
        \frac{\left(D-1\right)\left(3N+1\right)}{2D\left(N+1\right)^2}\label{Eqn:EvenLength2}
    \end{align}

    \begin{table}[htbp]
\centering
\caption{Post-training sparsification of Caltech-256 classifiers. Architectures are notated compactly as $w\!\times\!k$ for $k$ hidden layers of width $w$. $\Delta a$ is post-sparsification minus pre-sparsification accuracy in percentage points. Parameters count non-zero parameters, using a numerical threshold of $|\theta|>0.0001$. $\epsilon_{\mathrm{rel}}$ is \textit{Eqn.}~\ref{Eqn:RelativeChange}, and $D_{\mathrm{KL}}$ is \textit{Eqn.}~\ref{Eqn:KLmetric}. Values reported as mean $\pm$ standard error over 5 repeats.}
\scriptsize
\setlength{\tabcolsep}{3.5pt}
\renewcommand{\arraystretch}{1.12}
\resizebox{\textwidth}{!}{%
\begin{tabular}{lccccccc}
\toprule
Architecture & Acc. before & $\Delta a$ ($\!\times\!10^{-2}\,\mathrm{pp}$) & Params before & Params after & \% Reduced & $\epsilon_{\mathrm{rel}}$ ($\!\times\!10^{-2}\%$) & $D_{\mathrm{KL}}$ ($\!\times\!10^{-8}$) \\
\midrule
$[12288,500\!\times\!2,256]$ & $15.8\!\pm\!0.1$ & $+0.0\!\pm\!0.0$ & $6.52\mathrm{M}$ & $6.26\mathrm{M}$ & $4.1\%$ & $0.6\!\pm\!0.0$ & $0.4\!\pm\!0.1$ \\
$[12288,500\!\times\!3,256]$ & $15.2\!\pm\!0.1$ & $+0.0\!\pm\!0.0$ & $6.77\mathrm{M}$ & $6.51\mathrm{M}$ & $3.9\%$ & $0.6\!\pm\!0.0$ & $0.8\!\pm\!0.1$ \\
$[12288,500\!\times\!4,256]$ & $14.0\!\pm\!0.2$ & $-0.3\!\pm\!0.3$ & $7.02\mathrm{M}$ & $6.51\mathrm{M}$ & $7.3\%$ & $0.9\!\pm\!0.1$ & $1.5\!\pm\!0.5$ \\
$[12288,500\!\times\!5,256]$ & $12.7\!\pm\!0.1$ & $+0.3\!\pm\!0.3$ & $7.27\mathrm{M}$ & $6.76\mathrm{M}$ & $7.0\%$ & $0.5\!\pm\!0.0$ & $0.5\!\pm\!0.1$ \\
$[12288,500\!\times\!6,256]$ & $11.0\!\pm\!0.3$ & $-0.3\!\pm\!0.3$ & $7.52\mathrm{M}$ & $6.76\mathrm{M}$ & $10.1\%$ & $0.9\!\pm\!0.1$ & $1.7\!\pm\!0.4$ \\
$[12288,1000\!\times\!2,256]$ & $15.6\!\pm\!0.2$ & $-0.3\!\pm\!0.3$ & $13.54\mathrm{M}$ & $12.50\mathrm{M}$ & $7.7\%$ & $1.2\!\pm\!0.1$ & $1.9\!\pm\!0.2$ \\
$[12288,1000\!\times\!3,256]$ & $14.9\!\pm\!0.1$ & $+0.0\!\pm\!0.0$ & $14.54\mathrm{M}$ & $13.50\mathrm{M}$ & $7.1\%$ & $0.6\!\pm\!0.1$ & $0.8\!\pm\!0.2$ \\
$[12288,1000\!\times\!4,256]$ & $13.5\!\pm\!0.2$ & $+0.0\!\pm\!0.0$ & $15.54\mathrm{M}$ & $13.51\mathrm{M}$ & $13.0\%$ & $1.3\!\pm\!0.0$ & $2.1\!\pm\!0.2$ \\
$[12288,1000\!\times\!5,256]$ & $12.6\!\pm\!0.2$ & $-0.3\!\pm\!0.3$ & $16.54\mathrm{M}$ & $14.52\mathrm{M}$ & $12.2\%$ & $1.0\!\pm\!0.1$ & $1.9\!\pm\!0.4$ \\
$[12288,1000\!\times\!6,256]$ & $10.9\!\pm\!0.2$ & $+0.3\!\pm\!0.3$ & $17.54\mathrm{M}$ & $14.52\mathrm{M}$ & $17.2\%$ & $1.3\!\pm\!0.2$ & $2.2\!\pm\!0.7$ \\
\bottomrule
\end{tabular}%
}
\label{Tab:Caltech256Sparsification}
\end{table}

    Due to the larger initial layer for \textit{Tab.}~\ref{Tab:Caltech256Sparsification}, the sparsification is typically smaller than that achieved by \textit{Tabs.}~\ref{Tab:CIFAR10sparsification}~and~\ref{Tab:CIFAR100sparsification}. As expected, increasing width and depth increases the percentage sparsified. Most importantly, all results verify finite-width sparsity with function invariance, demonstrating negligible change in the final-layer output activations with an order of $10^{-8}$ KL-divergence in output distributions. Output depth appears to amplify small numerical differences, as previously expected, but these are often insignificant given the associated errors. Function drift, measured by $\epsilon_{\text{rel.}}$, is on order of $0.01\%$, and test-accuracy drift is similarly around $0.01\%$. This is approximately 1 sample classified differently in a 10000-sized test set (the test-set size of CIFAR-10 and CIFAR-100). This is consistent with a remarkable near-identical functionality, up to numerics and propagation thereof, during significant alternating-layer sparsification. Thus, one can reasonably assert that these networks are capable of being sparsified drastically without consequence on performance; moreover, function. 
    
    \begin{table}[htbp]
    \centering
   \caption{Post-training sparsification of CIFAR-100 classifiers. Architectures are notated compactly as $w\!\times\!k$ for $k$ hidden layers of width $w$. $\Delta a$ is post-sparsification minus pre-sparsification accuracy in percentage points. Parameters count non-zero parameters, using a numerical threshold of $|\theta|>0.0001$. $\epsilon_{\mathrm{rel}}$ is \textit{Eqn.}~\ref{Eqn:RelativeChange}, and $D_{\mathrm{KL}}$ is \textit{Eqn.}~\ref{Eqn:KLmetric}. Values reported as mean $\pm$ standard error over 5 repeats.}
    \scriptsize
    \setlength{\tabcolsep}{3.5pt}
    \renewcommand{\arraystretch}{1.12}
    \resizebox{\textwidth}{!}{%
    \begin{tabular}{lccccccc}
    \toprule
    Architecture & Acc. before & $\Delta a$ ($\!\times\!10^{-2}\,\mathrm{pp}$) & Params before & Params after & \% Reduced & $\epsilon_{\mathrm{rel}}$ ($\!\times\!10^{-2}\%$) & $D_{\mathrm{KL}}$ ($\!\times\!10^{-8}$) \\
    \midrule
    $[3072,500\!\times\!2,100]$ & $19.0\!\pm\!0.1$ & $+0.2\!\pm\!0.2$ & $1.84\mathrm{M}$ & $1.58\mathrm{M}$ & $13.7\%$ & $0.7\!\pm\!0.0$ & $0.6\!\pm\!0.0$ \\
    $[3072,500\!\times\!3,100]$ & $18.9\!\pm\!0.1$ & $-0.2\!\pm\!0.4$ & $2.09\mathrm{M}$ & $1.84\mathrm{M}$ & $12.1\%$ & $1.2\!\pm\!0.1$ & $2.5\!\pm\!0.3$ \\
    $[3072,500\!\times\!4,100]$ & $17.5\!\pm\!0.2$ & $-0.8\!\pm\!0.6$ & $2.34\mathrm{M}$ & $1.84\mathrm{M}$ & $21.4\%$ & $1.5\!\pm\!0.1$ & $3.5\!\pm\!0.4$ \\
    $[3072,500\!\times\!5,100]$ & $15.2\!\pm\!0.3$ & $+0.4\!\pm\!0.4$ & $2.59\mathrm{M}$ & $2.09\mathrm{M}$ & $19.4\%$ & $1.3\!\pm\!0.1$ & $2.9\!\pm\!0.2$ \\
    $[3072,500\!\times\!6,100]$ & $14.0\!\pm\!0.1$ & $+0.2\!\pm\!0.5$ & $2.84\mathrm{M}$ & $2.09\mathrm{M}$ & $26.4\%$ & $1.4\!\pm\!0.1$ & $3.2\!\pm\!0.5$ \\
    $[3072,1000\!\times\!2,100]$ & $18.9\!\pm\!0.1$ & $+0.4\!\pm\!0.2$ & $4.17\mathrm{M}$ & $3.17\mathrm{M}$ & $24.1\%$ & $1.2\!\pm\!0.0$ & $1.7\!\pm\!0.1$ \\
    $[3072,1000\!\times\!3,100]$ & $18.4\!\pm\!0.1$ & $-0.2\!\pm\!0.2$ & $5.17\mathrm{M}$ & $4.17\mathrm{M}$ & $19.4\%$ & $1.3\!\pm\!0.1$ & $3.4\!\pm\!0.8$ \\
    $[3072,1000\!\times\!4,100]$ & $16.9\!\pm\!0.1$ & $+0.6\!\pm\!0.5$ & $6.17\mathrm{M}$ & $4.17\mathrm{M}$ & $32.5\%$ & $1.9\!\pm\!0.1$ & $5.5\!\pm\!0.7$ \\
    $[3072,1000\!\times\!5,100]$ & $14.9\!\pm\!0.2$ & $-0.2\!\pm\!0.4$ & $7.17\mathrm{M}$ & $5.17\mathrm{M}$ & $27.9\%$ & $2.5\!\pm\!0.3$ & $12.1\!\pm\!3.4$ \\
    $[3072,1000\!\times\!6,100]$ & $13.9\!\pm\!0.1$ & $+0.2\!\pm\!0.2$ & $8.17\mathrm{M}$ & $5.17\mathrm{M}$ & $36.7\%$ & $2.3\!\pm\!0.2$ & $8.6\!\pm\!2.2$ \\
    \bottomrule
    \end{tabular}%
    }
    \label{Tab:CIFAR100sparsification}
\end{table}
    
    \begin{table}[htbp]
    \centering
    \caption{Post-training sparsification of CIFAR-10 classifiers. Architectures are notated compactly as $w\!\times\!k$ for $k$ hidden layers of width $w$. $\Delta a$ is post-sparsification minus pre-sparsification accuracy in percentage points. Parameters count non-zero parameters, using a numerical threshold of $|\theta|>0.0001$. $\epsilon_{\mathrm{rel}}$ is \textit{Eqn.}~\ref{Eqn:RelativeChange}, and $D_{\mathrm{KL}}$ is \textit{Eqn.}~\ref{Eqn:KLmetric}. Values reported as mean $\pm$ standard error over 5 repeats.}
    \scriptsize
    \setlength{\tabcolsep}{3.5pt}
    \renewcommand{\arraystretch}{1.12}
    \resizebox{\textwidth}{!}{%
    \begin{tabular}{lccccccc}
    \toprule
    Architecture & Acc. before & $\Delta a$ ($\!\times\!10^{-2}\,\mathrm{pp}$) & Params before & Params after & \% Reduced & $\epsilon_{\mathrm{rel}}$ ($\!\times\!10^{-2}\%$) & $D_{\mathrm{KL}}$ ($\!\times\!10^{-8}$) \\
    \midrule
    $[3072,500\!\times\!2,10]$ & $43.1\!\pm\!0.2$ & $-0.4\!\pm\!0.2$ & $1.79\mathrm{M}$ & $1.54\mathrm{M}$ & $14.0\%$ & $1.3\!\pm\!0.2$ & $1.3\!\pm\!0.4$ \\
    $[3072,500\!\times\!3,10]$ & $44.9\!\pm\!0.2$ & $-0.2\!\pm\!0.4$ & $2.04\mathrm{M}$ & $1.79\mathrm{M}$ & $12.3\%$ & $2.8\!\pm\!0.2$ & $6.0\!\pm\!0.7$ \\
    $[3072,500\!\times\!4,10]$ & $45.4\!\pm\!0.1$ & $+0.0\!\pm\!0.3$ & $2.29\mathrm{M}$ & $1.79\mathrm{M}$ & $21.8\%$ & $2.3\!\pm\!0.1$ & $5.6\!\pm\!0.7$ \\
    $[3072,500\!\times\!5,10]$ & $45.1\!\pm\!0.2$ & $+0.4\!\pm\!0.5$ & $2.54\mathrm{M}$ & $2.04\mathrm{M}$ & $19.7\%$ & $1.8\!\pm\!0.2$ & $3.3\!\pm\!0.6$ \\
    $[3072,500\!\times\!6,10]$ & $42.8\!\pm\!0.3$ & $+0.6\!\pm\!0.6$ & $2.79\mathrm{M}$ & $2.04\mathrm{M}$ & $26.9\%$ & $2.2\!\pm\!0.5$ & $4.4\!\pm\!2.3$ \\
    $[3072,1000\!\times\!2,10]$ & $43.2\!\pm\!0.2$ & $-0.2\!\pm\!0.5$ & $4.08\mathrm{M}$ & $3.08\mathrm{M}$ & $24.6\%$ & $1.8\!\pm\!0.2$ & $2.2\!\pm\!0.6$ \\
    $[3072,1000\!\times\!3,10]$ & $44.8\!\pm\!0.2$ & $-0.8\!\pm\!0.4$ & $5.08\mathrm{M}$ & $4.08\mathrm{M}$ & $19.7\%$ & $4.4\!\pm\!0.4$ & $14.0\!\pm\!2.3$ \\
    $[3072,1000\!\times\!4,10]$ & $45.2\!\pm\!0.1$ & $+0.4\!\pm\!0.6$ & $6.08\mathrm{M}$ & $4.08\mathrm{M}$ & $32.9\%$ & $3.1\!\pm\!0.3$ & $10.7\!\pm\!2.1$ \\
    $[3072,1000\!\times\!5,10]$ & $43.7\!\pm\!0.4$ & $+0.2\!\pm\!0.5$ & $7.08\mathrm{M}$ & $5.08\mathrm{M}$ & $28.3\%$ & $3.0\!\pm\!0.4$ & $7.4\!\pm\!1.5$ \\
    $[3072,1000\!\times\!6,10]$ & $41.8\!\pm\!0.1$ & $-0.8\!\pm\!0.2$ & $8.08\mathrm{M}$ & $5.08\mathrm{M}$ & $37.1\%$ & $3.3\!\pm\!0.6$ & $7.9\!\pm\!2.7$ \\
    \bottomrule
    \end{tabular}%
    }
    \label{Tab:CIFAR10sparsification}
\end{table}

    Overall, the networks of finite width and depth still achieve a $3.9\%$ to $37.1\%$ parameter reduction, without meaningful consequences. The alternating diagonalised architecture is also more computationally efficient in the forward pass, as diagonalisation yields a Hadamard product rather than full matrix multiplication. This yields $\mathcal{O}\left(n\right)$ over the standard $\mathcal{O}\left(mn\right)$ for a layer $\vec{y}=\mathbf{A}\vec{x}$, with $\mathbf{A}\in\mathbb{R}^{m\times n}$, $\vec{x}\in\mathbb{R}^n$ and $\vec{y}\in\mathbb{R}^m$ for $n=\min(m,n)$. Thus, this technique can be used before inference to obtain a more computationally and parameter-efficient isotropic multilayer perceptron architecture, with alternating layers scaling linearly in memory and compute rather than quadratically. These non-conventional lossless pruning results are a benefit of isotropic MLPs generally, independently of neuroadaptation, and can be similarly applied to convolution networks following \textit{App.}~\ref{App:Convolution}'s sketch.

\subsection{Experiment Six: Isotropic-Tanh Against Standard-Tanh Trends\label{App:ExperimentSix}}

    Already established in the previous sections (particularly \textit{App.}~\ref{App:ExperimentTwo}) is that standard-tanh, $\mathbf{f}\left(\vec{x};\left\{\hat{e}_i\right\}\right)=\sum_i\tanh\left(\hat{e}_i^T\vec{x}\right)\hat{e}_i$, does not operate comparably to isotropic-tanh, $\mathbf{f}\left(\vec{x}\right)=\tanh\left(\left\|\vec{x}\right\|_2\right)\hat{x}$, despite their superficial resemblance and function agreement for vectors colinear with basis directions, $\hat{e}_i$. 
    
    Nevertheless, comparisons can be undertaken across various datasets, but one should not assess either's broader functional form on a single instance alone. We stress that even for superficially similar functions, they are demonstrated to be inherently dissimilar.
    
    The results below consist of four repeats for each network-activation function permutation, across datasets of arguably increasing difficulty: MNIST, A-J EMNIST, Fashion-MNIST, CIFAR-10, CIFAR-100 and Caltech-256 --- each trained for $50$ epochs under otherwise identical conditions. The architecture column specifies the \textit{fixed} (non-neuroadapted) hidden-layer width $W$ and the number of hidden layers $D$ ($W\times D$), with input and output dimensions constrained by their respective datasets.

    \begin{table}[htbp]
    \centering
    \caption{Experiment Six: displays the final test-set accuracy for standard-tanh and isotropic-tanh for otherwise identical multilayer perceptrons across various datasets and architectures. Values are reported as mean $\pm$ standard error across $5$ repeats. Boxed indicates the column-wise highest accuracy (e.g., best isotropic-tanh performance across architectures, and likewise for standard-tanh). Bold indicates the pairwise highest accuracy (e.g., which was best between standard-tanh and isotropic-tanh).}
    \label{Tab:ExperimentEightProgress}

    \scriptsize
    \setlength{\tabcolsep}{2.2pt}
    \renewcommand{\arraystretch}{1.15}

    \resizebox{\textwidth}{!}{%
    \begin{tabular}{l cc@{\hspace{12pt}} cc@{\hspace{12pt}} cc@{\hspace{12pt}} cc@{\hspace{12pt}} cc@{\hspace{12pt}} cc}
        \toprule
        Architecture
        & \multicolumn{2}{c}{MNIST}
        & \multicolumn{2}{c}{EMNIST\_AJ}
        & \multicolumn{2}{c}{FMNIST}
        & \multicolumn{2}{c}{CIFAR-10}
        & \multicolumn{2}{c}{CIFAR-100}
        & \multicolumn{2}{c}{CALTECH-256} \\
        \cmidrule(lr){2-3}
        \cmidrule(lr){4-5}
        \cmidrule(lr){6-7}
        \cmidrule(lr){8-9}
        \cmidrule(lr){10-11}
        \cmidrule(lr){12-13}
        & Standard
        & Isotropic
        & Standard
        & Isotropic
        & Standard
        & Isotropic
        & Standard
        & Isotropic
        & Standard
        & Isotropic
        & Standard
        & Isotropic \\
        \midrule
        $500{\times}2$
        & $\boxed{\mathbf{97.18{\pm}0.03}}$
        & $94.38{\pm}0.04$
        & $\mathbf{93.26{\pm}0.09}$
        & $86.69{\pm}0.20$
        & $\mathbf{87.58{\pm}0.23}$
        & $85.69{\pm}0.14$
        & $\mathbf{44.70{\pm}0.11}$
        & $44.16{\pm}0.09$
        & $\boxed{\mathbf{18.70{\pm}0.07}}$
        & $18.60{\pm}0.11$
        & $\boxed{\mathbf{14.22{\pm}0.19}}$
        & $12.88{\pm}0.16$ \\

        $500{\times}3$
        & $\mathbf{97.07{\pm}0.04}$
        & $95.28{\pm}0.11$
        & $\mathbf{93.24{\pm}0.02}$
        & $87.81{\pm}0.18$
        & $\boxed{\mathbf{87.77{\pm}0.09}}$
        & $\boxed{86.13{\pm}0.16}$
        & $\boxed{45.19{\pm}0.13}$
        & $\mathbf{45.36{\pm}0.21}$
        & $18.62{\pm}0.15$
        & $\boxed{\mathbf{19.62{\pm}0.10}}$
        & $13.45{\pm}0.16$
        & $\boxed{\mathbf{13.63{\pm}0.12}}$ \\

        $500{\times}4$
        & $\mathbf{96.83{\pm}0.08}$
        & $95.25{\pm}0.09$
        & $\mathbf{93.26{\pm}0.05}$
        & $88.32{\pm}0.06$
        & $\mathbf{86.96{\pm}0.13}$
        & $86.04{\pm}0.25$
        & $44.83{\pm}0.18$
        & $\mathbf{45.92{\pm}0.16}$
        & $17.23{\pm}0.15$
        & $\mathbf{18.91{\pm}0.10}$
        & $12.63{\pm}0.21$
        & $\mathbf{12.82{\pm}0.15}$ \\

        $500{\times}5$
        & $\mathbf{96.61{\pm}0.08}$
        & $\boxed{95.37{\pm}0.07}$
        & $\boxed{\mathbf{93.27{\pm}0.09}}$
        & $\boxed{89.80{\pm}0.20}$
        & $\mathbf{86.66{\pm}0.30}$
        & $85.77{\pm}0.32$
        & $43.49{\pm}0.28$
        & $\mathbf{45.95{\pm}0.34}$
        & $15.97{\pm}0.09$
        & $\mathbf{17.22{\pm}0.05}$
        & $11.66{\pm}0.38$
        & $\mathbf{12.41{\pm}0.09}$ \\

        $1000{\times}2$
        & $\mathbf{96.98{\pm}0.05}$
        & $94.33{\pm}0.06$
        & $\mathbf{93.15{\pm}0.02}$
        & $86.62{\pm}0.17$
        & $\mathbf{86.66{\pm}0.27}$
        & $85.62{\pm}0.15$
        & $42.18{\pm}0.34$
        & $\mathbf{44.19{\pm}0.08}$
        & $17.74{\pm}0.06$
        & $\mathbf{18.68{\pm}0.10}$
        & $12.71{\pm}0.12$
        & $\mathbf{12.93{\pm}0.05}$ \\

        $1000{\times}3$
        & $\mathbf{96.77{\pm}0.07}$
        & $95.33{\pm}0.09$
        & $\mathbf{93.04{\pm}0.20}$
        & $88.19{\pm}0.10$
        & $\mathbf{86.72{\pm}0.16}$
        & $86.02{\pm}0.13$
        & $42.76{\pm}0.24$
        & $\mathbf{45.45{\pm}0.22}$
        & $16.68{\pm}0.10$
        & $\mathbf{19.59{\pm}0.10}$
        & $12.44{\pm}0.22$
        & $\mathbf{13.54{\pm}0.09}$ \\

        $1000{\times}4$
        & $\mathbf{96.74{\pm}0.04}$
        & $95.32{\pm}0.18$
        & $\mathbf{92.61{\pm}0.16}$
        & $88.49{\pm}0.05$
        & $\mathbf{86.08{\pm}0.18}$
        & $85.97{\pm}0.28$
        & $41.97{\pm}0.39$
        & $\boxed{\mathbf{46.02{\pm}0.12}}$
        & $14.45{\pm}0.19$
        & $\mathbf{18.73{\pm}0.11}$
        & $10.57{\pm}0.18$
        & $\mathbf{13.07{\pm}0.07}$ \\

        $1000{\times}5$
        & $\mathbf{96.60{\pm}0.13}$
        & $95.15{\pm}0.12$
        & $\mathbf{92.22{\pm}0.03}$
        & $89.69{\pm}0.11$
        & $85.60{\pm}0.12$
        & $\mathbf{85.98{\pm}0.22}$
        & $39.56{\pm}0.25$
        & $\mathbf{45.58{\pm}0.57}$
        & $12.31{\pm}0.21$
        & $\mathbf{17.02{\pm}0.15}$
        & $8.40{\pm}0.53$
        & $\mathbf{12.37{\pm}0.10}$ \\
        \bottomrule
    \end{tabular}%
    }
\end{table}
    The trend appears quite stable, showing a preference towards isotropic-tanh for more conventionally challenging image classification datasets, particularly for deeper MLPs --- with the converse for standard-tanh. This does demonstrate that their optimal regimes are dataset and architecture-dependent; however, isotropic-tanh is a more encouraging result for broader application, demonstrating preferable scalability with challenging datasets and deeper architectures. CIFAR and CALTECH256 results demonstrate a considerable advantage typically for isotropic-tanh, whilst MNIST and EMNIST-AJ favour standard-tanh, with a narrower performance gap for Fashion-MNIST.

    Further tendencies emerge: standard-tanh seems to decline considerably with increasing depth or width, a phenomenon much more suppressed in the isotropic formulation. This suggests that the isotropic-tanh may alleviate depth-related pathologies present in standard-tanh networks --- perhaps related to their Jacobians (\textit{Sec.}~\ref{Sec:Jacobians}) and their respective propensities for vanishing gradients \cite{Bengio1994}.

    Overall, this is a promising tendency for isotropic-tanh; it is superior for deeper-wider networks, whilst standard-tanh's regime is confined to narrower-shallower networks. This reliable superiority of isotropic tanh across these harder datasets and larger architectural regimes does alleviate some concerns raised in \textit{App.}~\ref{App:Limitations} regarding expressibility of isotropic functions; however, this must be considered only a single instance of isotropic-tanh and not yet expanded more broadly. This remains encouraging, considering that the standard tanh was a popular choice historically, before the advent of more modern activation functions. Thus, development from isotropic-tanh's foundation may be undertaken to a similar effect, via a naive search over function space or theoretical developments.

\newpage
\input{checklist.tex}


    
    
%

\end{document}

%% file: checklist.tex
\section*{NeurIPS Paper Checklist}

\begin{enumerate}

\item {\bf Claims}
    \item[] Question: Do the main claims made in the abstract and introduction accurately reflect the paper's contributions and scope?
    \item[] Answer: \answerYes{} 
    \item[] Justification: The abstract discusses several points: adaptive dense networks, a buffer of thresholded scaffold neurons, symmetry-led orthogonal isotropic activation functions, one-to-one ordered connectivity, function invariance, sparsification to asymptotic 50\%, adaptive networks to changing task demands, centred on primitive symmetry-prescriptions, neuron deindividuation, intrinsic length, parallel precomputed matrix-vector products for a generalised architecture and relation to interpretability (through ICA). Thus all of which is borne out theoretically or experimentally in the paper.

\item {\bf Limitations}
    \item[] Question: Does the paper discuss the limitations of the work performed by the authors?
    \item[] Answer: \answerYes{} 
    \item[] Justification: Dedicated limitations section in the appendices and main-text conclusion, covering the stated points below.

\item {\bf Theory assumptions and proofs}
    \item[] Question: For each theoretical result, does the paper provide the full set of assumptions and a complete (and correct) proof?
    \item[] Answer: \answerYes{} 
    \item[] Justification: The mathematics utilised in this paper is an implementation of standard results of group theory, representation theory, tensor-rank transformations, and singular value decomposition, applied to various aspects of the network. For claims stated, assumptions are clearly specified, including isotropic activation functions for commutation and asymptotic assumptions. Due to the standard nature of group- and representation-theoretic usage, these parts do not contain formal proofs, as they are well established in their fields and are only used in novel ways for these multilayer perceptrons. All usage and statements around linear algebra transforms are derived. It conforms to standard practice regarding this style of mathematically principled implementation paper.

    \item {\bf Experimental result reproducibility}
    \item[] Question: Does the paper fully disclose all the information needed to reproduce the main experimental results of the paper to the extent that it affects the main claims and/or conclusions of the paper (regardless of whether the code and data are provided or not)?
    \item[] Answer: \answerYes{} 
    \item[] Justification: Yes full experimental appendix section is provided, detailing all aspects needed for successful reproducibility along with justification around the experiments. The code is also provided, at the stated GitHub link.

\item {\bf Open access to data and code}
    \item[] Question: Does the paper provide open access to the data and code, with sufficient instructions to faithfully reproduce the main experimental results, as described in supplemental material?
    \item[] Answer: \answerYes{} 
    \item[] Justification: Yes, code is available at the stated GitHub link and all datasets are established publically-accessible datasets.

\item {\bf Experimental setting/details}
    \item[] Question: Does the paper specify all the training and test details (e.g., data splits, hyperparameters, how they were chosen, type of optimizer) necessary to understand the results?
    \item[] Answer: \answerYes{} 
    \item[] Justification: Please see appendix detailing all experimental setup.

\item {\bf Experiment statistical significance}
    \item[] Question: Does the paper report error bars suitably and correctly defined or other appropriate information about the statistical significance of the experiments?
    \item[] Answer: \answerYes{} 
    \item[] Justification: All plots and tabulated data include error bars. Exception being the ensemble ridgeline plot of experiment one, which cannot display errors by its nature, but is still an aggregate over many independent repeats.

\item {\bf Experiments compute resources}
    \item[] Question: For each experiment, does the paper provide sufficient information on the computer resources (type of compute workers, memory, time of execution) needed to reproduce the experiments?
    \item[] Answer: \answerYes{} 
    \item[] Justification: Time of execution was explicitly stated with errors, alongside relevant computing hardware.
    
\item {\bf Code of ethics}
    \item[] Question: Does the research conducted in the paper conform, in every respect, with the NeurIPS Code of Ethics \url{https://neurips.cc/public/EthicsGuidelines}?
    \item[] Answer: \answerYes{} 
    \item[] Justification: Yes, it conforms.

\item {\bf Broader impacts}
    \item[] Question: Does the paper discuss both potential positive societal impacts and negative societal impacts of the work performed?
    \item[] Answer: \answerNA{} 
    \item[] Justification: This is an adaptive architecture methodology, and a group/representation-theoretic approach to network primitives. It is evaluated on simple image-classification settings, and is not introduced as a deployed system, generative model, decision system etc. There is no direct pathway to societal harm.
    
    Neither of which constitutes a broader societal impact. 
    
\item {\bf Safeguards}
    \item[] Question: Does the paper describe safeguards that have been put in place for responsible release of data or models that have a high risk for misuse (e.g., pre-trained language models, image generators, or scraped datasets)?
    \item[] Answer: \answerNA{} 
    \item[] Justification: Models are simple image classifiers.

\item {\bf Licenses for existing assets}
    \item[] Question: Are the creators or original owners of assets (e.g., code, data, models), used in the paper, properly credited and are the license and terms of use explicitly mentioned and properly respected?
    \item[] Answer: \answerYes{} 
    \item[] Justification: Yes all datasets are properly cited and code is personally written by the authors.

\item {\bf New assets}
    \item[] Question: Are new assets introduced in the paper well documented and is the documentation provided alongside the assets?
    \item[] Answer: \answerNA{} 
    \item[] Justification: No new assets.

\item {\bf Crowdsourcing and research with human subjects}
    \item[] Question: For crowdsourcing experiments and research with human subjects, does the paper include the full text of instructions given to participants and screenshots, if applicable, as well as details about compensation (if any)? 
    \item[] Answer: \answerNA{} 
    \item[] Justification: No crowdsourcing.

\item {\bf Institutional review board (IRB) approvals or equivalent for research with human subjects}
    \item[] Question: Does the paper describe potential risks incurred by study participants, whether such risks were disclosed to the subjects, and whether Institutional Review Board (IRB) approvals (or an equivalent approval/review based on the requirements of your country or institution) were obtained?
    \item[] Answer: \answerNA{} 
    \item[] Justification: No human subjects.

\item {\bf Declaration of LLM usage}
    \item[] Question: Does the paper describe the usage of LLMs if it is an important, original, or non-standard component of the core methods in this research? Note that if the LLM is used only for writing, editing, or formatting purposes and does \emph{not} impact the core methodology, scientific rigor, or originality of the research, declaration is not required.
    \item[] Answer: \answerNA{} 
    \item[] Justification: LLM usage declared. Used in grammatical/synonym/phrasing finding capacity for the text, and used in some aspects of the code to improve the presentation of plots using matplotlib in a way described by the authors. Similarly, some code helper functions were written by LLM to save time but were reviewed carefully.

\end{enumerate}